\documentclass[%
 reprint,
 superscriptaddress,
 amsmath,amssymb,
 aps,
 pre,
]{revtex4-2}

\usepackage{dcolumn}
\usepackage{bm}
\usepackage{mathtools}
\usepackage{graphicx}
\usepackage{bm}
\usepackage{microtype}
\usepackage[dvipsnames]{xcolor}
\usepackage[colorlinks=true,allcolors=NavyBlue]{hyperref}

\DeclareMathOperator*{\argmin}{arg\,min}

\DeclareMathOperator{\tr}{tr}

\begin{document}

\title{Contrasting random and learned features in deep Bayesian linear regression}

\author{Jacob A. Zavatone-Veth}
\email{jzavatoneveth@g.harvard.edu}
\affiliation{Department of Physics, Harvard University, Cambridge, Massachusetts 02138, USA}
\affiliation{Center for Brain Science, Harvard University, Cambridge, Massachusetts 02138, USA}

\author{William L. Tong}
\email{wtong@g.harvard.edu}
\affiliation{John A. Paulson School of Engineering and Applied Sciences, Harvard University, Cambridge, Massachusetts 02138, USA}

\author{Cengiz Pehlevan}
\email{cpehlevan@seas.harvard.edu}
\affiliation{John A. Paulson School of Engineering and Applied Sciences, Harvard University, Cambridge, Massachusetts 02138, USA}
\affiliation{Center for Brain Science, Harvard University, Cambridge, Massachusetts 02138, USA}

\date{\today}

\begin{abstract}
Understanding how feature learning affects generalization is among the foremost goals of modern deep learning theory. Here, we study how the ability to learn representations affects the generalization performance of a simple class of models: deep Bayesian linear neural networks trained on unstructured Gaussian data. By comparing deep random feature models to deep networks in which all layers are trained, we provide a detailed characterization of the interplay between width, depth, data density, and prior mismatch. We show that both models display sample-wise double-descent behavior in the presence of label noise. Random feature models can also display model-wise double-descent if there are narrow bottleneck layers, while deep networks do not show these divergences. Random feature models can have particular widths that are optimal for generalization at a given data density, while making neural networks as wide or as narrow as possible is always optimal. Moreover, we show that the leading-order correction to the kernel-limit learning curve cannot distinguish between random feature models and deep networks in which all layers are trained. Taken together, our findings begin to elucidate how architectural details affect generalization performance in this simple class of deep regression models. 
\end{abstract}


\maketitle

\section{Introduction}

Deep neural networks (NNs) display a rich and often-perplexing spectrum of generalization behaviors. Highly overparameterized NNs may possess the expressivity to fit random noise, yet in practice can still generalize well to unseen data \cite{zhang2021understanding,belkin2019reconciling}. The ability of NNs to flexibly learn features from data is widely believed to be a critical contributor to their practical success \cite{yang2021feature,aitchison2020bigger,zhang2021understanding,belkin2019reconciling}, but the precise contributions of feature learning to their generalization behavior remain incompletely understood \cite{nakkiran2021deep,aitchison2020bigger,yang2019scaling,zhang2021understanding,belkin2019reconciling,yang2021feature,refinetti21fail,woodworth2020kernel,geiger2020lazy,geiger2020scaling}.

In recent years, intensive theoretical work has begun to elucidate the properties of deep networks in the limit of infinite hidden layer width. In this limit, a dramatic simplification occurs, and inference in deep networks is equivalent to kernel regression or classification  \cite{neal1996priors,williams1997computing,lee2018deep,matthews2018gaussian,jacot2018neural,lee2018deep,hron2020exact,yang2019scaling}. This correspondence has enabled detailed characterizations of inference at infinite width in both maximum-likelihood and fully Bayesian settings, providing new insights into the inductive biases that allow deep networks to overfit benignly \cite{mei2019generalization,hu2020universality,spigler2020asymptotic,canatar2021spectral,barbier2021performance,dascoli2021triple,adlam2020neural,adlam2020understanding,dascoli2020double,jin2022learning,loureiro2021learning}. Yet, understanding inference in the kernel limit is not sufficient, because kernel descriptions cannot capture feature learning \cite{lee2020finite,yang2021feature,refinetti21fail,woodworth2020kernel,geiger2020lazy}.

As a result, a growing number of recent works have aimed to study the behavior of networks near the kernel limit, with the hope that leading-order corrections to the large-width behavior might elucidate how width and depth affect inference \cite{antognini2019finite,yaida2020,zv2021exact,roberts2022principles,grosvenor2021edge,halverson2021neural,zv2021asymptotics,zv2021scale,li2021statistical,naveh2021predicting,naveh2021self,aitchison2020bigger,dyer2020asymptotics,aitken2020asymptotics}. Some of these works focus on the properties of the function-space prior distribution \cite{antognini2019finite,yaida2020,zv2021exact,grosvenor2021edge,roberts2022principles,halverson2021neural}, some consider maximum-likelihood inference with gradient descent \cite{dyer2020asymptotics,aitken2020asymptotics,roberts2022principles}, and some consider properties of the full Bayes posterior \cite{yaida2020,zv2021asymptotics,li2021statistical,naveh2021predicting,naveh2021self,roberts2022principles,aitchison2020bigger,zv2021scale}. This body of research has resulted in several conjectural conditions under which when narrower and deeper networks might perform better than their infinitely-wide cousins in the Bayesian setting, as measured by generalization for fixed data \cite{zv2021asymptotics,li2021statistical,naveh2021predicting} or by some alternative criterion based on entropic considerations \cite{roberts2022principles}. 

However, previous studies of Bayesian neural network generalization near the kernel limit have not clearly differentiated the effect of width on feature learning from its other potential effects on inference. Concretely, it is not clear whether potential improvements in generalization afforded by the leading finite-width correction reflect the benefits of feature learning, or if a similar gain would be observed in random feature models, where only the readout layer is trained. Here, we explore how random and learned features affect generalization in the simplest class of Bayesian NNs---deep linear models---when trained on unstructured, noisy data. By developing a detailed understanding of this simple setting, one might hope to gain intuition that may prove useful in studying more complex networks \cite{saxe2013exact,fukumizu1998effect,nakkiran2019more,hastie2019surprises,advani2020high,zv2021exact,wilson2020bayesian,wenzel2020good,izmailov2021bayesian}. 

In this work, we study the asymptotic generalization performance of deep linear Bayesian regression for data generated with an isotropic Gaussian covariate model. Using the replica trick \cite{mezard1987spin,engel2001statistical}, we compute learning curves for simple linear regression, deep linear Gaussian random feature (RF) models, and deep linear NNs. Our results are obtained using an isotropic Gaussian likelihood in the limit of small likelihood variance, which renders this analysis analytically tractable \cite{zv2021asymptotics,zv2021scale}. Using alternative replica-free methods and numerical simulation, we show that the predictions obtained under a replica-symmetric (RS) \emph{Ansatz} are accurate for all three model classes. In particular, the RS result for learning curves of NNs with hidden layers of equal widths is consistent with results obtained by Li and Sompolinsky \cite{li2021statistical} using a different approximation method. 

In the presence of label noise, both RF and NN models display sample-wise non-monotonicity in their learning curves. As we work in a high-dimensional limit, this non-monontonicity is of a particularly extreme form: the generalization error diverges at a particular data density. In keeping with modern deep learning parlance, we refer to this behavior as ``double-descent," though this monotonicity can arise from distinct effects in different settings \cite{belkin2019reconciling,krogh1992generalization,hastie2019surprises,nakkiran2019more,nakkiran2021deep,dascoli2020double,dascoli2021triple,adlam2020neural,adlam2020understanding,canatar2021spectral,geiger2020scaling,mei2019generalization}. If one introduces a bottleneck layer that is narrower than the input dimension, an RF model will display model-wise double-descent behavior at fixed data density---or equivalently sample-wise double-descent at fixed width---even in the absence of label noise, while an NN model will not show this divergence. This distinct small-width behavior shows one advantage afforded by the flexibility to learn features. For both models, we analyze how optimal network architecture depends on data density and prior mismatch. We show that, at a given data density, RF models have a particular optimal width for fixed depth and optimal depth for fixed width that minimizes the generalization error. In contrast, it is always optimal to take an NN to be as wide or as narrow as possible, depending on the regime.

We further analyze models of arbitrary depth perturbatively in the limit in which the network depth and dataset size are small relative to the hidden layer widths, connecting these results to those of previous work on fixed-dataset perturbation theory \cite{zv2021asymptotics}. We find that the leading order correction to the large-width behavior of RF and NN models is identical, hence first-order perturbation theory for the generalization error cannot distinguish between random and learned features. To distinguish between training only the readout layer and training all layers, one must go to second order in perturbation theory. Therefore, at large widths, the ability to perform representation learning provides only a small advantage in generalization performance in these simple models relative to random features, which is invisible in first-order perturbation theory. In total, our results provide new insight into how the generalization behavior of deep Bayesian linear regression in high dimensions depends on architectural details. Moreover, they shed light onto which qualitative features of generalization behavior can or cannot be captured by low-order perturbative corrections \cite{zv2021exact}.

\section{Problem setting}

In this section, we introduce the three classes of regression models we consider in this work, as well as our generative data model. Our notation throughout is standard; we use $\Vert \cdot \Vert$ to denote the Euclidean norm, $\mathbf{I}_{d}$ to denote the $d \times d$ identity matrix, and $\mathbf{1}$ to denote the vector with all elements equal to one. 

\subsection{Regression models and parameter priors}

In this work, we consider three classes of scalar Bayesian linear regression models for a scalar-valued function of $d$-dimensional inputs. All three of these model classes are of the form
\begin{align}
\begin{split}
    g_{\mathbf{w}}(\mathbf{x}) = \frac{1}{\sqrt{d}} \mathbf{w}^{\top} \mathbf{x}, 
\end{split}
\end{align}
and differ in the parameterization of the `end-to-end' weight vector $\mathbf{w} \in \mathbb{R}^{d}$. We  will choose parameter priors such that $\mathbb{E} \Vert \mathbf{w} \Vert^{2} = \sigma^{2} d$ for each model, where $\sigma > 0$ is a hyperparameter which sets the prior variance of the network outputs. We remark that each model is positive-homogeneous in its parameters, hence this choice is made without loss of generality. In all cases, the parameter priors are isotropic and Gaussian, as is standard in Bayesian deep learning \cite{wilson2020bayesian,neal1996priors,izmailov2021bayesian,wenzel2020good,zv2021exact,lee2018deep,matthews2018gaussian,williams1997computing}. 

Below, we list the three classes of models we consider, and introduce a two-letter abbreviation for each:
\begin{itemize}

\item[(LR)] Simple Bayesian linear regression. For this model, the end-to-end weight vector is directly parameterized as
\begin{align}
    \mathbf{w}_{\textrm{LR}} = \sigma \mathbf{v} 
\end{align}
for a trainable parameter vector $\mathbf{v} \in\mathbb{R}^{d}$ with isotropic Gaussian prior distribution
\begin{align}
    \mathbf{v} \sim \mathcal{N}(\mathbf{0},\mathbf{I}_{d}).
\end{align}
Previous works have extensively studied this model in both maximum-likelihood and fully Bayesian settings \cite{krogh1992generalization,nakkiran2019more,hastie2019surprises,canatar2021spectral,advani2016statistical,barbier2021performance}, hence we include it as a baseline against which we will compare our results for more complicated models. 

\item[(RF)] Deep Bayesian random feature models. For these models, the weight vector is parameterized as 
\begin{align}
    \mathbf{w}_{\textrm{RF}} = \frac{\sigma}{\sqrt{n_{1} \cdots n_{\ell}}} \mathbf{U}_{1} \cdots \mathbf{U}_{\ell} \mathbf{v} 
\end{align}
for matrices $\mathbf{U}_{1} \in \mathbb{R}^{d \times n_{1}}$,  $\mathbf{U}_{2} \in \mathbb{R}^{n_{1} \times n_{2}}$,\ldots, $\mathbf{U}_{\ell} \in \mathbb{R}^{n_{\ell-1} \times n_{\ell}}$ and a vector $\mathbf{v} \in \mathbb{R}^{n_{\ell}}$. Here, $\ell \in \mathbb{N}_{>0}$ is the network depth, while $n_{1},\ldots,n_{\ell} \in \mathbb{N}_{>0}$ are the hidden layer widths. For the RF model, only the readout weight vector $\mathbf{v}$ is trainable, while the hidden layer weights $\mathbf{U}_{l}$ are fixed and random. We choose an isotropic Gaussian prior for the readout weights
\begin{align}
    \mathbf{v} \sim \mathcal{N}(\mathbf{0},\mathbf{I}_{n_{\ell}}),
\end{align}
while the hidden layer weights are drawn from a fixed isotropic Gaussian distribution
\begin{align}
    (\mathbf{U}_{l})_{ij} &\sim \mathcal{N}(0,1) \qquad (l = 1, \ldots, \ell).
\end{align}

\item[(NN)] Deep Bayesian linear neural networks. For these models, the weight vector is parameterized as 
\begin{align}
    \mathbf{w}_{\textrm{NN}} = \frac{\sigma}{\sqrt{n_{1} \cdots n_{\ell}}} \mathbf{U}_{1} \cdots \mathbf{U}_{\ell} \mathbf{v} . 
\end{align}
Though NNs are parameterized identically to the RF models above, they differ in that all of the weights are trainable, not only the readout. We again choose isotropic Gaussian prior distributions
\begin{align}
    (\mathbf{U}_{l})_{ij} &\sim \mathcal{N}(0,1) \qquad (l = 1, \ldots, \ell),
    \\
    \mathbf{v} &\sim \mathcal{N}(\mathbf{0},\mathbf{I}_{n_{\ell}}).
\end{align}
From a physical perspective, the hidden layer weights in the RF model are `quenched' disorder, whereas they are `annealed' disorder in NNs \cite{mezard1987spin,engel2001statistical}. 

\end{itemize}

For all models, we denote expectation with respect to the prior distribution of the trainable parameters by $\mathbb{E}_{\mathcal{W}}$. 

\subsection{Data model and the Bayes posterior}

We train all models on a dataset $\{(\mathbf{x}_{\mu},y_{\mu})\}_{\mu=1}^{p}$ of $p$ examples, generated according to a standard isotropic Gaussian covariate model \cite{hastie2019surprises,barbier2021performance,advani2016statistical,krogh1992generalization,nakkiran2019more,advani2020high}. In this model, the example inputs are independent and identically distributed samples from a standard Gaussian distribution:
\begin{align}
    \mathbf{x}_{\mu} \sim \mathcal{N}(\mathbf{0},\mathbf{I}_{d}), 
\end{align}
while the labels are generated by a ground truth linear model, possibly corrupted by additive Gaussian noise: 
\begin{align}
    y_{\mu} = \frac{1}{\sqrt{d}} \mathbf{w}_{\ast}^{\top} \mathbf{x}_{\mu} + \eta \xi_{\mu},
\end{align}
where $\eta \geq 0$ sets the noise variance. The noise variables are independent and identically distributed as
\begin{align}
    \xi_{\mu} \sim \mathcal{N}(0,1),
\end{align}
and are independent of the inputs. We take the `teacher' weight vector $\mathbf{w}_{\ast}$ to have fixed norm $\Vert \mathbf{w}_{\ast} \Vert^2 = d$. In some places, we will average over teacher weights distributed uniformly on the sphere (i.e., $\mathbf{w}_{\ast} \sim \mathcal{U}[\mathbb{S}^{d-1}(\sqrt{d})]$), though our main results will hold pointwise for any $\mathbf{w}_{\ast}$ on the sphere. We will collect the training inputs and outputs into a matrix $(\mathbf{X})_{\mu j} = (\mathbf{x}_{\mu})_{j}$ and a vector $(\mathbf{y})_{\mu} = y_{\mu}$, respectively. 

For a dataset thusly generated, we introduce an isotropic Gaussian likelihood of variance $1/\beta$:
\begin{align}
    p(\{(\mathbf{x}_{\mu},y_{\mu})\}_{\mu=1}^{p} \,|\,\mathcal{W}) \propto \exp\left(-\frac{\beta}{2} \sum_{\mu=1}^{p} [g_{\mathbf{w}}(\mathbf{x}_{\mu}) - y_{\mu}]^2 \right),
\end{align}
where $\mathcal{W}$ denotes the set of trainable parameters for a given model, and the normalization constant is implied. We will refer to $\beta$ as the  `inverse temperature' by standard analogy with statistical mechanics \cite{zv2021asymptotics,krogh1992generalization,engel2001statistical,biehl1998phase,solla1992learning,levin1990statistical}. Then, the partition function of the resulting Bayes posterior is given as
\begin{align}
    Z = \mathbb{E}_{\mathcal{W}} \exp\left(-\frac{\beta}{2} \sum_{\mu=1}^{p} [g_{\mathbf{w}}(\mathbf{x}_{\mu}) - y_{\mu}]^2 \right).
\end{align}
We denote expectations with respect to this Bayes posterior by $\langle \cdot \rangle$.

\subsection{Generalization error in the thermodynamic limit}

With the initial setup of the previous sections, we can now introduce our concrete objective. We consider a proportional asymptotic limit in which the input dimension $d$, the dataset size $p$, and (for NN and RF models) the hidden layer widths $n_{1}, \ldots, n_{\ell}$ tend to infinity for fixed depth $\ell$ and fixed ratios 
\begin{align}
    \alpha &\equiv p/d = \mathcal{O}(1),
    \\
    \gamma_{l} &\equiv n_{l}/d = \mathcal{O}(1) \qquad (l = 1, \ldots, \ell).
\end{align}
Moreover, we focus on the zero-temperature limit $\beta \to \infty$, in which the likelihood tends to a constraint that the network interpolates the training set with probability one. In the noise-free case, this limiting likelihood is matched to the true generative model of the data, but it is clearly mismatched in the presence of label noise. This limit has been considered in several recent studies of deep linear Bayesian neural networks \cite{zv2021asymptotics,zv2021scale,li2021statistical,aitchison2020bigger,roberts2022principles}.

Our goal is to study the average-case generalization error $\epsilon$ of the resulting model, as measured by the deviation of its end-to-end weight vector $\mathbf{w}$ from the true teacher weight vector $\mathbf{w}_{\ast}$:
\begin{align} \label{eqn:generalization_error}
    \epsilon = \lim_{\beta \to \infty} \lim_{d,p,n_{1},\ldots,n_{\ell} \to \infty} \mathbb{E}_{\mathcal{D}} \left\langle \frac{1}{d} \Vert \mathbf{w} - \mathbf{w}_{\ast} \Vert^{2} \right\rangle.
\end{align}
Here, $\mathbb{E}_{\mathcal{D}}$ denotes expectation with respect to all quenched disorder for a given model. For all models, this includes the training inputs and label noise; for the RF model, it also includes the hidden layer weights. We emphasize that this choice means that we do not ensemble RF model predictions over realizations of the features; rather, we consider the average-case generalization error of individual realizations \cite{mei2019generalization,dascoli2020double,dascoli2021triple,adlam2020neural,adlam2020understanding}. 

We remark that \eqref{eqn:generalization_error} is the average-case error of the Gibbs estimator (i.e., a single sample from the posterior); one could instead consider the error of the mean estimator $\langle \mathbf{w} \rangle$. For the LR and RF models, this corresponds to studying Bayesian minimum mean-squared error (MMSE) inference \cite{advani2016statistical,barbier2021performance}. As one has the thermal bias-variance decomposition
\begin{align}
    \langle \Vert \mathbf{w} - \mathbf{w}_{\ast} \Vert^2 \rangle &= \Vert \langle \mathbf{w} \rangle - \mathbf{w}_{\ast} \Vert^2 \nonumber\\&\quad + \tr(\langle \mathbf{w} \mathbf{w}^{\top} \rangle - \langle \mathbf{w} \rangle \langle \mathbf{w} \rangle^{\top}),
\end{align}
our results include an additional contribution to the generalization error from the posterior covariance $\langle \mathbf{w} \mathbf{w}^{\top} \rangle - \langle \mathbf{w} \rangle \langle \mathbf{w} \rangle^{\top}$ of the end-to-end weight vector, which is not identically zero. If one considered an alternative low-temperature limit in which the prior variance is proportional to $1/\beta$, then this additional contribution would vanish in the low-temperature limit. Our choice of scaling is motivated by the considerations described in our previous work \cite{zv2021asymptotics}, and is the one classically used in studies of the statistical mechanics of Bayesian inference \cite{biehl1998phase,solla1992learning,levin1990statistical,li2021statistical}. This choice is important as it affects the relationship of our results to those in the setting of ridge regression. As discussed in Appendices \ref{app:rf_posterior_expectations} and \ref{app:nn_posterior_expectations}, in our previous work \cite{zv2021asymptotics}, and in previous works of Advani and Ganguli \cite{advani2016statistical} and Barbier \emph{et al.} \cite{barbier2021performance}, the zero-temperature limit of the MMSE estimator would in this case coincide with the ridge regression estimator.

We compute the limiting average generalization error using the replica method, a non-rigorous but powerful heuristic that has seen broad use in statistical mechanical studies of inference \cite{engel2001statistical,mezard1987spin,canatar2021spectral,dascoli2020double,loureiro2021learning}. As our main results can be understood independently of calculation through which they were obtained, we relegate the details to Appendices \ref{app:replica_framework} and \ref{app:rs_saddle_point}. We note the important caveat that our main results are obtained under a replica-symmetric \emph{Ansatz}. We expect this assumption to hold exactly for the LR and RF models by virtue of the concavity of their log-posteriors, but replica symmetry may be broken in deep linear NNs \cite{barbier2021strong,mezard1987spin}. We will not address this possibility analytically by considering \emph{Ans\"atze} with broken replica symmetry \cite{mezard1987spin}, but will instead simply compare the RS predictions against results obtained through a combination of alternative analytical methods and numerics.

\section{Learning curves for the LR model} \label{sec:lr_model}

We begin by briefly describing the learning curve of the simple LR model. Our result extends the classic result of Krogh and Hertz \cite{krogh1992generalization} for ridge regression in the ridgeless limit to the Bayesian setting: 
\begin{align}\label{eqn:lr_learning_curve}
    \epsilon_{\textrm{LR}} = 
    \begin{dcases} 
    (1 + \sigma^{2}) (1-\alpha) + \frac{\alpha}{1-\alpha} \eta^{2}, &\textrm{if}\ \alpha < 1
    \\
    \frac{1}{\alpha - 1} \eta^{2},  &\textrm{if}\ \alpha > 1. 
    \end{dcases}
\end{align}
For this simple model, the learning curve can also be computed directly by first evaluating the posterior average defining $\epsilon_{\textrm{LR}}$ for a fixed realization of the disorder, and then averaging the result over the disorder in the zero-temperature limit (see Appendix \ref{app:rf_posterior_expectations} for details). The result of \cite{krogh1992generalization} can be recovered from \eqref{eqn:lr_learning_curve} by setting $\sigma = 0$. We provide further discussion of the relationship between the Bayesian LR model in the zero-temperature limit and ridge regression in the ridgeless limit in Appendix \ref{app:rf_posterior_expectations}. 

Therefore, as \cite{krogh1992generalization} found in the ridge regression setting, the LR model exhibits sample-wise double-descent behavior---i.e., non-monotonicity in $\epsilon_{\textrm{LR}}$ as a function of $\alpha$ \cite{belkin2019reconciling,nakkiran2021deep}---in the presence of label noise. In the thermodynamic limit, the double-descent behavior is particularly striking: $\epsilon_{\textrm{LR}}$ diverges as $\alpha \to 1$. In the absence of noise, $\epsilon_{\textrm{LR}}$ decreases monotonically from $1+\sigma^2$ to $0$ as $\alpha \uparrow 1$, and then remains at zero for all $\alpha > 1$. We remark that, for this and subsequent models, we will not conduct a detailed analysis of what happens precisely at exceptional points, e.g., $\alpha = 1$. In the ridge regression setting, the phase transition at $\alpha = 1$ has recently been analyzed in detail by Canatar \emph{et al.} \cite{canatar2021spectral}. We also direct the interested reader to an expository note by Nakkiran \cite{nakkiran2019more} for further intuitions on double-descent in ridge regression, and to work by Hastie \emph{et al.} \cite{hastie2019surprises} for a detailed rigorous analysis. We will take the model-wise double-descent behavior of the LR model as a benchmark for our subsequent analyses of the more complex RF and NN models. 

\section{Learning curves for the RF model} \label{sec:rf_model}

\subsection{Learning curve and double-descent behavior }

\begin{figure*}
    \centering
    \includegraphics[width=4in]{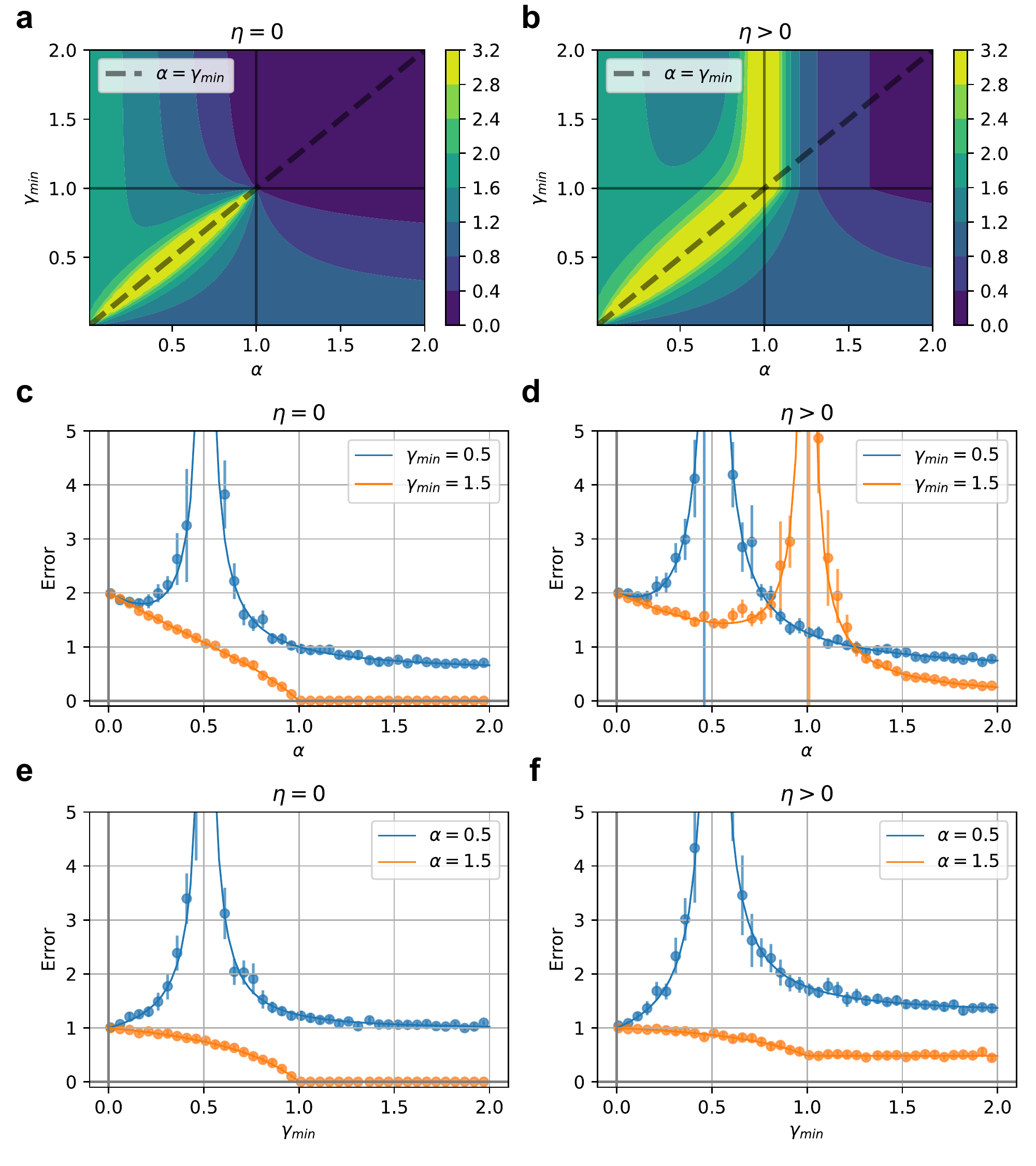}
    \caption{Sample- and model-wise double-descent in deep Bayesian random feature models.
    \textbf{(a).} Contour plot in $(\alpha,\gamma)$-space of the theoretical error surface $\epsilon_{\textrm{RF}}$ \eqref{eqn:rf_learning_curve} for a single-hidden-layer RF model in the absence of label noise ($\eta = 0$). For all panels, we set the input dimensionality $d = 100$ and prior variance $\sigma^2 = 1$. For details of our numerical methods, see Appendix \ref{app:numerical_methods}.
    \textbf{(b).} As in (a)., but in the presence of label noise ($\eta = 0.5$).
    \textbf{(c).} Horizontal cross sections of above (a). Theory curves are overlayed with experiment points, plotted with $\pm 2$ SE bars. 
    \textbf{(d).} Horizontal cross sections of above (b).
    \textbf{(e).} Vertical cross sections of above (a).
    \textbf{(f).} Vertical cross sections of above (b).
    }
    \label{fig:rf_learning_curve}
\end{figure*}

For RF models, we obtain a closed-form expression for the learning curve at any depth. Let $\gamma_{\textrm{min}} = \min\{\gamma_{1},\ldots,\gamma_{\ell}\}$ be the minimum hidden layer width. Then, we find that
\begin{widetext}
\begin{align} \label{eqn:rf_learning_curve}
    \epsilon_{\textrm{RF}} = 
    \begin{dcases}
        (1 - \alpha) \left(1 + \sigma^{2} \prod_{l=1}^{\ell} \frac{\gamma_{l}-\alpha}{\gamma_{l}} + \sum_{l=1}^{\ell} \frac{\alpha}{\gamma_{l}-\alpha} \right) +  \left(\frac{\alpha}{1 - \alpha} + \sum_{l=1}^{\ell} \frac{\alpha}{\gamma_{l}-\alpha} \right) \eta^{2} , & \textrm{if}\ \alpha < \min\{1, \gamma_{\textrm{min}}\}
        \\
        \alpha \frac{1 - \gamma_{\textrm{min}}}{\alpha - \gamma_{\textrm{min}}} + \frac{\gamma_{\textrm{min}}}{\alpha - \gamma_{\textrm{min}}} \eta^{2}, & \textrm{if}\ \alpha > \gamma_{\textrm{min}}\ \textrm{and}\ \gamma_{\textrm{min}} < 1
        \\
        \frac{1}{\alpha - 1} \eta^{2}, & \textrm{if}\  \alpha > 1 \ \textrm{and}\ \gamma_{\textrm{min}} > 1.
    \end{dcases}
\end{align}
\end{widetext}

\begin{figure*}
    \centering
    \includegraphics[width=4in]{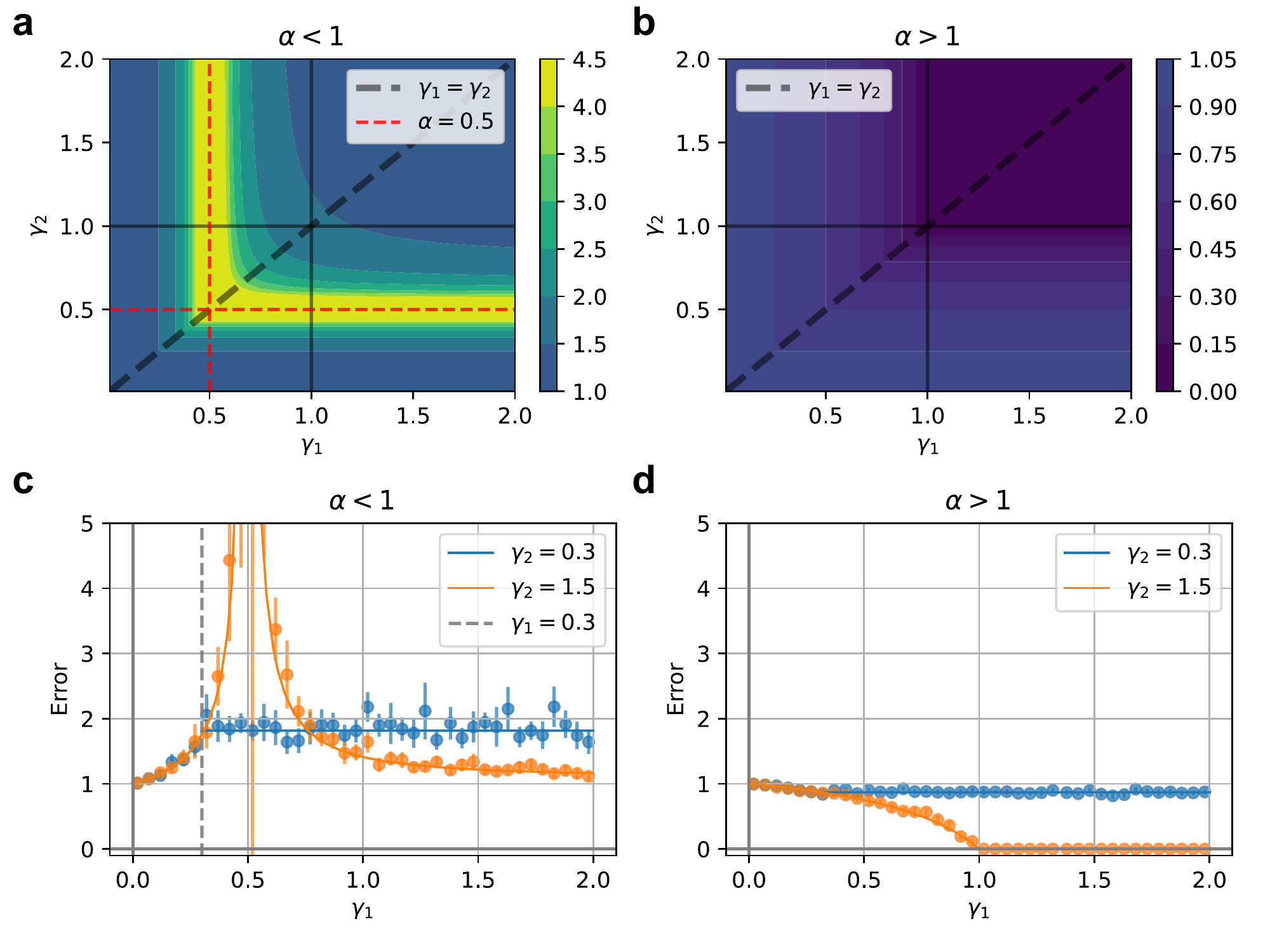}
    \caption{Double-descent in deep random feature models depends on the narrowest hidden layer. 
    \textbf{(a).} Contour plot in $(\gamma_1,\gamma_2)$-space of the theoretical error surface $\epsilon_{\textrm{RF}}$ \eqref{eqn:rf_learning_curve} for a deep RF model with two hidden layers and $\alpha=0.5$. For all panels, we set the input dimensionality $d = 100$, prior variance $\sigma^2 = 1$, and no label noise ($\eta = 0$). For details of our numerical methods, see Appendix \ref{app:numerical_methods}.
    \textbf{(b).} As in (a)., but with $\alpha=1.5$.
    \textbf{(c).} Horizontal cross sections of above (a). Theory curves are overlayed with experiment points, plotted with $\pm 2$ SE bars.
    \textbf{(d).} Horizontal cross sections of above (b).
    }
    \label{fig:rf_narrow}
\end{figure*}

We validate the accuracy of this RS result by comparing it against the result of an alternative semi-analytical approach. As shown in Appendix \ref{app:rf_posterior_expectations}, the zero-temperature posterior average in \eqref{eqn:generalization_error} can be computed for a fixed realization of the disorder. Even without explicitly evaluating the disorder average, this shows that the RF model should display the three phases indicated by the RS result \eqref{eqn:rf_learning_curve}, and confirms the prediction for in which of the phases the learning curve should depend on the prior variance $\sigma^2$ (see Appendix \ref{app:rf_posterior_expectations}). Importantly, the RF model learning curve \eqref{eqn:generalization_error} does not depend on the ordering of the hidden layer widths, which follows from the fact that the random Gaussian hidden layer weight matrices weakly commute \cite{akemann2013products,ipsen2014commutation}. For conceptual clarity, we therefore refer to the cases in which different hidden layers are the narrowest as a single phase. To quantitatively test the accuracy of the RS result, the disorder average can be evaluated numerically using sampling (see Appendix \ref{app:numerical_methods}). As shown in Figures \ref{fig:rf_learning_curve} and \ref{fig:rf_narrow}, we observe excellent agreement over a broad range of parameter values. These results are consistent with our expectation that the RS \emph{Ansatz} should yield accurate results for the RF models \cite{mezard1987spin,barbier2021strong,advani2020high,canatar2021spectral}.

While the LR model only exhibits double-descent behavior in the presence of label noise \eqref{eqn:lr_learning_curve}, the RF model can also exhibit double-descent behavior in the absence of label noise if any one of the hidden layers is narrower than the input dimension, i.e., $\gamma_{\textrm{min}} < 1$. This phenomenon occurs in a model-wise fashion at fixed data density: if one considers a decreasing sequence of widths $\gamma_{\textrm{min}}$ at fixed $\alpha$, $\epsilon_{\textrm{RF}}$ will diverge as $\gamma_{\textrm{min}} \downarrow \alpha$ (Figure \ref{fig:rf_learning_curve}a,c,e). Equivalently, this divergence can be observed in a sample-wise fashion at fixed width, with $\epsilon_{\textrm{RF}} \to \infty$ as $\alpha \to \gamma_{\textrm{min}}$. Moreover, as illustrated in Figure \ref{fig:rf_narrow}, it is determined by the width of the narrowest hidden layer. If one adds more bottleneck layers, then the expression for the generalization error in the regime $\alpha < \gamma_{\textrm{min}}$ will formally include more poles \eqref{eqn:rf_learning_curve}, but these poles will not be visible as one varies the size of the training set or the width of the narrowest bottleneck. 

Like the LR model, the RF model exhibits sample-wise double-descent behavior in the presence of label noise (Figure \ref{fig:rf_learning_curve}). However, if there is a bottleneck layer with width $\gamma_{\textrm{min}} < 1$, then the addition of label noise does not introduce additional divergences in $\epsilon_{\textrm{RF}}$ beyond that at $\alpha \to \gamma_{\textrm{min}}$; the pole at $\alpha = 1$ is visible only if $\gamma_{\textrm{min}} > 1$ (Figure \ref{fig:rf_learning_curve}b,d,e). This is clearly illustrated by comparing learning curves of two-layer ($\ell=1$) RF models with $\gamma_{\textrm{min}} = 1/2$ in the absence (Figure \ref{fig:rf_learning_curve}c) and presence (Figure \ref{fig:rf_learning_curve}d) of label noise: $\epsilon_{\textrm{RF}}$ increases with the addition of noise, but the only visible divergence is at $\alpha \to \gamma_{\textrm{min}}$. The presence of only a single divergence in $\epsilon_{\textrm{RF}}$ for two-layer models is consistent with work by d'Ascoli and colleagues on the phenomenology of double-descent in two-layer RF models trained with ridge regression \cite{dascoli2021triple}.

\subsection{Large-width behavior }

We now analyze the behavior of RF models at large widths. As $\gamma_{1},\ldots,\gamma_{\ell} \to \infty$, $\epsilon_{\textrm{RF}} \to \epsilon_{\textrm{LR}}$ for any fixed $\alpha$, $\sigma$, and $\eta$. We will refer to this simplification---the reduction of the linear curve of a deep linear model to that of simple linear regression---as the \emph{kernel limit} \cite{neal1996priors,matthews2018gaussian,lee2018deep,hron2020exact,jacot2018neural}. To obtain a more precise understanding of the behavior of the RF model near the kernel limit, we expand \eqref{eqn:rf_learning_curve} in the regime $\gamma_{1},\cdots,\gamma_{\ell} \gg 1$. If $\alpha > 1$, then we have $\epsilon_{\textrm{RF}} = \epsilon_{\textrm{LR}}$ in this regime. If $\alpha < 1$, we have
\begin{align} \label{eqn:rf_perturbation}
    \epsilon_{\textrm{RF}} &= \epsilon_{\textrm{LR}} + 
    [(1 - \alpha) (1 - \sigma^{2}) + \eta^{2}] \sum_{l=1}^{\ell} \frac{\alpha}{\gamma_{l}} + \mathcal{O}\left(\frac{\alpha^{2}}{\gamma^2}\right),
\end{align}
where $\mathcal{O}(\alpha^2/\gamma^2)$ denotes terms that include two or more factors of any combination of the layer widths. 

For an RF model of equal hidden layer widths $\gamma_{1} = \cdots = \gamma_{\ell} = \gamma$, the leading correction scales as $\ell \alpha/\gamma$. For this simple architecture, we can also study the scaling of higher-order corrections relatively easily. In the regime $\alpha < \min\{1,\gamma\}$, the learning curve \eqref{eqn:rf_learning_curve} can be written compactly as
\begin{align} \label{eqn:rf_equal_width_generalization}
    \frac{\epsilon_{\textrm{RF}} - \epsilon_{\textrm{LR}}}{1 - \alpha + \eta^2} &= \tilde{\sigma}^{2} \left[ \left(\frac{\gamma-\alpha}{\gamma}\right)^{\ell} - 1 \right] + \ell \frac{\alpha}{\gamma-\alpha} ,
\end{align}
where we have defined the re-scaled prior variance
\begin{align} \label{eqn:sigma_tilde}
    \tilde{\sigma}^2 \equiv \frac{ \sigma^{2} }{1 + \eta^{2}/(1-\alpha)}.
\end{align}
Then, for $\alpha/\gamma < 1$, we can read off the full series expansion using the binomial theorem and the geometric series: 
\begin{align} \label{eqn:rf_all_order_series}
    \frac{\epsilon_{\textrm{RF}} - \epsilon_{\textrm{LR}}}{1 - \alpha + \eta^2} &= \sum_{j=1}^{\infty} \left[ (-1)^{j} \tilde{\sigma}^{2}  \binom{\ell}{j} + \ell \right] \frac{\alpha^{j}}{\gamma^{j}},
\end{align}
where we note that $\binom{\ell}{j} = 0$ if $j > \ell$. Noting that $\binom{\ell}{j} = \mathcal{O}(\ell^{j})$ for $\ell \gg j$, we can see that the dominant term for large $\ell$ at each order in $\alpha / \gamma$ will scale with $\ell \alpha / \gamma$, up to around $\mathcal{O}(\alpha^{\ell}/\gamma^{\ell})$. Therefore, depth will have an important effect on how quickly the kernel limit is approached with varying width. At small $\ell$, the $j$-th order term will simply scale as $\ell (\alpha/\gamma)^{j}$ for all $j > \ell$, hence the effect of depth on the approach to the kernel limit can be neglected in this regime.

\subsection{Optimal width and depth}

\begin{figure*}
    \centering
    \includegraphics[width=4in]{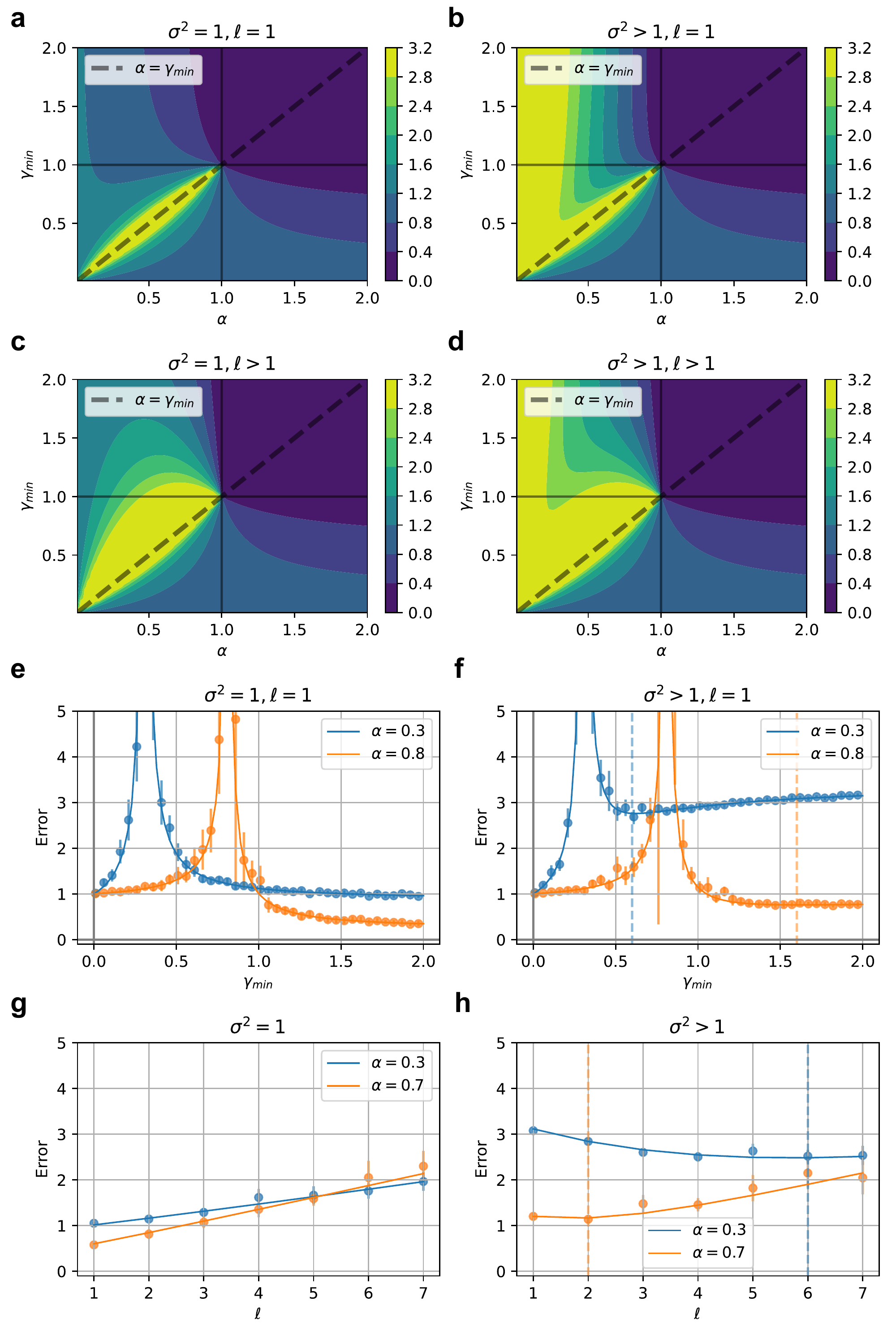}
    \caption{Optimal RF model architecture depends on target-prior mismatch.
    \textbf{(a).} Contour plot in $(\alpha,\gamma)$-space of the theoretical error surface $\epsilon_{\textrm{RF}}$ \eqref{eqn:rf_learning_curve} for a single-hidden-layer RF model with prior variance $\sigma^2 = 1$. For all panels, we have no label noise ($\eta = 0$) and set the input dimensionality $d = 100$. For details of our numerical methods, see Appendix \ref{app:numerical_methods}.
    \textbf{(b).} As in (a)., but for a single-hidden-layer RF model with higher prior variance ($\sigma^2 = 4$).
    \textbf{(c).} As in (a)., but for a deep RF model ($\ell = 5$) and prior variance $\sigma^2 = 1$.
    \textbf{(d).} As in (a)., but for a deep RF model ($\ell = 5$) and with higher prior variance ($\sigma^2 = 4$).
    \textbf{(e).} Vertical cross sections of above (a). Theory curves are overlayed with experiment points, plotted with $\pm 2$ SE bars.
    \textbf{(f).} Vertical cross sections of above (b). Optimal widths computed from equation \ref{eqn:rf_optimal_width} are marked with dashed vertical lines for each respective setting of $\alpha$.
    \textbf{(g).} Error across different depths for prior variance $\sigma^2 = 1$ and fixed width $\gamma = 1.5$
    \textbf{(h).} Error across different depths for prior variacne $\sigma^2 = 4$ and fixed width $\gamma = 1.5$. Optimal depths computed from equation $\ref{eq:optimal_depth}$ are marked with dashed vertical lines for each respective setting of $\alpha$.
    }
    \label{fig:rf_optimal}
\end{figure*}

With the formula \eqref{eqn:rf_learning_curve} for the generalization error in hand, we can determine the optimal hidden layer width for fixed depth, noise variance, and prior variance. We focus on the regime $\alpha < \min\{1,\gamma_{\textrm{min}}\}$, in which the generalization error always depends non-trivially on width. In Appendix \ref{app:rf_optimal_width}, we show that the optimal architecture for an RF model depends on the rescaled prior variance $\tilde{\sigma}^{2}$ defined in \eqref{eqn:sigma_tilde}. If $\tilde{\sigma} \leq 1$, then $\partial \epsilon_{\textrm{RF}}/\partial \gamma_{l} < 0$ for all $l$ and all widths in this regime, hence increasing width always improves generalization. Thus, in this regime, the best RF model is one that behaves identically to an LR model (Figure \ref{fig:rf_optimal}). If $\tilde{\sigma}^{2} > 1$, then $\epsilon_{\textrm{RF}}$ is minimized by taking all $\gamma_{1} =  \cdots = \gamma_{\ell} = \gamma_{\star}$ for
\begin{align} \label{eqn:rf_optimal_width}
    \gamma_{\star} = \frac{\tilde{\sigma}^{2/(\ell+1)}}{\tilde{\sigma}^{2/(\ell+1)} - 1} \alpha .
\end{align}
We note that the leading term in the perturbative expansion \eqref{eqn:rf_perturbation} would predict that generalization performance improves with increasing width if $\tilde{\sigma} < 1$, is invariant under changes of width if $\tilde{\sigma} = 1$, and degrades with increasing width if $\tilde{\sigma} > 1$. Thus, in this case the leading-order perturbative correction captures some, but not all, of the effect of width.

In the absence of noise, this yields a simple qualitative picture in which the optimal width is related to the mismatch between the scale of the prior and the target weight vector: if $\mathbb{E}_{\mathcal{W}} \Vert \mathbf{w} \Vert^2 = \sigma^{2} d \leq d = \Vert \mathbf{w}_{\ast} \Vert^2$, then wider networks are always better, while otherwise one can obtain improved generalization performance by using an RF model rather than an LR model. This occurs because of the trade-off between the terms with linear and exponential dependence on depth in \eqref{eqn:rf_learning_curve}. Label noise has the effect of increasing the effective scale of the target in a way that depends on the data density $\alpha$: as $\alpha \uparrow 1$, wider models are always better in the presence of label noise. This behavior is illustrated in Figure \ref{fig:rf_optimal}.

Similarly, one can also optimize the depth for fixed width, noise variance, and prior variance. To do so, it is convenient to assume that all layers are of the same width $\gamma_{1} = \cdots = \gamma_{\ell} = \gamma$, which allows us to analytically continue $\epsilon_{\textrm{RF}}$ as a function of the depth $\ell$. Again specializing to the regime $\alpha < \min\{1,\gamma\}$, the generalization error is given by \eqref{eqn:rf_equal_width_generalization}. It is then easy to see that the LR model learning curve \eqref{eqn:lr_learning_curve} is recovered upon $\ell \downarrow 0$. In Appendix \ref{app:rf_optimal_depth}, we show that, if $\tilde{\sigma} \leq 1$, $\epsilon_{\textrm{RF}}$ is a monotonically increasing function of $\ell$, hence shallower RF models always generalize better. This is consistent with our result above for optimal width, because taking $\gamma_{l} \to \infty$ for some $l$ effectively reduces the depth of the RF model by one, by eliminating that layer's contribution to \eqref{eqn:rf_equal_width_generalization}. If $\tilde{\sigma} > 1$, then the optimal depth is given by 
\begin{align} \label{eq:optimal_depth}
    \ell_{\star} = \begin{dcases} 
        j\ \textrm{or}\ j - 1, & \textrm{if}\ \frac{\log(\tilde{\sigma}^2)}{\log[\gamma/(\gamma-\alpha)]} = j \in \mathbb{N}_{>0}
        \\
        \left\lfloor \frac{\log(\tilde{\sigma}^2)}{\log[\gamma/(\gamma-\alpha)]} \right\rfloor, & \textrm{otherwise}.
    \end{dcases}
\end{align}
In the former condition, taking $\ell = j$ or $\ell = j-1$ will yield identical generalization error. Moreover, for the condition $\log(\tilde{\sigma}^{2}) / \log[\gamma/(\gamma-\alpha)] \in \mathbb{N}_{>0}$ to hold, the network width must be of the form
\begin{align}
    \gamma = \frac{\tilde{\sigma}^{2/j}}{\tilde{\sigma}^{2/j} - 1} \alpha ,
\end{align}
which is consistent with the result for optimal width at fixed depth given in \eqref{eqn:rf_optimal_width}. Therefore, much like we found in our analysis of optimal width, the optimal depth of an RF model is related to the match between the scale of the prior and of the target. This behavior is illustrated in Figure \ref{fig:rf_optimal}.

\section{Learning curves for the NN model}

\subsection{Learning curve and double-descent behavior} 

For the NN model, we do not obtain a simple closed-form solution for the RS learning curve at general depth. As shown in Appendix \ref{app:nn_rs_saddle_point}, we find that the solution is of the form
\begin{align} \label{eqn:nn_learning_curve}
    \epsilon_{\textrm{NN}} = \epsilon_{\textrm{LR}} + 
    \begin{dcases} 
    z - \sigma^2 (1-\alpha), &\textrm{if}\ \alpha < 1
    \\
    0,  &\textrm{if}\ \alpha > 1,
    \end{dcases}
\end{align}
where $z = z(\alpha, \sigma^{2}, \eta^{2},\gamma_{1},\ldots,\gamma_{\ell})$ is a non-negative real root of the polynomial 
\begin{align} \label{eqn:z_polynomial}
    z^{\ell+1} = \sigma^{2} (1 - \alpha) \prod_{l=1}^{\ell} \left[\frac{\gamma_{l} - \alpha}{\gamma_{l}} z + \frac{\alpha (1 - \alpha + \eta^{2})}{\gamma_{l}} \right].
\end{align}
We defer more detailed discussion of which root should be selected to Appendix \ref{app:nn_rs_saddle_point}, where we show that one required condition on the solution is that
\begin{align}
    (\gamma_{l} - \alpha) z + \alpha (1 - \alpha + \eta^{2}) > 0 
\end{align}
for all $l$. For a network with a single hidden layer ($\ell = 1$), \eqref{eqn:z_polynomial} is quadratic, and we can easily obtain
\begin{align}
    \frac{z}{1 - \alpha + \eta^2 } &= \frac{\tilde{\sigma}^{2} (\gamma_{1} - \alpha) + \sqrt{\tilde{\sigma}^{4} (\gamma_{1} - \alpha)^{2} + 4 \alpha \gamma_{1} \tilde{\sigma}^{2} }}{2 \gamma_{1}},
\end{align}
where $\tilde{\sigma}^{2}$ is defined as in \eqref{eqn:sigma_tilde}. 

\begin{figure*}[t]
    \centering
    \includegraphics[width=4in]{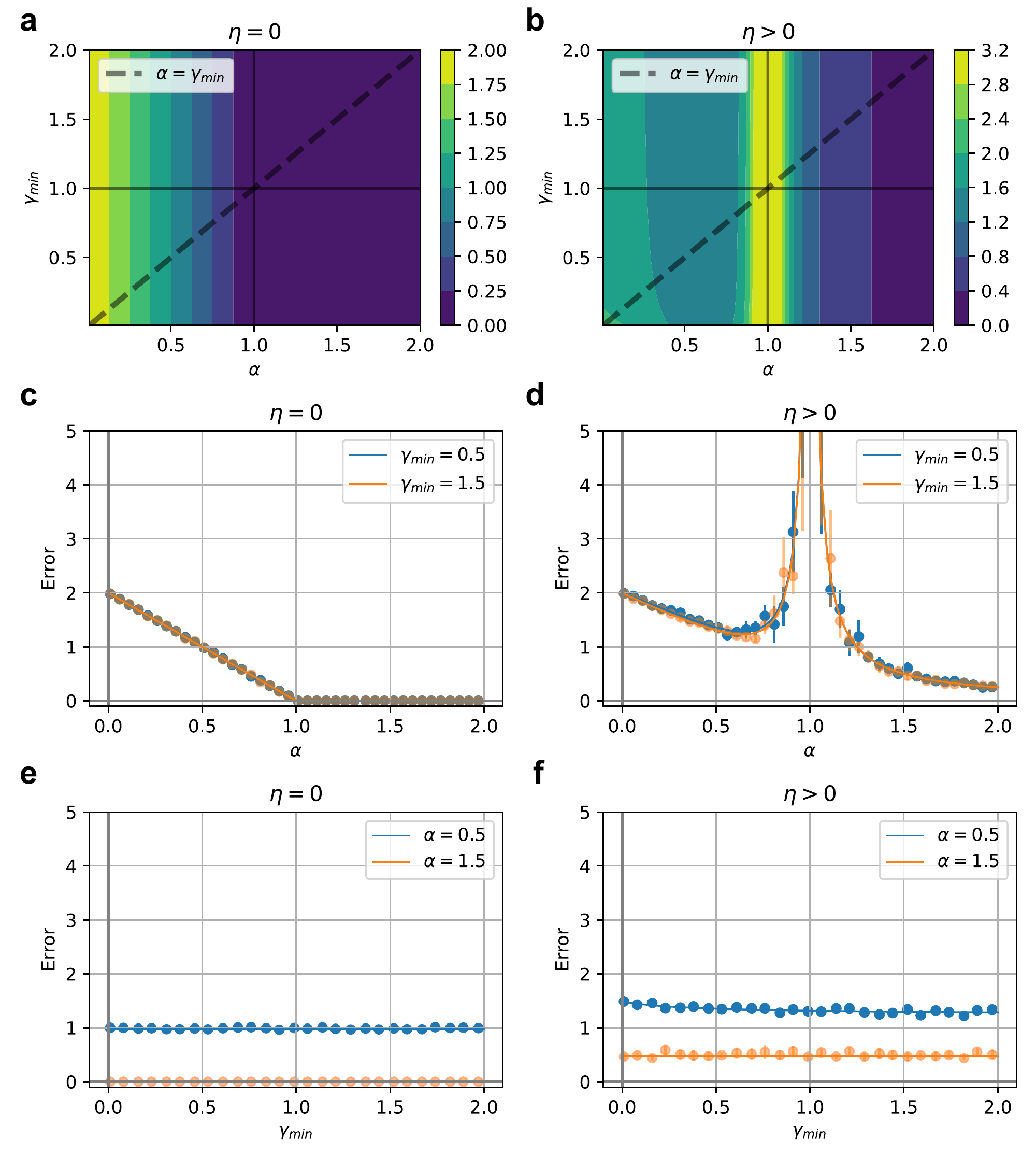}
    \caption{Sample-wise double-descent in deep Bayesian neural networks.
    \textbf{(a).} Contour plot in $(\alpha,\gamma)$-space of the theoretical error surface $\epsilon_{\textrm{NN}}$ \eqref{eqn:nn_learning_curve} for a two-layer NN model in the absence of label noise ($\eta = 0$). For all panels, we set the input dimensionality $d = 100$ and prior variance $\sigma^2 = 1$. For details of our numerical methods, see Appendix \ref{app:numerical_methods}.
    \textbf{(b).} As in (a)., but in the presence of label noise ($\eta = 0.5$). 
    \textbf{(c).} Horizontal cross sections of above (a). Theory curves are overlayed with experiment points, plotted with $\pm 2$ SE bars. 
    \textbf{(d).} Horizontal cross sections of above (b).
    \textbf{(e).} Vertical cross sections of above (a).
    \textbf{(f).} Vertical cross sections of above (b).
    }
    \label{fig:nn_learning_curve}
\end{figure*}

The special case of \eqref{eqn:nn_learning_curve} for networks with hidden layers of equal widths follows from results obtained through a rather different approach in a recent study by Li and Sompolinsky \cite{li2021statistical}. Concretely, they use an iterative saddle-point argument to approximate the posterior expectation in \eqref{eqn:generalization_error} for fixed data, and then apply that result to a random Gaussian covariate model under what amounts to the assumption that the quantity $\mathbf{y}^{\top} (\mathbf{X} \mathbf{X}^{\top})^{-1} \mathbf{y}$ concentrates rapidly. In Appendix \ref{app:nn_posterior_expectations}, we provide a detailed discussion of the mapping between the polynomial condition in terms of which their result is expressed and the RS condition \eqref{eqn:z_polynomial}. In Appendix \ref{app:nn_posterior_expectations}, we also use a finite-size fixed-data approach derived from our previous work \cite{zv2021scale} to show that the learning curve should be of the form \eqref{eqn:nn_learning_curve}. Concretely, this approach gives an expression for $z$ as the thermodynamic limit of a dataset average of a ratio of prior averages, with the remaining components of the learning curve exactly matching the RS prediction. Taken together, these result suggests that the RS prediction for the learning curve correctly captures at least the coarse behavior of generalization in NNs.

To further probe whether the RS prediction is quantitatively accurate, we evaluate the finite-size data average numerically. As shown in Figures \ref{fig:nn_learning_curve} and \ref{fig:nn_narrow}, and in supplemental figures provided in Appendix \ref{app:numerical_methods}, we observe good agreement for two-layer networks. To probe the accuracy of the RS prediction for deeper networks, we solve the polynomial \eqref{eqn:z_polynomial} numerically. As shown in Figures \ref{fig:nn_learning_curve} and \ref{fig:nn_narrow}, we again observe good agreement. Therefore, both alternative heuristic analytical approaches and numerical results are consistent with the RS learning curve, suggesting that it provides a reasonably accurate picture of generalization in deep NNs.

\begin{figure*}
    \centering
    \includegraphics[width=4in]{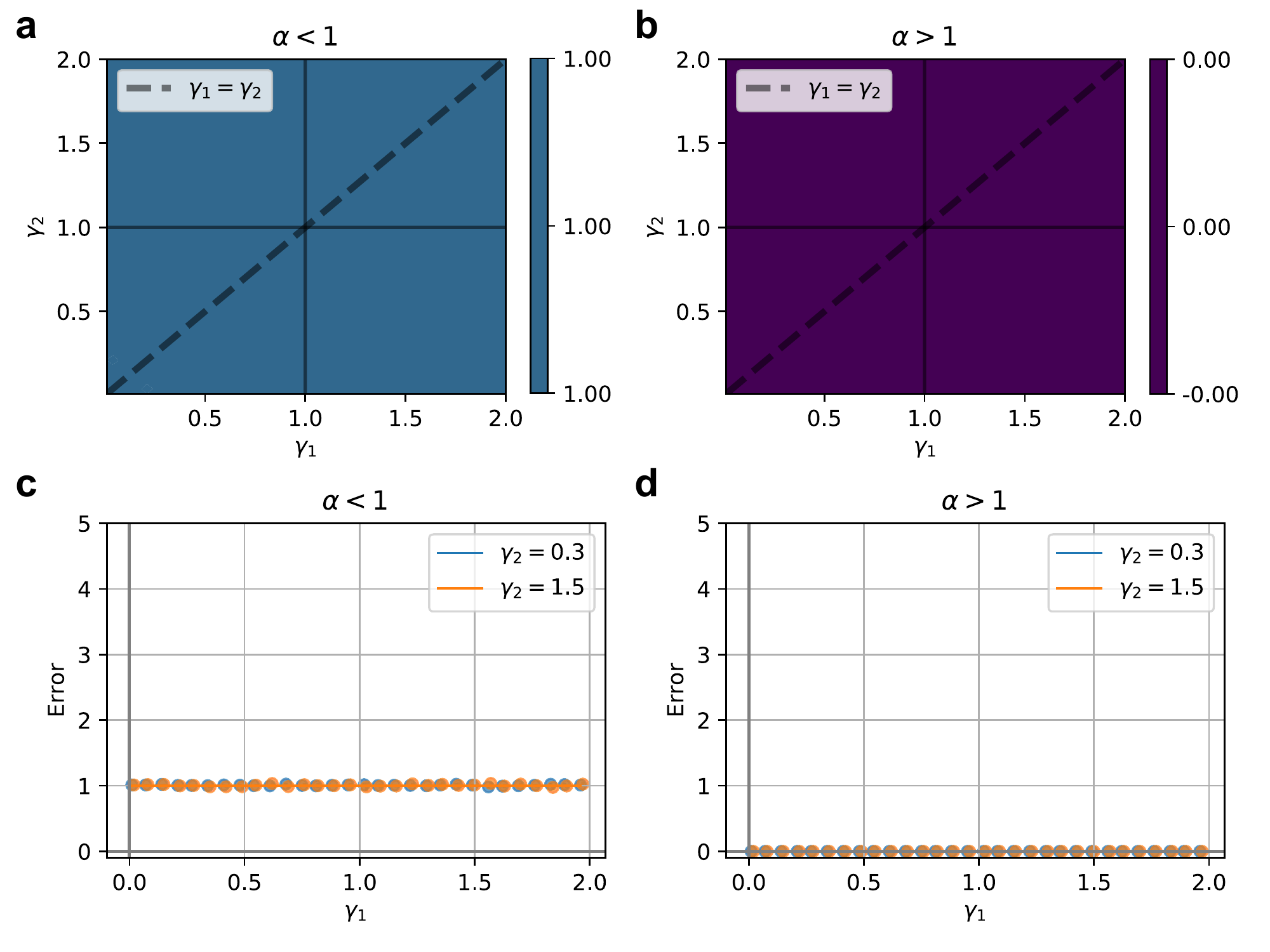}
    \caption{Bottleneck layers do not induce model-wise double-descent in deep NNs. 
    \textbf{(a).} Contour plot in $(\gamma_1,\gamma_2)$-space of the theoretical error surface $\epsilon_{\textrm{NN}}$ \eqref{eqn:nn_learning_curve} for a deep Bayesian NN model with two hidden layers and $\alpha=0.5$. For all panels, we set the input dimensionality $d = 100$, prior variance $\sigma^2 = 1$, and no label noise ($\eta = 0$). For details of our numerical methods, see Appendix \ref{app:numerical_methods}.
    \textbf{(b).} As in (a)., but with $\alpha = 1.5$. 
    \textbf{(c).} Horizontal cross sections of above (a). Theory curves are overlayed with experiment points, plotted with $\pm 2$ SE bars. 
    \textbf{(d).} Horizontal cross sections of above (b).
    }
    \label{fig:nn_narrow}
\end{figure*}

Like the previously-studied models, we see that label noise can induce sample-wise double-descent, with $\epsilon_{\textrm{NN}} \to \infty$ as $\alpha \to 1$ (Figure \ref{fig:nn_learning_curve}). However, unlike for the RF model, having relatively narrow hidden layers does not introduce the possibility of divergences other than at $\alpha = 1$, as $z$ should remain bounded. This is illustrated in Figure \ref{fig:nn_narrow}, where we repeat the analysis of Figure \ref{fig:rf_narrow}, but do not observe similar model-wise divergences. Moreover, this means that the NN model does not display sample-wise divergences in the absence of label noise. Therefore, training the hidden layers affords the advantage of avoiding the possible model- and sample-wise divergences that can arise in RF models with narrow bottlenecks. This sharp contrast makes sense, since in the RF model the presence of layers width $\gamma_{l} < 1$ introduces a true bottleneck, while in the NN model one could in principle find a solution where, in all layers except the first, exactly one weight is nonzero, and the model essentially reduces to shallow linear regression. The existence of this solution reflects the fact that, from the standpoint of expressivity, NN models should be able to perform as well as LR models, and differences in performance reflect the behavior of the inference algorithm \cite{saxe2013exact}. Indeed, if $\sigma = 1$ and $\eta = 0$, we have the solution $z = 1-\alpha$, and $\epsilon_{\textrm{NN}} = \epsilon_{\textrm{LR}}$. Therefore, in this special case, the RS result predicts that depth has no effect on generalization performance. This behavior is clearly illustrated by Figure \ref{fig:nn_narrow}, where the generalization error of a three-layer NN remains constant as the widths of the two hidden layers are varied. Even at non-zero noise levels, Figure \ref{fig:nn_learning_curve} illustrates that width has a relatively minimal effect of generalization performance when $\sigma = 1$. 

\subsection{Large-width behavior}

Beyond the special cases mentioned above, we observe that, in the limit $\gamma_{1},\ldots,\gamma_{\ell} \to \infty$, we have the solution $z=\sigma^{2}(1-\alpha)$ for any fixed $\alpha$, $\sigma$, and $\eta$. Therefore, as we found for the RF model, the NN model's generalization performance reduces to that of the shallow LR model in this large-width limit: $\epsilon_{\textrm{NN}} \to \epsilon_{\textrm{LR}}$. In the regime $\alpha < 1$, $\gamma_{1},\ldots,\gamma_{\ell} \gg 1$, we can obtain a perturbative solution for the learning curve (see Appendix \ref{app:rs_saddle_point}), which is given as
\begin{align} \label{eqn:nn_perturbation}
    \epsilon_{\textrm{NN}} &= \epsilon_{\textrm{LR}}
    + [(1 - \alpha) (1-\sigma^{2}) + \eta^2 ] \sum_{l=1}^{\ell} \frac{\alpha}{\gamma_{l}} + \mathcal{O}\left(\frac{\alpha^{2}}{\gamma^2}\right)
\end{align}
to leading order. This result can be compared to the leading-order perturbative computation of the zero-temperature learning curve for fixed data in our previous work \cite{zv2021asymptotics}. As shown in Appendix \ref{app:perturbative_comparison}, averaging the result of \cite{zv2021asymptotics} over data recovers the $\mathcal{O}(\alpha/\gamma)$ term resulting from the replica method computation. This suggests that the RS prediction for the NN model learning curve is accurate at large widths. Heuristically, this makes sense because the concavity of the log-posterior is restored in the limit $\gamma_{1},\ldots,\gamma_{\ell} \to \infty$.

This limiting result has several interesting features. First, paralleling our analysis of the RF model at large widths, the closeness of the NN model's learning curve to that of simple linear regression is determined by a combination of depth, dataset size and width. Second, not only do the RS learning curves for NN and RF models agree at infinite width, but the leading order corrections agree (i.e., the term that is linear in $\alpha/\gamma_{l}$; see \eqref{eqn:rf_perturbation}). Thus, if one tracked only the generalization error, one could not differentiate between training only the readout layer and training all of the layers simply by considering the leading order perturbative correction. One could of course distinguish between these two models by considering leading-order corrections to observables that explicitly measure task-relevant feature learning in early hidden layers, such as the kernels considered in our previous work \cite{zv2021asymptotics}.

\subsection{Generalization gap between RF and NN models}

\begin{figure}[tb]
    \centering
    \includegraphics[width=3in]{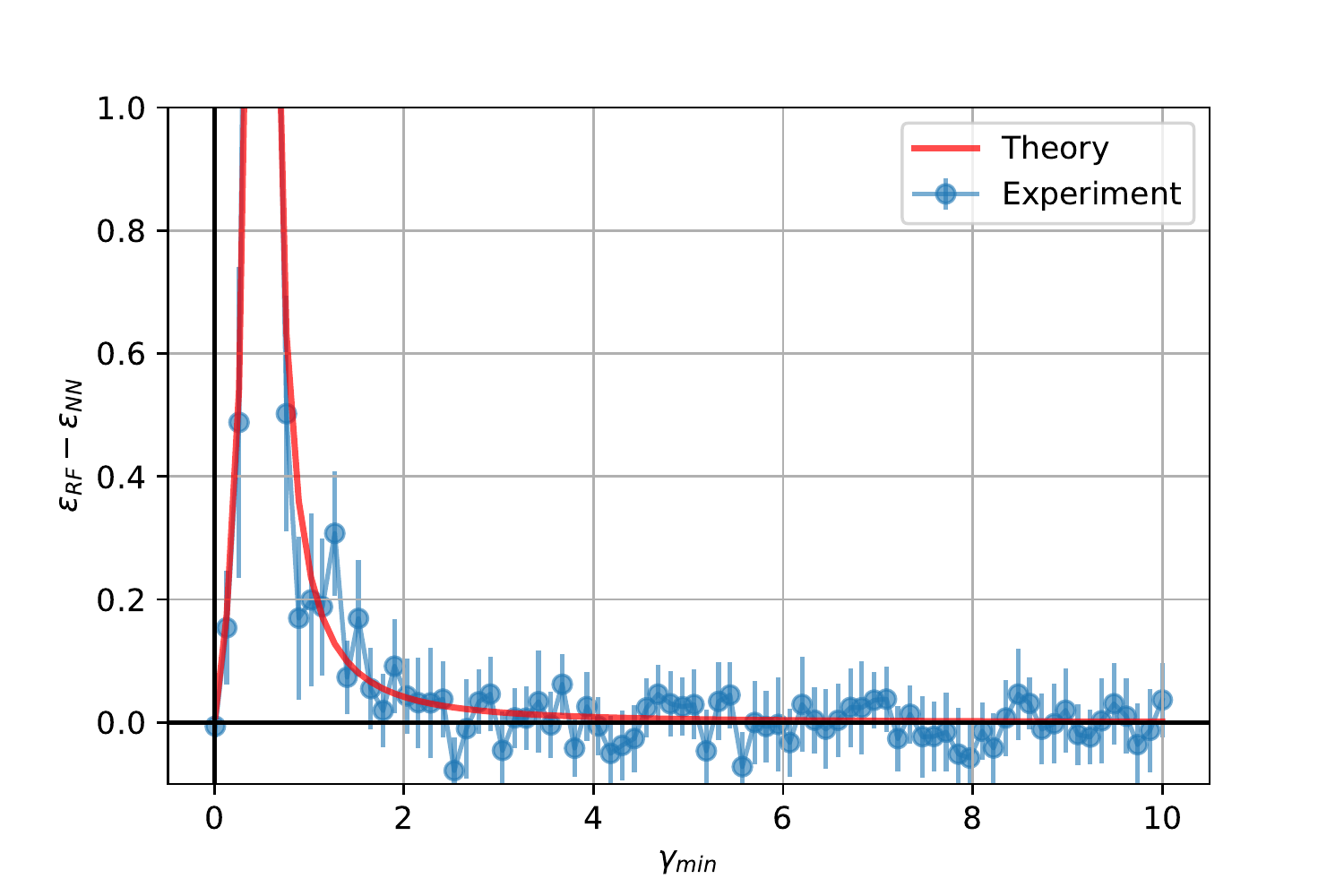}
    \caption{The generalization gap between RF and NN models approaches zero with increasing width. The difference $\epsilon_{\textrm{RF}} - \epsilon_{\textrm{NN}}$ remains positive or consistent with zero within standard error throughout, highlighting the advantage in training all layers. See Appendix \ref{app:numerical_methods} for details of our numerical methods.}
    \label{fig:rf_nn_gap}
\end{figure}

To distinguish between RF and NN models based on generalization performance, one must therefore go to higher order in perturbation theory. For convenience and clarity, we specialize to the case of networks with equal hidden layer widths $\gamma_{1}=\gamma_{2}=\cdots=\gamma_{\ell} = \gamma$. Then, we find that
\begin{align}
    \frac{\epsilon_{\textrm{NN}} - \epsilon_{\textrm{LR}}}{1-\alpha+\eta^2} &= (1-\tilde{\sigma}^2) \frac{\ell \alpha}{\gamma} \nonumber\\&\quad + \left(\frac{\ell(\ell-1) \tilde{\sigma}^2}{2} - \frac{\ell(\ell+1) }{2 \tilde{\sigma}^2} + \ell \right) \frac{\alpha^2}{\gamma^2} \nonumber\\&\quad + \mathcal{O}\left(\frac{\alpha^3}{\gamma^3}\right),
\end{align}
where $\tilde{\sigma}$ is defined as in \eqref{eqn:sigma_tilde}. In contrast, by truncating \eqref{eqn:rf_all_order_series} to this order, we can see the corresponding RF model has generalization error
\begin{align}
    \frac{\epsilon_{\textrm{RF}} - \epsilon_{\textrm{LR}}}{1-\alpha+\eta^2} &= (1-\tilde{\sigma}^2) \frac{\ell \alpha}{\gamma} \nonumber\\&\quad + \left(\frac{\ell (\ell-1) \tilde{\sigma}^2}{2} + \ell\right) \frac{\alpha^2}{\gamma^2} \nonumber\\&\quad + \mathcal{O}\left(\frac{\alpha^{3}}{\gamma^3}\right).
\end{align}
Therefore, the next-to-leading order correction can distinguish between RF and NN models. Moreover, the gap in the generalization performance of the two models is, to the given order, 
\begin{align}
    \frac{\epsilon_{\textrm{RF}} - \epsilon_{\textrm{NN}}}{1-\alpha+\eta^2} = \frac{\ell(\ell+1) }{2 \tilde{\sigma}^2} \frac{\alpha^2}{\gamma^2} + \mathcal{O}\left(\frac{\alpha^{3}}{\gamma^3}\right). 
\end{align}
The coefficient of the leading term is always positive, hence at very large widths training both layers should produce a small benefit relative to simply training the readout. In the two-layer case, one can use the closed-form solution for the RS generalization error to show that the generalization gap $\epsilon_{\textrm{RF}} - \epsilon_{\textrm{NN}}$ is strictly positive, except at vanishing load or in the limit $\gamma_{1} \to \infty$ (see Appendix \ref{app:generalization_gap}). These results suggest that training all layers of a deep linear network can yield improved generalization relative to training only the last layer, even if the widths are large enough such that the RF model does not display double-descent in the absence of noise. See Figure \ref{fig:rf_nn_gap} for an illustration of this behavior.

\subsection{Optimal width and depth}

The leading correction term in \eqref{eqn:nn_perturbation} predicts that generalization error always decreases with increasing width (respectively increases with increasing depth) if $\tilde{\sigma} < 1$, is constant if $\tilde{\sigma} = 1$, and always increases (respectively decreases) if $\tilde{\sigma} > 1$. As shown in Appendix \ref{app:perturbative_comparison}, this condition is the dataset-averaged version of the fixed-data condition noted by Li and Sompolinsky \cite{li2021statistical} and in our previous perturbative work \cite{zv2021asymptotics}. In Appendix \ref{app:nn_optimal_width}, we show in detail that this condition captures the behavior of the full RS generalization error of an NN with one hidden layer; for other depths it follows from an argument based on implicit differentiation given by Li and Sompolinsky \cite{li2021statistical}. Therefore, like in our study of the RF model, the optimal width of an NN depends on the match between the scales of the prior and of the target. However, unlike for an RF model, the RS result suggests that the optimal generalization performance for an NN is obtained either by taking $\gamma_{l} \to \infty$ or by taking $\gamma_{l} \downarrow 0$, behavior which is fully predicted by the leading perturbative correction. This behavior is illustrated in Figure \ref{fig:nn_optimal}.

\section{Discussion and conclusions}

\subsection{Summary of results}

\begin{figure*}
    \centering
    \includegraphics[width=4in]{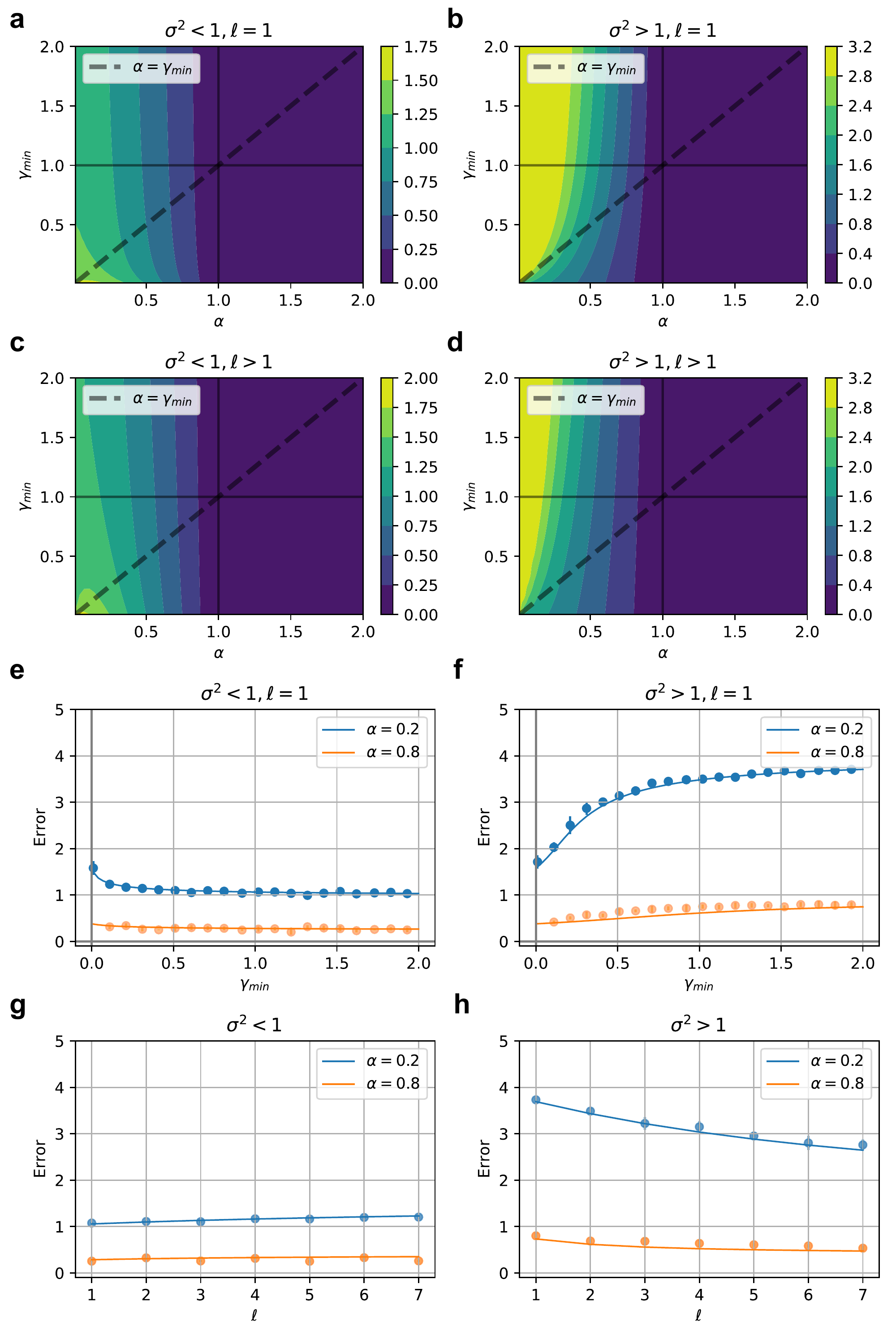}
    \caption{Optimal NN model architecture depends on target-prior mismatch.
    \textbf{(a).} Contour plot in $(\alpha,\gamma)$-space of the theoretical error surface $\epsilon_{\textrm{NN}}$ \eqref{eqn:nn_learning_curve} for a single-hidden-layer NN model with prior variance $\sigma^2 = 1/4$. For all panels, we have no label noise ($\eta = 0$) and set the input dimensionality $d = 100$. For details of our numerical methods, see Appendix \ref{app:numerical_methods}.
    \textbf{(b).} As in (a)., but for a single-hidden-layer NN model with higher prior variance ($\sigma^2 = 4$).
    \textbf{(c).} As in (a)., but for a deep NN model ($\ell = 5$) and prior variance $\sigma^2 = 1/4$.
    \textbf{(d).} As in (a)., but for a deep NN model ($\ell = 5$) and with higher prior variance ($\sigma^2 = 4$).
    \textbf{(e).} Vertical cross sections of above (a). Theory curves are overlayed with experiment points, plotted with $\pm 2$ SE bars.
    \textbf{(f).} Vertical cross sections of above (b).
    \textbf{(g).} Error across different depths for prior variance $\sigma^2 = 1/4$ and fixed width $\gamma = 1.5$
    \textbf{(h).} Error across different depths for prior variance $\sigma^2 = 4$ and fixed width $\gamma = 1.5$.
    }
    \label{fig:nn_optimal}
\end{figure*}

In this work, we studied the statistical mechanics of inference in deep Bayesian linear models. We characterized the learning curves of deep linear random feature models and deep linear neural networks for isotropic Gaussian covariates, using a combination of the replica trick and replica-free methods. Our primary results for how deep Bayesian linear models with random and learned features differ or resemble may be summarized as follows: 
\begin{itemize}
    \item  In the presence of label noise, both RF and NN models display sample-wise double-descent (Figures \ref{fig:rf_learning_curve} and \ref{fig:nn_learning_curve}). For RF models, the presence of a bottleneck layer with width less than the input dimension induces model-wise double-descent at fixed dataset size and sample-wise double descent at fixed width (Figures \ref{fig:rf_learning_curve} and \ref{fig:rf_narrow}), while bottlenecks do not affect the double-descent behavior of NN models (Figures \ref{fig:nn_learning_curve} and \ref{fig:nn_narrow}). In particular, NN models do not display model-wise double-descent, and do not display sample-wise double-descent in the absence of label noise.
    
    \item 
    For both RF and NN models, the effect of width on generalization depends on the match between the prior variance and the true scale of the targets, with wider networks yielding better generalization when the prior variance is less than the average target scale (Figures \ref{fig:rf_optimal} and \ref{fig:nn_optimal}). For NN models, taking the network to be as wide or as narrow as possible is always optimal. In contrast, when the prior variance is greater than the average target scale, there is a particular width that yields optimal generalization in RF models. 
    
    \item 
    Similarly, the optimal depth for both models depends on prior-target mismatch. Paralleling the case of optimal width, deeper models always perform worse when the prior variance is less than the average target scale (Figures \ref{fig:rf_optimal} and \ref{fig:nn_optimal}). When prior variance is greater than the average target scale, shallower models perform better. In this regime, as in the case of optimal width, there is a particular depth that yields optimal RF model generalization for fixed width, prior variance, and data density. 
    
    \item 
    Both RF and NN models display kernel-limit behavior---i.e., their learning curves reduce to those of shallow linear regression---when the depth and dataset size are small relative to the hidden layer width. Moreover, for both classes of deep models, the $\mathcal{O}(\ell\alpha/\gamma)$  perturbative correction captures much of the gross qualitative behavior of the learning curve as a function of prior variance, width, and depth. 
    
    \item 
    The learning curves of wide RF and NN models coincide not only in the limit $\ell\alpha/\gamma \downarrow 0$, but have identical leading-order corrections in $\ell\alpha/\gamma$. Training all layers improves generalization relative to training only the readout, but this gap is an $\mathcal{O}(\ell^2\alpha^2/\gamma^2)$ effect (Figure \ref{fig:rf_nn_gap}). 
    
\end{itemize}

\subsection{Prior work}

As noted above, our results for deep linear neural networks partially overlap with those obtained previously by Li and Sompolinsky \cite{li2021statistical}. Specifically, the RS learning curve for networks of equal hidden layer width agrees with the result they obtained through an alternative heuristic, and, as a result, their criteria for when generalization improves or degrades with width and depth coincide with those obtained here.  However, they did not analyze in detail how the kernel limit is approached as $\ell \alpha/\gamma \downarrow 0$, and did not consider random feature models. Our results therefore complement their study by providing a more granular picture of how generalization performance for random datasets depends on model architecture. Moreover, the agreement between their approximations, our RS results, and our numerical simulations is consistent with the conjecture that the RS learning curve is reasonably accurate.

Double-descent phenomena have recently garnered significant interest in deep learning \cite{belkin2019reconciling,krogh1992generalization,nakkiran2019more,nakkiran2021deep,dascoli2020double,dascoli2021triple,adlam2020understanding,canatar2021spectral,geiger2020scaling,mei2019generalization}. In high-dimensional random feature models like those considered in \S\ref{sec:rf_model}, divergences in the generalization error can arise through interactions between randomness in the features and randomness in the training data \cite{mei2019generalization,dascoli2020double,dascoli2021triple,adlam2020neural,adlam2020understanding}. Moreover, as noted in our discussion of simple linear regression models in \S\ref{sec:lr_model}, divergences can arise in models without additional feature randomness---including kernel regressors with deterministic nonlinear features---due to overfitting of noisy labels \cite{krogh1992generalization,nakkiran2019more,hastie2019surprises,canatar2021spectral,dascoli2020double,adlam2020neural,adlam2020understanding,dascoli2021triple}. Disentangling the causes of non-monotonicitic generalization performance observed in experimental settings for realistic data models remains an interesting subject for further study \cite{belkin2019reconciling,krogh1992generalization,nakkiran2019more,nakkiran2021deep,dascoli2020double,dascoli2021triple,adlam2020understanding,canatar2021spectral,geiger2020scaling,mei2019generalization}.

The statistical mechanics of inference in shallow linear models with more general priors and likelihoods was investigated in detail by Advani and Ganguli \cite{advani2016statistical}, who showed a correspondence between the performance of Bayesian MMSE inference and a class of algorithms known as M-estimators. The effect of prior mismatch on the performance of the shallow MMSE estimator has also been considered in recent rigorous work by Barbier \emph{et al.} \cite{barbier2021performance}. However, neither of these works considered the effect of depth on inference. 

Here, we considered a proportional asymptotic limit in which the input dimension $d$, dataset size $p$, and hidden layer widths $n_{\ell}$ tend jointly to infinity with fixed limiting ratios $\alpha = p/d = \mathcal{O}(1)$ and $\gamma_{l} = n_{l}/d = \mathcal{O}(1)$ and fixed depth $\ell$. Moreover, we have only considered networks with scalar output. In this setting, kernel-machine behavior---i.e., approximate reduction of the learning curves of deep models to those of simple linear regression \cite{neal1996priors,williams1997computing,lee2018deep,matthews2018gaussian,hron2020exact}---emerges when the ratio $\ell \alpha/\gamma_{l}$ is vanishingly small. This is consistent with our observations in prior work \cite{zv2021asymptotics,zv2021scale}, and with those of works that considered large depths but fixed dataset size \cite{roberts2022principles} or large dataset size but fixed depth \cite{li2021statistical,naveh2021self}. Given the large scale of contemporary regression and classification datasets (e.g., \cite{russakovsky2015imagenet}), careful consideration of limits in which the output dimension and dataset size are not vanishingly small relative to hidden layer width warrants further study. 

This regime has thus far proven challenging to access perturbatively, as large deviations from the kernel limit may emerge \cite{zv2021asymptotics,li2021statistical,aitchison2020bigger,zv2021scale,naveh2021self}. Existing fixed-data approaches to regimes in which either the dataset size \cite{naveh2021self,li2021statistical,zv2021scale} or the output dimension \cite{aitchison2020bigger,zv2021scale} is not negligible relative to hidden layer width rely on saddle-point approximations that may break down when both of these parameters are large. New approaches will therefore be required to study networks in this limit non-perturbatively. With such results in hand, it will be interesting to test whether existing perturbative predictions do in fact capture qualitative features of how generalization depends on network architecture and other hyperparameters. For the simple models considered here, we found that small-sample-size perturbation theory does in fact yield largely correct predictions for when wider networks generalize better, even at large sample size.

\subsection{Outlook}

We conclude by noting that our work has several important limitations, which will be interesting to address in future work. First, our approach is highly specialized to deep linear networks, and would not extend easily to nonlinear models. Though the utility of linear networks as a model system for studying the effect of depth on inference has been clearly established  \cite{saxe2013exact,fukumizu1998effect,advani2020high,zv2021asymptotics,li2021statistical}, rigorous characterization of the effect of nonlinearity on inference in deep Bayesian neural networks remains a largely open problem \cite{zv2021asymptotics,zv2021exact,zv2021activation,naveh2021predicting,naveh2021self,li2021statistical,roberts2022principles,aitchison2020bigger}. Second, we have assumed that the covariates are drawn from an isotropic Gaussian distribution. Though this is a standard generative model in theoretical studies of inference \cite{krogh1992generalization,hastie2019surprises,barbier2021performance,advani2016statistical,nakkiran2019more,advani2020high}, it is undoubtedly not reflective of real-world data. Extending results of this form to more realistic generative models will be an interesting objective for future work \cite{canatar2021spectral,loureiro2021learning,gold2020manifold}. We remark that some of the fixed-data results of Appendices \ref{app:rf_posterior_expectations} and \ref{app:nn_posterior_expectations} would extend immediately to anisotropic and non-Gaussian data provided that the requisite invertibility conditions hold. While BNNs are finding practical applications in physics and elsewhere \cite{cranmer2021bayesian}, another important direction for future work will be to develop a rigorous theoretical understanding of how results on the generalization performance of BNNs, like those obtained here, relate to the generalization performance of networks trained with stochastic gradient-based algorithms, a link that remains incompletely understood \cite{wilson2020bayesian,izmailov2021bayesian,mingard2021almost,he2021ensembles}. Finally, our replica theory approach is of course non-rigorous. For the RF model, we do not expect replica symmetry to be broken, and conjecture that our results might be rigorously justifiable \cite{mezard1987spin,barbier2021strong,canatar2021spectral,engel2001statistical,akemann2013products,ipsen2014commutation}. Moreover, our replica-free analytical approaches and numerical experiments suggest that our RS results for NNs are at the very least a reasonable approximation for their true generalization performance. With that in mind, careful exploration of the possibility of replica symmetry breaking will be an interesting topic for further investigation. 

\emph{Note added.} Following the appearance of our work in preprint form, results on the behavior of the ridge regression estimator---which in this case would coincide with the limiting MMSE estimator---for a model with a single layer of Gaussian linear random features were announced by Rocks and Mehta \cite{rocks2022bias}.  

\begin{acknowledgments}

We thank A. Atanasov, B. Bordelon, and B. S. Ruben for helpful comments on our manuscript. This work was supported by a Google Faculty Research Award and NSF Award \#2134157. A subset of the computations in this paper were performed using the Harvard University FAS Division of Science Research Computing Group's Cannon HPC cluster.
\end{acknowledgments}

\appendix

\section{Replica theory framework}\label{app:replica_framework}

In this appendix, we introduce the replica theory framework we use to compute learning curves. We direct the interested reader to \cite{mezard1987spin} for more details on replica theory. Our starting point is the partition function of the Bayes posterior:
\begin{align}
    Z = \mathbb{E}_{\mathcal{W}} \exp\left(-\frac{\beta}{2} \sum_{\mu=1}^{p} [g_{\mathbf{w}}(\mathbf{x}_{\mu}) - y_{\mu}]^2 \right).
\end{align}
In the limit of interest, we expect the quenched free energy 
\begin{align}
    f = - \lim_{d,p,n_{1},\ldots,n_{\ell} \to \infty} \frac{1}{d} \log Z,
\end{align}
to be self-averaging, i.e., $f = \mathbb{E}_{\mathcal{D}} f$ with probability one \cite{mezard1987spin,engel2001statistical}. To compute the limiting quenched average, we will resort to the replica trick, which proceeds via the identity $\mathbb{E}_{\mathcal{D}} \log Z = \lim_{m \to 0} \log(\mathbb{E}_{\mathcal{D}} Z^{m})/m$, yielding
\begin{align}
    f = - \lim_{m \to 0} \lim_{d,p,n_{1},\ldots,n_{\ell} \to \infty} \frac{1}{dm} \log \mathbb{E}_{\mathcal{D}} Z^{m}
\end{align}
after a non-rigorous interchange of the limits $m \to 0$ and $d,p,n_{1},\ldots,n_{\ell} \to \infty$. We evaluate the moments $\mathbb{E}_{\mathcal{D}} Z^{m}$ for positive integer $m$, and assume that they can be analytically continued to $m \to 0$. 

In this appendix, we show that we can write the disorder-averaged replicated partition function $\mathbb{E}_{\mathcal{D}} Z^{m}$ for each model as an integral over some set of order parameter matrices. We then introduce the replica-symmetric \emph{Ansatz} under which we will solve the resulting saddle-point equations in Appendix \ref{app:rs_saddle_point}. 

\begin{widetext}

\subsection{Integrating out the data}

We first integrate out the data. Introducing replicas indexed by $a = 1, \ldots, m$, the object of interest is the disorder-averaged replicated partition function: 
\begin{align}
    \mathbb{E}_{\mathcal{D}} Z^{m} = \mathbb{E}_{\{\mathcal{W}^{a}\}} \mathbb{E}_{\mathcal{D}} \exp\left(-\frac{\beta}{2} \sum_{a=1}^{m} \sum_{\mu=1}^{p} [g_{\mathbf{w}^{a}}(\mathbf{x}_{\mu}) - y_{\mu}]^2 \right).
\end{align}
where $\mathbf{w}^{a}$ denotes the end-to-end weight vector with appropriate replica indices for a given model. Using the fact that the training examples are independent and identically distributed, we have
\begin{align}
    \mathbb{E}_{\mathcal{D}} Z^{m}
    &= \mathbb{E}_{\{\mathcal{W}^{a}\}} \mathbb{E}_{\mathcal{D} \setminus \mathcal{X}} \left[ \mathbb{E}_{\mathbf{x},\xi} \exp\left(-\frac{\beta}{2} \sum_{a=1}^{m} [g_{\mathbf{w}^{a}}(\mathbf{x}) - y]^2 \right) \right]^{p},
\end{align}
where we use $\mathbb{E}_{\mathcal{D} \setminus \mathcal{X}}$ as shorthand for expectation with respect to all quenched disorder other than the training inputs and label noise. The expectations over $\mathbf{x}$ and $\xi$ are Gaussian integrals, hence we can evaluate them explicitly. After simplifying the resulting determinant with the aid of the matrix determinant lemma, the Weinstein-Aronzjan identity, and the push-through identity \cite{horn2012matrix}, this yields
\begin{align}
    \mathbb{E}_{\mathbf{x},\xi} \exp\left(-\frac{\beta}{2} \sum_{a=1}^{m} [g_{\mathbf{w}^{a}}(\mathbf{x}) - y]^2 \right) = \det(\mathbf{I}_{m} + \beta \mathbf{Q})^{-1/2} \left[ 1 + \beta \eta^{2} \mathbf{1}^{\top} (\mathbf{I}_{m} + \beta \mathbf{Q})^{-1} \mathbf{1} \right]^{-1/2},
\end{align}
where we have defined the $m \times m$ overlap matrix
\begin{align}
    Q^{ab} \equiv \frac{1}{d} (\mathbf{w}^{a} - \mathbf{w}_{\ast}) \cdot (\mathbf{w}^{b} - \mathbf{w}_{\ast}).
\end{align}
Enforcing the definition of the order parameter matrix $\mathbf{Q}$ by introducing corresponding Lagrange multipliers $\hat{\mathbf{Q}}$ \cite{mezard1987spin}, we therefore have
\begin{align}
    \mathbb{E}_{\mathcal{D}} Z^{m} = \int \frac{d\mathbf{Q}\,d\hat{\mathbf{Q}}}{(4 \pi i/d)^{m(m+1)/2}} & \exp\left(-\frac{d}{2} [ \tr(\mathbf{Q} \hat{\mathbf{Q}}) + \alpha m G_{1}(\mathbf{Q}) ] \right) S(\hat{\mathbf{Q}})
\end{align}
where we have defined
\begin{align}
    G_{1}(\mathbf{Q}) &\equiv \frac{1}{m} \log\det(\mathbf{I}_{m} + \beta \mathbf{Q}) + \frac{1}{m} \log\left[ 1 + \beta \eta^{2} \mathbf{1}^{\top} (\beta^{-1} \mathbf{I}_{m} + \mathbf{Q})^{-1} \mathbf{1} \right]
\end{align}
and
\begin{align}
    S(\hat{\mathbf{Q}}) \equiv \mathbb{E}_{\{\mathcal{W}^{a}\}} \mathbb{E}_{\mathcal{D} \setminus \mathcal{X}}  \exp\left(\frac{1}{2} \sum_{a,b} \hat{Q}^{ab} (\mathbf{w}^{a} - \mathbf{w}_{\ast}) \cdot (\mathbf{w}^{b} - \mathbf{w}_{\ast}) \right) .
\end{align}
Here, the integrals over $\mathbf{Q}$ and $\hat{\mathbf{Q}}$ are taken over the spaces of real and imaginary $m \times m$ symmetric matrices, respectively. Our remaining task is to integrate out the weights, which we will do for each of the three models of interest in the following sections. 

\subsection{Integrating out the weights for the LR model}

For simple linear regression, there is no quenched disorder other than the training inputs and we have $\mathbb{E}_{\{\mathcal{W}^{a}\}} = \mathbb{E}_{\{\mathbf{w}^{a}\}}$. In this case, all integrals are Gaussian, and we can easily compute
\begin{align}
    S(\hat{\mathbf{Q}}) &= \det(\mathbf{I}_{m} - \sigma^{2} \hat{\mathbf{Q}})^{-d/2} \exp\left[\frac{1}{2} \mathbf{1}^{\top} \hat{\mathbf{Q}}  ( \mathbf{I}_{m} - \sigma^{2} \hat{\mathbf{Q}})^{-1}  \mathbf{1} \Vert \mathbf{w}_{\ast} \Vert_{2}^{2} \right].
\end{align}
Using the assumption that $\Vert \mathbf{w}_{\ast} \Vert_{2}^{2} = d$ and defining
\begin{align}
    m\, G_{2}(\mathbf{Q},\hat{\mathbf{Q}}) &\equiv \tr(\mathbf{Q} \hat{\mathbf{Q}}) + \det(\mathbf{I}_{m} - \sigma^{2} \hat{\mathbf{Q}}) - \mathbf{1}^{\top} \hat{\mathbf{Q}}  ( \mathbf{I}_{m} - \sigma^{2} \hat{\mathbf{Q}})^{-1} \mathbf{1}
\end{align}
we can write the averaged replicated partition function of a single-layer network as
\begin{align}
    \mathbb{E}_{\mathcal{D}} Z_{\textrm{LR}}^{m} = \int \frac{d\mathbf{Q}\,d\hat{\mathbf{Q}}}{(4 \pi i/d)^{m(m+1)/2}} \exp\left(-\frac{1}{2} d m [ \alpha G_{1}(\mathbf{Q}) + G_{2}(\mathbf{Q},\hat{\mathbf{Q}}) ] \right) .
\end{align}

\subsection{Integrating out the weights for the RF model}

We now consider deep random feature models, for which we have $\mathbb{E}_{\mathcal{D} \setminus \mathcal{X}} = \mathbb{E}_{\mathbf{U}_{1},\ldots,\mathbf{U}_{\ell}}$ and $\mathbb{E}_{\{\mathcal{W}^{a}\}} = \mathbb{E}_{\{\mathbf{v}^{a}\}}$ We of course have
\begin{align}
    S &= \exp\left(\frac{1}{2} \mathbf{1}^{\top}\hat{\mathbf{Q}} \mathbf{1} \Vert \mathbf{w}_{\ast} \Vert_{2}^{2} \right) \mathbb{E}_{\mathbf{U}_{1},\ldots,\mathbf{U}_{\ell},\{\mathbf{v}^{a}\}} \exp\left(\frac{1}{2} \sum_{a,b} \hat{Q}^{ab} \mathbf{w}^{a} \cdot \mathbf{w}^{b} - \sum_{a,b} \hat{Q}^{ab} \mathbf{w}_{\ast} \cdot \mathbf{w}^{a} \right).
\end{align}
By introducing order parameters
\begin{align}
    C_{1}^{ab} \equiv \frac{1}{n_{1} \cdots n_{\ell}} (\mathbf{U}_{2} \cdots \mathbf{U}_{\ell} \mathbf{v}^{a}) \cdot (\mathbf{U}_{2} \cdots \mathbf{U}_{\ell} \mathbf{v}^{b})
\end{align}
via Fourier representations of the Dirac distribution with corresponding Lagrange multipliers $\hat{\mathbf{C}}_{1}$, we can integrate out $\mathbf{U}_{1}$, yielding
\begin{align}
    S &= \int \frac{d\mathbf{C}_{1}\,d\hat{\mathbf{C}}_{1}}{(4 \pi i /n_{1})^{m(m+1)/2}} \exp\left(-\frac{n_{1}}{2} \tr(\mathbf{C}_{1} \hat{\mathbf{C}}_{1}) - \frac{d}{2} \log\det(\mathbf{I}_{m} - \sigma^{2} \mathbf{C}_{1} \hat{\mathbf{Q}} ) + \frac{ \Vert \mathbf{w}_{\ast} \Vert_{2}^{2}}{2} \mathbf{1}^{\top} \hat{\mathbf{Q}} (\mathbf{I}_{m} - \sigma^{2} \mathbf{C}_{1} \hat{\mathbf{Q}})^{-1} \mathbf{1} \right)
    \nonumber\\&\qquad \times \mathbb{E}_{\mathbf{U}_{2},\ldots \mathbf{U}_{\ell},\{\mathbf{v}^{a}\}} \exp\left(\frac{1}{2 n_{2} \cdots n_{\ell}} \sum_{a,b} \hat{C}_{1}^{ab} (\mathbf{U}_{2} \cdots \mathbf{U}_{\ell} \mathbf{v}^{a})^{\top} \mathbf{U}_{2} \cdots \mathbf{U}_{\ell} \mathbf{v}^{b}\right). 
\end{align}
It is easy to see that we can iterate this procedure forward through the network, introducing order parameters
\begin{align}
    C_{l}^{ab} \equiv \frac{1}{n_{l} \cdots n_{\ell}} (\mathbf{U}_{l+1} \cdots \mathbf{U}_{\ell} \mathbf{v}^{a}) \cdot (\mathbf{U}_{l+1} \cdots \mathbf{U}_{\ell} \mathbf{v}^{b})
\end{align}
for $l = 1, \ldots, \ell-1$ and
\begin{align}
    C_{\ell}^{ab} \equiv \frac{1}{n_{\ell}} \mathbf{v}^{a} \cdot \mathbf{v}^{b}
\end{align}
along with corresponding Lagrange multipliers, yielding
\begin{align}
    S(\hat{\mathbf{Q}}) &= \int \frac{d\mathbf{C}_{1}\,d\hat{\mathbf{C}}_{1}}{(4 \pi i /n_{1})^{m(m+1)/2}} \cdots \int \frac{d\mathbf{C}_{\ell}\,d\hat{\mathbf{C}}_{\ell}}{(4 \pi i /n_{\ell})^{m(m+1)/2}} 
    \nonumber\\&\qquad \times \exp\left(- \frac{d}{2} \log\det(\mathbf{I}_{m} - \sigma^{2} \mathbf{C}_{1} \hat{\mathbf{Q}} ) + \frac{ \Vert \mathbf{w}_{\ast} \Vert_{2}^{2}}{2} \mathbf{1}^{\top} \hat{\mathbf{Q}} (\mathbf{I}_{m} - \sigma^{2} \mathbf{C}_{1} \hat{\mathbf{Q}})^{-1} \mathbf{1} \right)  \nonumber\\&\qquad \times \exp\left(-\frac{1}{2} \sum_{l=1}^{\ell-1} n_{l} \bigg[ \tr(\mathbf{C}_{l} \hat{\mathbf{C}}_{l}) + \log\det(\mathbf{I}_{m} - \mathbf{C}_{l+1} \hat{\mathbf{C}}_{l}) \bigg] \right) \nonumber\\&\qquad \times \exp\left( -\frac{1}{2} n_{\ell} \bigg[\tr(\mathbf{C}_{\ell} \hat{\mathbf{C}}_{\ell}) + \log\det(\mathbf{I}_{m} - \hat{\mathbf{C}}_{\ell}) \bigg] \right).
\end{align}
Then, using the assumption that $\Vert \mathbf{w}_{\ast} \Vert_{2}^{2} = d$ and defining 
\begin{align}
    m\, G_{2}(\mathbf{Q},\hat{\mathbf{Q}}, \mathbf{C}_{1} ) &\equiv  \tr(\mathbf{Q} \hat{\mathbf{Q}}) + \log\det(\mathbf{I}_{m} - \sigma^{2} \mathbf{C}_{1} \hat{\mathbf{Q}} ) - \mathbf{1}^{\top} \hat{\mathbf{Q}} (\mathbf{I}_{m} - \sigma^{2} \mathbf{C}_{1} \hat{\mathbf{Q}})^{-1} \mathbf{1}
\end{align}
and
\begin{align}
    m\, G_{3}(\mathbf{C}_{1}, \hat{\mathbf{C}}_{1}, \ldots, \mathbf{C}_{\ell},\hat{\mathbf{C}}_{\ell}) &\equiv \sum_{l=1}^{\ell-1} \gamma_{l} \bigg[ \tr(\mathbf{C}_{l} \hat{\mathbf{C}}_{l}) + \log\det(\mathbf{I}_{m} - \mathbf{C}_{l+1} \hat{\mathbf{C}}_{l}) \bigg] + \gamma_{\ell} \bigg[\tr(\mathbf{C}_{\ell} \hat{\mathbf{C}}_{\ell}) + \log\det(\mathbf{I}_{m} - \hat{\mathbf{C}}_{\ell}) \bigg], 
\end{align}
we can write the averaged replicated partition function as
\begin{align}
    \mathbb{E}_{\mathcal{D}} Z_{\textrm{RF}}^{m} &= \int \frac{d\mathbf{Q}\,d\hat{\mathbf{Q}}}{(4 \pi i/d)^{m(m+1)/2}} \int \frac{d\mathbf{C}_{1}\,d\hat{\mathbf{C}}_{1}}{(4 \pi i /n_{1})^{m(m+1)/2}} \cdots \int \frac{d\mathbf{C}_{\ell}\,d\hat{\mathbf{C}}_{\ell}}{(4 \pi i /n_{\ell})^{m(m+1)/2}}  \nonumber\\&\qquad \times \exp\left(-\frac{1}{2} d m F(\mathbf{Q},\hat{\mathbf{Q}}, \mathbf{C}_{1}, \hat{\mathbf{C}}_{1}, \ldots, \mathbf{C}_{\ell},\hat{\mathbf{C}}_{\ell} ) \right) ,
\end{align}
where we have defined
\begin{align}
    F(\mathbf{Q},\hat{\mathbf{Q}}, \mathbf{C}_{1}, \hat{\mathbf{C}}_{1}, \ldots, \mathbf{C}_{\ell},\hat{\mathbf{C}}_{\ell} ) \equiv \alpha G_{1}(\mathbf{Q}) + G_{2}(\mathbf{Q},\hat{\mathbf{Q}}, \mathbf{C}_{1}) + G_{3}(\mathbf{C}_{1}, \hat{\mathbf{C}}_{1}, \ldots, \mathbf{C}_{\ell},\hat{\mathbf{C}}_{\ell}).
\end{align}
We note that we have intentionally split the entropic contribution to the replica free energy into two pieces. The first, $G_{2}(\mathbf{Q},\hat{\mathbf{Q}},\mathbf{C}_{1})$, reduces to the entropic contribution for simple linear regression upon fixing $\hat{\mathbf{C}}_{1} = \mathbf{I}_{m}$. The second, $G_{3}$, captures the effect of depth. 

\subsection{Integrating out the weights for the NN model}

We now consider deep networks, for which there is no quenched disorder other than the training inputs, and $\mathbb{E}_{\{\mathcal{W}^{a}\}} = \mathbb{E}_{\{\mathbf{U}_{1}^{a},\ldots,\mathbf{U}_{\ell}^{a},\mathbf{v}^{a}\}}$. In this case, we have
\begin{align}
    S &= \exp\left(\frac{1}{2} \mathbf{1}^{\top}\hat{\mathbf{Q}} \mathbf{1} \Vert \mathbf{w}_{\ast} \Vert_{2}^{2} \right) \mathbb{E}_{\{\mathbf{U}_{1}^{a},\ldots,\mathbf{U}_{\ell}^{a},\mathbf{v}^{a}\}} \exp\left(\frac{1}{2} \sum_{a,b} \hat{Q}^{ab} \mathbf{w}^{a} \cdot \mathbf{w}^{b} - \sum_{a,b} \hat{Q}^{ab} \mathbf{w}_{\ast} \cdot \mathbf{w}^{a} \right) .
\end{align}
As we did for the RF model, we start by integrating out $\mathbf{U}_{1}^{a}$. We introduce analogous order parameters
\begin{align}
    C_{1}^{ab} \equiv \frac{1}{n_{1} \cdots n_{\ell}} (\mathbf{U}_{2}^{a} \cdots \mathbf{U}_{\ell}^{a} \mathbf{v}^{a}) \cdot (\mathbf{U}_{2}^{b} \cdots \mathbf{U}_{\ell}^{b} \mathbf{v}^{b}) .
\end{align}
However, importantly, as the weights $\mathbf{U}_{1}^{a}$ are annealed rather than quenched, the covariance of $h_{j}^{a}$ is replica-diagonal. For clarity of notation, we define the diagonal matrix
\begin{align}
    D_{1}^{ab} = \delta_{ab} C_{1}^{ab}.
\end{align}
Then, introducing a corresponding diagonal matrix of Lagrange multipliers $\hat{\mathbf{D}}_{1}$, we have
\begin{align}
    S &= \int \frac{d\mathbf{D}_{1}\,d\hat{\mathbf{D}}_{1}}{(4 \pi i /n_{1})^{m/2}} \exp\left(-\frac{n_{1}}{2} \tr(\mathbf{D}_{1} \hat{\mathbf{D}}_{1}) - \frac{d}{2} \log\det(\mathbf{I}_{m} - \sigma^{2} \mathbf{D}_{1} \hat{\mathbf{Q}} ) + \frac{ \Vert \mathbf{w}_{\ast} \Vert_{2}^{2}}{2} \mathbf{1}^{\top} \hat{\mathbf{Q}} (\mathbf{I}_{m} - \sigma^{2} \mathbf{D}_{1} \hat{\mathbf{Q}})^{-1} \mathbf{1} \right)
    \nonumber\\&\qquad\times \mathbb{E}_{\{\mathbf{U}_{2}^{a},\ldots \mathbf{U}_{\ell}^{a},\mathbf{v}^{a}\}} \exp\left(\frac{1}{2 n_{2} \cdots n_{\ell}} \sum_{a} \hat{D}_{1}^{aa} (\mathbf{U}_{2}^{a} \cdots \mathbf{U}_{2}^{b} \mathbf{v}^{a})^{\top} \mathbf{U}_{2}^{a} \cdots \mathbf{U}_{\ell}^{a} \mathbf{v}^{a}\right)
\end{align}
upon integrating out $\mathbf{U}_{1}^{a}$. We can see that we can follow much the same procedure to integrate out the remaining weights as we did for the random feature model, except for the fact that we only consider the replica-diagonal component of the overlaps, which are re-defined to include the replica indices of the hidden layer weights, i.e., 
\begin{align}
    C_{l}^{ab} \equiv \frac{1}{n_{l} \cdots n_{\ell}} (\mathbf{U}_{l+1}^{a} \cdots \mathbf{U}_{\ell}^{a} \mathbf{v}^{a}) \cdot (\mathbf{U}_{l+1}^{b} \cdots \mathbf{U}_{\ell}^{b} \mathbf{v}^{b})
\end{align}
for $l = 1, \ldots, \ell-1$ and
\begin{align}
    C_{\ell}^{ab} \equiv \frac{1}{n_{\ell}} \mathbf{v}^{a} \cdot \mathbf{v}^{b}.
\end{align}
We therefore obtain
\begin{align}
    \mathbb{E}_{\mathcal{D}} Z_{\textrm{NN}}^{m} &= \int \frac{d\mathbf{Q}\,d\hat{\mathbf{Q}}}{(4 \pi i/d)^{m(m+1)/2}} \int \frac{d\mathbf{D}_{1}\,d\hat{\mathbf{D}}_{1}}{(4 \pi i /n_{1})^{m/2}} \cdots \int \frac{d\mathbf{D}_{\ell}\,d\hat{\mathbf{D}}_{\ell}}{(4 \pi i /n_{\ell})^{m/2}}  \nonumber\\&\qquad \times \exp\left(-\frac{1}{2} d m F(\mathbf{Q},\hat{\mathbf{Q}}, \mathbf{D}_{1}, \hat{\mathbf{D}}_{1}, \ldots, \mathbf{D}_{\ell},\hat{\mathbf{D}}_{\ell} ) \right) 
\end{align}
where $F$ is the same as for the random feature model and the matrices $\mathbf{D}_{l}$ and $\hat{\mathbf{D}}_{l}$ are constrained to be replica-diagonal. This difference reflects the fact that the hidden layer weights of the NN model are annealed, rather than being quenched as in the RF model. 

\end{widetext}

\subsection{The replica-symmetric \emph{Ansatz}}

In the thermodynamic limit, we evaluate the integral over the appropriate order parameters and Lagrange multipliers for each model via the method of steepest descent. Importantly, we note that the diagonal components of the order parameters $Q^{aa}$ give the posterior-averaged generalization errors of the replicas, as in the thermodynamic limit the mean value of these parameters is given by the saddle-point equations. Our eventual objective is therefore simply to evaluate the saddle-point values of $\mathbf{Q}$ in the zero-temperature limit, and we will not consider the resulting values of the free energy. 

As usual in the replica method, we seek extrema in the limit $m \to 0$ of a constrained form, known as the replica-symmetric (RS) \emph{Ansatz} \cite{mezard1987spin}. For all three models, the RS \emph{Ansatz} for the variables $\mathbf{Q}$ and $\hat{\mathbf{Q}}$ is simply
\begin{align}
    \mathbf{Q}_{\textrm{RS}} &= (Q-q) \mathbf{I}_{m} + q \mathbf{1}\mathbf{1}^{\top}
    \\
    \hat{\mathbf{Q}}_{\textrm{RS}} &= (\hat{Q}-\hat{q}) \mathbf{I}_{m} + \hat{q} \mathbf{1} \mathbf{1}^{\top}.
\end{align}
For a deep random feature model, the RS \emph{Ansatz} for the remaining order parameters is
\begin{align}
    \mathbf{C}_{l,\textrm{RS}} &= (C_{l}-c_{l}) \mathbf{I}_{m} + c_{l} \mathbf{1}\mathbf{1}^{\top} && (l=1,\ldots,\ell)
    \\
    \hat{\mathbf{C}}_{l,\textrm{RS}} &= (\hat{C}_{l}-\hat{c}_{l}) \mathbf{I}_{m} + \hat{c}_{l} \mathbf{1} \mathbf{1}^{\top} && (l=1,\ldots,\ell),
\end{align}
while, for a deep network, the RS \emph{Ansatz} for the remaining order parameters is
\begin{align}
    \mathbf{D}_{l,\textrm{RS}} &= C_{l} \mathbf{I}_{m} && (l=1,\ldots,\ell)
    \\
    \hat{\mathbf{D}}_{l,\textrm{RS}} &= \hat{C}_{l} \mathbf{I}_{m} && (l=1,\ldots,\ell) ,
\end{align}
as we consider only the replica-diagonal components of the overlaps $\mathbf{C}_{l}$. With this \emph{Ansatz}, one can simplify the expressions for the free energy and the saddle-point equations in the limit $m \to 0$. These manipulations are standard exercises using the properties of RS matrices \cite{mezard1987spin}, hence we will only report the results (in Appendix \ref{app:rs_saddle_point}). 

We remark briefly on the conditions under which the RS order parameters make physical sense given their definitions. We must have $Q \geq 0$ and $C_{l} \geq 0$ for all $l$, as these quantities are the squares of norms of vectors. If $C_{l} = 0$ for any $l$, the norm of the end-to-end weight vector tends to zero, and we must have a trivial solution with $Q = 1$. We must also have $Q - q \geq 0$ and $C_{l} - c_{l} \geq 0$ for all $l$. Moreover, if $Q - q = 0$ (respectively $C_{l} - c_{l} = 0$ for some $l$), then we must have $q \geq 0$ (respectively $c_{l} \geq 0$), to obtain a nontrivial physical solution.

\section{Solution of the replica-symmetric saddle point equations}\label{app:rs_saddle_point}

In this appendix, we analyze the replica-symmetric saddle point equations in the zero-temperature limit. 

\subsection{LR model}

For simple linear regression, the RS saddle point is given by a 4-dimensional system of equations, which decouples into a two-dimensional nonlinear system for the replica-nonuniform components $z \equiv Q-q$ and $\hat{z} \equiv \hat{Q} - \hat{q}$,
\begin{align}
    \hat{z} &=  - \frac{\alpha}{\beta^{-1} + z}
    \\
    z &= \frac{\sigma^{2} }{1 - \sigma^{2} \hat{z}}, 
\end{align}
and a linear system for the replica-uniform components $q$ and $\hat{q}$:
\begin{align}
    \hat{q} &= \frac{\alpha (q+\eta^{2})}{(\beta^{-1} + z)^2} 
    \\
    q &= \frac{1 + \sigma^{4} \hat{q}}{(1 - \sigma^{2} \hat{z})^2} . 
\end{align}
Using the expression for $\hat{z}$ as a function of $z$, we obtain the quadratic condition
\begin{align}
    z^2 -  [ \sigma^{2} (1-\alpha) - \beta^{-1} ] z - \sigma^{2} \beta^{-1} = 0 . 
\end{align}
The two solutions $z_{\pm}$ to this quadratic equation have zero-temperature limits
\begin{align}
    \lim_{\beta \to \infty} z_{\pm} = \frac{1}{2} \sigma^{2} (1 - \alpha \pm |\alpha-1|) .
\end{align}
For $\alpha > 1$, $z_{-}$ is negative, and is therefore unphysical. If $0 < \alpha < 1$,
\begin{align}
    z_{+} = \sigma^{2} (1-\alpha) + \frac{\alpha}{1-\alpha} \frac{1}{\beta} + \mathcal{O}(\beta^{-2}),
\end{align}
while if $\alpha > 1$,
\begin{align}
    z_{+} = \frac{1}{\alpha - 1} \frac{1}{\beta} + \mathcal{O}(\beta^{-2}).
\end{align}
These are the two low-temperature scalings we would expect to be self-consistent given the saddle point equations; we could alternatively derive the above solutions by assuming these scalings for $z$. 

Considering the replica-uniform components, we use the expressions for $1-\sigma^2 \hat{z}$ and $\hat{q}$ as functions of $z$ and $q$ to write
\begin{align}
    q 
    = \frac{z^2}{\sigma^{4}} (1 + \sigma^{4} \hat{q})
    = \frac{z^2}{\sigma^{4}} + \frac{\alpha z^2}{(\beta^{-1} + z)^2} (q+\eta^{2}).
\end{align}
For the solution with $z \sim \mathcal{O}(1)$, we then have
\begin{align}
    (1-\alpha) q = \frac{z^2}{\sigma^{4}} + \alpha \eta^{2} 
\end{align}
hence, recalling that this scaling yields $z = \sigma^{2} (1-\alpha)$ and is valid for $0 < \alpha < 1$,
\begin{align}
    q = 1-\alpha + \frac{\alpha}{1-\alpha} \eta^{2}.  
\end{align}
For the solution with $z \sim r/\beta$ with $r \sim \mathcal{O}(1)$, we have
\begin{align}
    \left[1 - \alpha \left(\frac{r}{1+r} \right)^{2} \right] q = \frac{r^2}{\beta^{2} \sigma^{4}} + \alpha \left(\frac{r}{1 + r} \right)^{2} \eta^{2} ,
\end{align}
hence, recalling that this scaling yields $r = 1/(\alpha-1)$ and is valid for $\alpha > 1$,
\begin{align}
    \frac{\alpha - 1}{\alpha} q = \frac{1}{\beta^{2} \sigma^{4} (\alpha-1)^2} + \frac{1}{\alpha} \eta^{2} ,
\end{align}
or
\begin{align}
    q = \frac{1}{\alpha - 1} \eta^{2}. 
\end{align}
Combining these results, we obtain a zero-temperature solution which gives the result for $\epsilon = Q$ reported in the main text. 

\subsection{RF model}

For a deep random feature model, the RS saddle-point is given by a $4(\ell+1)$-dimensional system of equations. As in the single-layer case, this system decouples into two sets of equations. Defining
\begin{align}
    z &\equiv Q - q
    \\
    \hat{z} &\equiv \hat{Q} - \hat{q}
    \\
    w_{l} &\equiv C_{l} - c_{l} 
    \\
    \hat{w}_{l} &\equiv \hat{C}_{l} - \hat{c}_{l},
\end{align}
the deviations from uniformity across replicas are determined by the $2 (\ell+1)$-dimensional closed system
\begin{align} 
    \hat{z} &= - \frac{\alpha}{\beta^{-1} + z}
    \\
    z &= \frac{\sigma^{2} w_{1} }{1 - \sigma^{2} w_{1} \hat{z}} 
    \\
    \hat{w}_{1} &= \frac{\sigma^{2}}{\gamma_{1}} \frac{\hat{z}}{1 - \sigma^{2} w_{1} \hat{z}}
    \\
    \hat{w}_{l} &= \frac{\gamma_{l-1}}{\gamma_{l}} \frac{\hat{w}_{l-1}}{1-\hat{w}_{l-1} w_{l}} && (l=2,\ldots,\ell) 
    \\
    w_{l} &= \frac{w_{l+1}}{1-w_{l+1} \hat{w}_{l}} && (l=1,\ldots,\ell-1) 
    \\
    w_{\ell} &= \frac{1}{1 - \hat{w}_{\ell}} .
\end{align}
We can then solve the remaining equations for the replica-uniform components,
\begin{align} \label{eqn:rf_replica_uniform_sp}
    \hat{q} &= \frac{\alpha (q+\eta^{2})}{(\beta^{-1} + Q - q)^2}
    \\
    q &= \frac{1 + \sigma^{2} c_{1}  + \sigma^{4} w_{1}^2 \hat{q}}{(1 - \sigma^{2} w_{1} \hat{z})^2}
    \\
    \hat{c}_{1} &= \frac{\sigma^{2}}{\gamma_{1}} \frac{\hat{q} + \hat{z}^{2} (1 + \sigma^{2} c_{1})}{(1 - \sigma^{2} w_{1} \hat{z})^2} 
    \\
    \hat{c}_{l} &= \frac{\gamma_{l-1}}{\gamma_{l}} \frac{\hat{c}_{l-1} + \hat{w}_{l-1}^2 c_{l} }{(1-\hat{w}_{l-1} w_{l})^{2}} && (l=2,\ldots,\ell) 
    \\
    c_{l} &= \frac{ c_{l+1} +  w_{l+1}^2 \hat{c}_{l}}{(1-w_{l+1} \hat{w}_{l})^2} && (l=1,\ldots,\ell-1)
    \\
    c_{\ell} &= \frac{\hat{c}_{\ell}}{(1 - \hat{w}_{l})^2},
\end{align}
for fixed values of these parameters. Importantly, we note that this is a set of $2 (\ell+1)$ linear equations for $2(\ell+1)$ variables. 

\subsubsection{Solving for the replica-nonuniform components}

We first consider the replica nonuniform components. We start by noting that the equations for $z$ and $\hat{z}$ yield
\begin{align}
    w_{1} = \frac{1}{\sigma^2} \frac{z}{1 + z \hat{z}} ,
\end{align}
hence the equation for $\hat{w}_{1}$ yields
\begin{align}
    \hat{w}_{1} &= \frac{\sigma^2}{\gamma_{1}} \hat{z} (1 + z \hat{z}).
\end{align}
If $\ell = 1$, then we are nearly done. The condition $w_{1} = 1/(1-\hat{w}_{1})$ gives
\begin{align}
    \hat{w}_{1} 
    = 1 - \frac{1}{w_{1}} 
    = \frac{z - \sigma^{2} (1 + z \hat{z})}{z} 
\end{align}
whence 
\begin{align}
    \frac{\sigma^2}{\gamma_{1}} \hat{z} (1 + z \hat{z}) = \frac{z - \sigma^{2} (1 + z \hat{z})}{z} ,
\end{align}
and therefore
\begin{align}
    z = \sigma^{2} \frac{(1 + z \hat{z}) (\gamma_{1} + z \hat{z})}{\gamma_{1}} .
\end{align}

If $\ell > 1$, we observe that a solution with any $w_{l} = 0$ must have all $w_{l} = 0$ and $z = 0$. Similarly, a solution with one $\hat{w}_{l} = 0$ must have all $\hat{w}_{l} = 0$ and $\hat{z} = 0$. As $w_{\ell} = 1/(1-\hat{w}_{\ell})$, these situations cannot coexist. Moreover, neither is self-consistent unless $\alpha = 0$ or $\beta$ is strictly infinite or zero. 

With this observation in mind, we will eliminate the Lagrange multipliers $\hat{w}_{l}$. Formally defining $w_{l+1} \equiv 1$ for convenience, we have
\begin{align}
    \hat{w}_{l} = \frac{w_{l} - w_{l+1}}{w_{l}w_{l+1}}  
\end{align}
for $l = 1, \ldots, \ell$. Then, for $l = 2, \ldots, \ell$, the equation
\begin{align}
    \hat{w}_{l} &= \frac{\gamma_{l-1}}{\gamma_{l}} \frac{\hat{w}_{l-1}}{1-\hat{w}_{l-1} w_{l}}
\end{align}
yields
\begin{align}
    \frac{w_{l} - w_{l+1}}{w_{l}w_{l+1}} = \frac{\gamma_{l-1}}{\gamma_{l}} \frac{w_{l-1}}{w_{l}} \frac{w_{l-1} - w_{l}}{w_{l-1} w_{l}} .
\end{align}
We now consider the equation
\begin{align}
    \hat{w}_{1} = \frac{\sigma^2}{\gamma_{1}} \hat{z} (1 + z \hat{z}) .
\end{align}
As 
\begin{align}
    1 + z \hat{z} = \frac{1}{\sigma^2} \frac{z}{w_{1}} ,
\end{align}
we can re-write this as
\begin{align}
    \hat{w}_{1} = \frac{z \hat{z}}{\gamma_{1} w_{1}} ,
\end{align}
which in turn implies that
\begin{align}
    \frac{w_{1} - w_{2}}{w_{1}w_{2}} = \frac{z \hat{z}}{\gamma_{1} w_{1}}.
\end{align}
Then, 
\begin{align}
    \frac{w_{2} - w_{3}}{w_{2}w_{3}} 
    = \frac{\gamma_{1}}{\gamma_{2}} \frac{w_{1}}{w_{2}} \frac{w_{1} - w_{2}}{w_{1} w_{2}}
    = \frac{z \hat{z}}{\gamma_{2} w_{2}}
\end{align}
hence we can see that
\begin{align}
    \hat{w}_{l} = \frac{w_{l} - w_{l+1}}{w_{l} w_{l+1}} = \frac{z \hat{z}}{\gamma_{l} w_{l}}
\end{align}
for $l = 1, \ldots, \ell$. This yields the backward recurrence
\begin{align}
    w_{l} = \frac{\gamma_{l} + z \hat{z}}{\gamma_{l}} w_{l+1} ,
\end{align}
which can be solved using the endpoint condition $w_{\ell+1} \equiv 1$, yielding
\begin{align}
    w_{l} = \frac{(\gamma_{l} + z \hat{z}) (\gamma_{l+1} + z \hat{z}) \cdots (\gamma_{\ell} + z \hat{z})}{\gamma_{l} \gamma_{l+1} \cdots \gamma_{\ell}} . 
\end{align}
In particular,
\begin{align}
    w_{1} = \frac{(\gamma_{1} + z \hat{z}) (\gamma_{2} + z \hat{z}) \cdots (\gamma_{\ell} + z \hat{z})}{\gamma_{1} \gamma_{2} \cdots \gamma_{\ell}} , 
\end{align}
which yields a self-consistent equation for $z$
\begin{align}
    z = \sigma^{2} \frac{(1 + z \hat{z}) (\gamma_{1} + z \hat{z}) (\gamma_{2} + z \hat{z}) \cdots (\gamma_{\ell} + z \hat{z})}{\gamma_{1} \gamma_{2} \cdots \gamma_{\ell}}  
\end{align}
using the condition
\begin{align}
    \hat{z} = - \frac{\alpha}{\beta^{-1} + z} .
\end{align}
This coincides with the result we obtained earlier for $\ell = 1$. 

As in the single-layer case, it can be seen that the self-consistent scalings for $z$ in the zero-temperature limit are $z \sim \mathcal{O}(1)$ and $z \sim \mathcal{O}(1/\beta)$. If we take $\beta \to \infty$ with $z \sim \mathcal{O}(1)$, we simply have $z \hat{z} \to - \alpha$, which gives
\begin{align}
    z = \sigma^{2} \frac{(1 - \alpha ) (\gamma_{1} - \alpha ) (\gamma_{2} - \alpha ) \cdots (\gamma_{\ell} - \alpha )}{\gamma_{1} \gamma_{2} \cdots \gamma_{\ell}}  .
\end{align}
This scaling gives
\begin{align}
    w_{l} \to \frac{(\gamma_{l} - \alpha ) (\gamma_{l+1} - \alpha ) \cdots (\gamma_{\ell} - \alpha )}{\gamma_{l} \gamma_{l+1} \cdots \gamma_{\ell}}  \sim \mathcal{O}(1)
\end{align}
for all $l$. As physical solutions have $z \geq 0$ and all $w_{l} \geq 0$, this solution is sensible in the regime $\alpha < \min\{1, \gamma_{1}, \gamma_{2}, \ldots, \gamma_{l}\}$. 

If we take $z \sim r/\beta$ for $r \sim \mathcal{O}(1)$, we have
\begin{align}
    z \hat{z} \to - \frac{\alpha r}{1+r}
\end{align}
and the limiting equation
\begin{align}
    0 = \frac{\sigma^2}{\gamma_{1} \gamma_{2} \cdots \gamma_{\ell}} \left(1 - \frac{\alpha r}{1+r}\right) \prod_{l=1}^{\ell} \left(\gamma_{l} - \frac{\alpha r}{1+r}\right) .
\end{align}
This yields $\ell+1$ solutions
\begin{align}
    r_{0} &= \frac{1}{\alpha - 1}
    \\
    r_{l_{\ast}} &= \frac{\gamma_{l_{\ast}}}{\alpha - \gamma_{l_{\ast}}} \qquad (l_{\ast} = 1,\ldots,\ell). 
\end{align}
For the zeroth solution with $r_{0} = \frac{1}{\alpha - 1}$, we have $z \hat{z} \to - 1$, and thus
\begin{align}
    w_{l} \to \frac{(\gamma_{l} - 1) (\gamma_{l+1} - 1) \cdots (\gamma_{\ell} - 1)}{\gamma_{l} \gamma_{l+1} \cdots \gamma_{\ell}} \sim \mathcal{O}(1)
\end{align}
This solution is therefore physical for $\alpha > 1$ and all $\gamma_{l} > 1$. For the $l_{\ast}$-th such solution, we have $z \hat{z} \to - \gamma_{l_{\ast}}$, hence 
\begin{align}
    w_{l} \to \frac{(\gamma_{l} - \gamma_{l_{\ast}}) (\gamma_{l+1} - \gamma_{l_{\ast}}) \cdots (\gamma_{\ell} - \gamma_{l_{\ast}})}{\gamma_{l} \gamma_{l+1} \cdots \gamma_{\ell}} . 
\end{align}
Thus, we have $w_{l} \to 0$ for all $l \leq l_{\ast}$. For $l > l_{\ast}$, $w_{l} \sim \mathcal{O}(1)$, and we must have $\gamma_{l} \geq \gamma_{l_{\ast}}$ for all $l > l_{\ast}$ such that $w_{l} \geq 0$. Moreover, we must have $\alpha > \gamma_{l_{\ast}}$, such that $r_{l_{\ast}} > 0$. We will obtain further conditions on the validity of these solutions from solving for the replica-uniform components.

\subsubsection{Solving for the replica-uniform components}

We now consider the linear system of equations \eqref{eqn:rf_replica_uniform_sp} that determines the replica-uniform components in terms of the non-uniform components. We start by noting that we can $\hat{c}_{1}$ express a function of $q$ alone, eliminating the dependence on $c_{1}$ using the equation for $q$: 
\begin{align}
    \hat{c}_{1} = \frac{\sigma^{2}}{\gamma_{1}} \hat{z}^{2} \left( \frac{1 + 2 z \hat{z}}{\alpha} (q + \eta^2) + q \right)
\end{align}
where we have used the fact that $\hat{q} = \hat{z}^{2} (q + \eta^2)/\alpha$.

If $\ell = 1$, we can use the equation $w_{1} = 1/(1 - \hat{w}_{1})$ to obtain
\begin{align}
    c_{1} = \frac{\hat{c}_{1}}{(1 - \hat{w}_{1})^2} = w_{1}^{2} \hat{c}_{1},
\end{align}
which will give a closed equation for $q$. If $\ell > 1$, our task is somewhat more complex. We eliminate the Lagrange multipliers via
\begin{align}
    \hat{c}_{l} &= \left(\frac{1-w_{l+1} \hat{w}_{l}}{w_{l+1}}\right )^2 c_{l} - \frac{1}{w_{l+1}^2} c_{l+1}  \qquad (l=1,\ldots,\ell)
\end{align}
where we have defined $w_{\ell+1} \equiv 1$ and $c_{\ell+1} \equiv 0$ for convenience. Then, for $l = 2, \ldots, \ell$, the equation
\begin{align}
    \hat{c}_{l} = \frac{\gamma_{l-1}}{\gamma_{l}} \frac{\hat{c}_{l-1} + \hat{w}_{l-1}^2 c_{l} }{(1-\hat{w}_{l-1} w_{l})^{2}}
\end{align}
yields the three-term recurrence 
\begin{align}
    \frac{\gamma_{l-1}}{\gamma_{l} w_{l}^{2}} c_{l-1} 
    &= \left[ \frac{(1-w_{l+1} \hat{w}_{l})^2}{w_{l+1}^2} + \frac{\gamma_{l-1}}{\gamma_{l}}  \frac{1 - w_{l}^{2} \hat{w}_{l-1}^{2}}{w_{l}^2 (1-\hat{w}_{l-1} w_{l})^{2}} \right] c_{l} \nonumber\\&\quad - \frac{1}{w_{l+1}^2} c_{l+1}
\end{align}
with initial difference condition
\begin{align}
    \hat{c}_{1}(q) = \left(\frac{1-w_{2} \hat{w}_{1}}{w_{2}}\right )^2 c_{1} - \frac{1}{w_{2}^2} c_{2} 
\end{align}
and endpoint condition $c_{\ell+1} = 0$. Substituting in $\hat{w}_{l} = \frac{z \hat{z}}{\gamma_{l} w_{l}}$ and using the backward recurrence $w_{l} = \frac{\gamma_{l} + z \hat{z}}{\gamma_{l}} w_{l+1} $, we have
\begin{align}
    \frac{\gamma_{l-1}}{\gamma_{l} } c_{l-1} 
    &= \left[ 1 + \frac{\gamma_{l-1}}{\gamma_{l}}  \frac{(\gamma_{l-1} + z \hat{z})^2 - (z\hat{z})^{2}}{\gamma_{l-1}^{2}} \right] c_{l} \nonumber\\&\quad  - \frac{(\gamma_{l} + z \hat{z})^2}{\gamma_{l}^{2}} c_{l+1} . 
\end{align}
Similarly, we can simplify the initial difference condition to 
\begin{align}
    \hat{c}_{1}(q) = \frac{1}{w_{1}^2} c_{1} - \frac{1}{w_{2}^2} c_{2}, 
\end{align}
hence, substituting in $w_{1} = \frac{1}{\sigma^2} \frac{z}{1 + z \hat{z}}$, we have
\begin{align}
    w_{1}^{2} \hat{c}_{1}(q) 
    = c_{1} - \left(\frac{\gamma_{1} + z \hat{z}}{\gamma_{1}}\right)^2 c_{2}. 
\end{align}
To simplify our remaining task, we define new variables $u_{l}$ such that
\begin{align}
    c_{l} = \gamma_{1} w_{1}^{2} \hat{c}_{1}(q) u_{l}. 
\end{align}
If $\ell = 1$, we simplify have $u_{1} = 1/\gamma_{1}$. For $\ell > 1$, these variables are determined by the recurrence
\begin{align}
    \frac{\gamma_{l-1}}{\gamma_{l} } u_{l-1} 
    &= \left[ 1 + \frac{\gamma_{l-1}}{\gamma_{l}}  \frac{(\gamma_{l-1} + z \hat{z})^2 - (z\hat{z})^{2}}{\gamma_{l-1}^{2}} \right] u_{l} \nonumber\\&\quad - \frac{(\gamma_{l} + z \hat{z})^2}{\gamma_{l}^{2}} u_{l+1} 
\end{align}
with initial difference condition 
\begin{align}
    \frac{1}{\gamma_{1}} = u_{1} - \left(\frac{\gamma_{1} + z \hat{z}}{\gamma_{1}}\right)^{2} u_{2} 
\end{align}
and endpoint condition $u_{\ell+1} = 0$. We note that the initial difference and endpoint conditions give the consistent result $u_{1} = 1/\gamma_{1}$ when $\ell = 1$.

Given a solution to the recurrence for the variables $u_{l}$, we then have a closed equation for $q$:
\begin{align}
    q &= (1 + z \hat{z})^2 \nonumber\\&\quad + (z\hat{z})^{2} \left(\frac{(1 + \alpha + 2 z \hat{z}) u_{1} + 1}{\alpha}   (q + \eta^2) - \eta^{2} \right). 
\end{align}
With this solution in hand, we can then obtain $c_{l}$ via the relation $c_{l} = \gamma_{1} w_{1}^{2} \hat{c}_{1}(q) u_{l}$.

We now consider the zero-temperature limits of interest. With $z \sim \mathcal{O}(1)$, we have $z \hat{z} \to - \alpha$. The limiting equation for $q$ is then 
\begin{align}
    q &= (1 - \alpha)^{2} + \alpha [(1 -\alpha) u_{1} + 1] (q + \eta^2) - \eta^{2} \alpha^2 u_{1} ,
\end{align}
which yields
\begin{align}
    q = (1 - \alpha) \left[1 + \frac{\alpha u_{1}}{1 - \alpha u_{1}} \right] +  \left[\frac{\alpha}{1 - \alpha} + \frac{\alpha u_{1}}{1-\alpha u_{1}} \right] \eta^{2}. 
\end{align}
Considering the recurrence for $u_{l}$, we have
\begin{align}
    \frac{\gamma_{l-1}}{\gamma_{l} } u_{l-1} = \frac{\gamma_{l} + \gamma_{l-1} - 2 \alpha}{\gamma_{l}} u_{l}  - \frac{(\gamma_{l} - \alpha)^2}{\gamma_{l}^{2}} u_{l+1} .
\end{align}
for $l = 2, \ldots, \ell$, and the initial difference condition
\begin{align}
    \frac{1}{\gamma_{1}} = u_{1} - \left(\frac{\gamma_{1} -\alpha}{\gamma_{1}}\right)^{2} u_{2} .
\end{align}
We can re-express this recurrence as
\begin{align}
    \frac{\gamma_{l} - \alpha}{\gamma_{l}} u_{l+1} - u_{l}
    &= \frac{\gamma_{l-1}}{\gamma_{l}} \frac{\gamma_{l}}{\gamma_{l} - \alpha} \left( \frac{\gamma_{l-1} - \alpha}{\gamma_{l-1}} u_{l} -  u_{l-1} \right) , 
\end{align}
which can easily be iterated backward, yielding 
\begin{align}
    u_{l} &= \frac{\gamma_{l} - \alpha}{\gamma_{l}} u_{l+1} \nonumber\\&\quad + \frac{1}{\gamma_{l}} \frac{\gamma_{l}}{\gamma_{l} - \alpha} \frac{\gamma_{l-1}}{\gamma_{l-1} - \alpha} \cdots \frac{\gamma_{2}}{\gamma_{2} - \alpha} [\gamma_{1} u_{1} - (\gamma_{1} - \alpha) u_{2} ]
\end{align}
for $l = 2, \ldots, \ell$. Then, the termination condition $u_{\ell+1} = 0$ implies that
\begin{align}
    u_{\ell} = \frac{1}{\gamma_{\ell}} \frac{\gamma_{\ell}}{\gamma_{\ell} - \alpha} \frac{\gamma_{\ell-1}}{\gamma_{\ell-1} - \alpha} \cdots \frac{\gamma_{2}}{\gamma_{2} - \alpha} [\gamma_{1} u_{1} - (\gamma_{1} - \alpha) u_{2} ],
\end{align}
hence, iterating one step backwards, we find that
\begin{align}
    u_{\ell-1} &= \left( \frac{1}{\gamma_{\ell} - \alpha}  + \frac{1}{\gamma_{\ell-1} - \alpha} \right) \frac{\gamma_{\ell-2}}{\gamma_{\ell-2} - \alpha} \cdots \frac{\gamma_{2}}{\gamma_{2} - \alpha} \nonumber\\&\quad \times  [\gamma_{1} u_{1} - (\gamma_{1} - \alpha) u_{2} ].
\end{align}
It is now easy to see that we can iterate this process backwards, yielding
\begin{align}
    u_{l} &= \left( \sum_{j=l}^{\ell} \frac{1}{\gamma_{l} - \alpha} \right) \frac{\gamma_{l-1}}{\gamma_{l-1} - \alpha} \cdots \frac{\gamma_{2}}{\gamma_{2} - \alpha} \nonumber\\&\qquad \times [\gamma_{1} u_{1} - (\gamma_{1} - \alpha) u_{2} ].
\end{align}
for $l=2,\ldots,\ell$, with
\begin{align}
    u_{2} = [\gamma_{1} u_{1} - (\gamma_{1} - \alpha) u_{2} ] \sum_{j=2}^{\ell} \frac{1}{\gamma_{l} - \alpha} 
\end{align}
in particular. Using the initial difference condition to express $u_{2}$ in terms of $u_{1}$, we then obtain a closed equation for $u_{1}$:
\begin{align}
    u_{1} - \frac{1}{\gamma_{1}} = \frac{\gamma_{1} - \alpha}{\gamma_{1}} (1 - \alpha u_{1}) \sum_{j=2}^{\ell} \frac{1}{\gamma_{l} - \alpha} .
\end{align}
As this equation is linear, it is easy to solve, yielding
\begin{align}
    1 - \alpha u_{1} 
    &= \frac{\gamma_{1} - \alpha}{\gamma_{1} + \alpha (\gamma_{1} - \alpha) \sum_{j=2}^{\ell} \frac{1}{\gamma_{l} - \alpha}}
    \\
    &= \frac{1}{1 + \sum_{j=1}^{\ell} \frac{\alpha}{\gamma_{l} - \alpha}}
\end{align}
under the assumption that $\alpha \neq \gamma_{l}$ for all $l$. Then, we have
\begin{align}
    \frac{\alpha u_{1}}{1-\alpha u_{1}} = \sum_{j=1}^{\ell} \frac{\alpha}{\gamma_{l} - \alpha} ,
\end{align}
which yields
\begin{align}
    q &= (1 - \alpha) \left(1 + \sum_{j=1}^{\ell} \frac{\alpha}{\gamma_{l} - \alpha} \right)  \nonumber\\&\quad + \left(\frac{\alpha}{1 - \alpha} + \sum_{j=1}^{\ell} \frac{\alpha}{\gamma_{l} - \alpha} \right) \eta^{2}. 
\end{align}
Moreover, for $l = 2, \ldots, \ell$, we have the solution 
\begin{align}
    u_{l} 
    &= \left( \sum_{j=l}^{\ell} \frac{1}{\gamma_{l} - \alpha} \right) \frac{\gamma_{l-1}}{\gamma_{l-1} - \alpha} \cdots \frac{\gamma_{1}}{\gamma_{1} - \alpha} (1 - \alpha u_{1}) 
\end{align}
in terms of the solution for $1-\alpha u_{1}$. Then, in terms of these solutions for $u_{l}$, we have
\begin{align}
    c_{l} = \frac{ \alpha}{\sigma^2} \frac{1 - \alpha + \eta^2}{1-\alpha} \left(1 + \sum_{j=1}^{\ell} \frac{\alpha}{\gamma_{l} - \alpha}\right) u_{l} . 
\end{align}

We now consider the solutions with $z \sim r/\beta$ for $r \sim \mathcal{O}(1)$. We first consider the solution with
\begin{align}
    r_{0} &= \frac{1}{\alpha - 1}. 
\end{align}
For this solution, $z \hat{z} \to -1$, and the limiting self-consistent equation for $q$ reduces to
\begin{align}
    q = \frac{( \alpha - 1) u_{1} + 1}{\alpha}  (q + \eta^2) - \eta^{2} u_{1},
\end{align}
which yields
\begin{align}
    q = \frac{1}{\alpha - 1} \eta^{2}
\end{align}
for any $u_{1}$. This is non-negative throughout the expected region of physical validity ($\alpha > 1$), and therefore the overall solution makes sense given that $z \to 0$ in this regime. The recurrence for $u_{l}$ simplifies to 
\begin{align}
    \frac{\gamma_{l-1}}{\gamma_{l} } u_{l-1} 
    &= \left[ 1 + \frac{\gamma_{l-1}}{\gamma_{l}}  \frac{(\gamma_{l-1} - 1)^2 - 1}{\gamma_{l-1}^{2}} \right] u_{l}  \nonumber\\&\quad - \frac{(\gamma_{l} - 1)^2}{\gamma_{l}^{2}} u_{l+1} 
\end{align}
with initial difference condition 
\begin{align}
     \frac{1}{\gamma_{1}} = u_{1} - \left(\frac{\gamma_{1} - 1}{\gamma_{1}}\right)^{2} u_{2} 
\end{align}
and endpoint condition $u_{\ell+1} = 0$. This is exactly analogous to the recurrence we obtained when considering the solution with $z \sim \mathcal{O}(1)$ with $\alpha$ set to 1, hence we conclude immediately that $u_{1}$ is given by
\begin{align}
    1 - u_{1} = \frac{1}{1 + \sum_{j=1}^{\ell} \frac{1}{\gamma_{l} - 1}},
\end{align}
while
\begin{align}
    u_{l} 
    = \left( \sum_{j=l}^{\ell} \frac{1}{\gamma_{l} - 1} \right) \frac{\gamma_{l-1}}{\gamma_{l-1} - 1} \cdots \frac{\gamma_{1}}{\gamma_{1} - 1}  (1 - u_{1})
\end{align}
for $l = 2,\ldots,\ell$. These results are positive throughout the region of interest, i.e., $\gamma_{l} > 1$ for all $l$. For this solution, we must be somewhat careful in simplifying the equation for $c_{l}$. We have the limiting equation
\begin{align}
    c_{l} = \frac{1}{\sigma^2} \bigg[ 1 + \frac{1}{\alpha} q + \sigma^{2} c_{1} \bigg] u_{l} .
\end{align}
This yields a self-consistent equation for $c_{1}$, which gives
\begin{align}
    c_{1} 
    &= \frac{1}{\sigma^{2}} \frac{q + \alpha}{\alpha} \frac{u_{1}}{1-u_{1}} 
    = \frac{1}{\sigma^{2}} \frac{q + \alpha}{\alpha} \sum_{l=1}^{\ell} \frac{1}{\gamma_{l} - 1}.
\end{align}
Then, we have
\begin{align}
    c_{l} 
    &= \frac{1}{\sigma^2} \frac{q + \alpha}{\alpha} \frac{u_{l}}{1 - u_{1}}
    = \frac{1}{\sigma^2} \left(1 + \frac{\eta^2}{\alpha (\alpha-1)}\right) \frac{u_{l}}{1 - u_{1}},
\end{align}
which gives
\begin{align}
    c_{l} = \frac{1}{\sigma^2} \left(1 + \frac{\eta^2}{\alpha (\alpha-1)}\right) \frac{\gamma_{l-1}}{\gamma_{l-1} - 1} \cdots \frac{\gamma_{1}}{\gamma_{1} - 1}  \sum_{j=l}^{\ell} \frac{1}{\gamma_{l} - 1}
\end{align}
for all $l$, where the empty product is interpreted as unity. These results are positive throughout the region of physical validity we expect from our analysis of the replica-nonuniform components; recalling that $w_{l} > 0$ for these solutions, no further conditions are imposed.

We now consider the solutions with 
\begin{align}
    r_{l_{\ast}} &= \frac{\gamma_{l_{\ast}}}{\alpha - \gamma_{l_{\ast}}}  
\end{align}
for some $l_{\ast} = 1, \ldots, \ell$. For these solutions, we have $z \hat{z} \to  - \gamma_{l_{\ast}}$ and $w_{l} \to 0$ for all $l \leq l_{\ast}$. From our previous analysis, we have the condition $\gamma_{l} \geq \gamma_{l_{\ast}}$ for all $l > l_{\ast}$. If $l_{\ast} = 1$, the initial difference condition reduces to 
\begin{align}
    u_{1} = \frac{1}{\gamma_{1}}
\end{align}
and 
\begin{align}
    \frac{\gamma_{l-1}}{\gamma_{l} } u_{l-1} 
    &= \left[ 1 + \frac{\gamma_{l-1}}{\gamma_{l}}  \frac{(\gamma_{l-1} - \gamma_{1})^2 -\gamma_{1}^{2}}{\gamma_{l-1}^{2}} \right] u_{l}  \nonumber\\&\quad - \frac{(\gamma_{l} - \gamma_{1})^2}{\gamma_{l}^{2}} u_{l+1} 
\end{align}
If $l_{\ast} > 1$, we have the recurrence 
\begin{align}
    \frac{\gamma_{l-1}}{\gamma_{l} } u_{l-1} 
    &= \left[ 1 + \frac{\gamma_{l-1}}{\gamma_{l}}  \frac{(\gamma_{l-1} - \gamma_{l_{\ast}})^2 - \gamma_{l_{\ast}}^{2}}{\gamma_{l-1}^{2}} \right] u_{l}  \nonumber\\&\quad - \frac{(\gamma_{l} - \gamma_{l_{\ast}})^2}{\gamma_{l}^{2}} u_{l+1} 
\end{align}
with initial difference condition 
\begin{align}
    \frac{1}{\gamma_{1}} = u_{1} - \left(\frac{\gamma_{1} - \gamma_{l_{\ast}}}{\gamma_{1}}\right)^{2} u_{2} 
\end{align}
and endpoint condition $u_{\ell+1} = 0$. Precisely at $l_{\ast}$, we have the simplification
\begin{align}
    u_{l_{\ast}-1} = \frac{ \gamma_{l_{\ast}-1} - \gamma_{l_{\ast}}}{\gamma_{l_{\ast}-1}} u_{l_{\ast}} . 
\end{align}
Iterating one step backward, we find that
\begin{align}
    u_{l_{\ast}-2} = \frac{\gamma_{l_{\ast}-2} - \gamma_{l_{\ast}}}{\gamma_{l_{\ast}-2}} u_{l_{\ast}-1} . 
\end{align}
It is then easy to see that we can iterate further back to obtain, for $l < l_{\ast}$, 
\begin{align}
    u_{l} = \frac{\gamma_{l} - \gamma_{l_{\ast}}}{\gamma_{l}} u_{l+1}. 
\end{align}
In particular, we have
\begin{align}
    u_{1} = \frac{\gamma_{1} - \gamma_{l_{\ast}}}{\gamma_{1}} u_{2}.
\end{align}
hence the initial difference equation yields
\begin{align}
    u_{1} = \frac{1}{\gamma_{l_{\ast}}}. 
\end{align}
For $2 \leq l \leq l_{\star}$, this yields the solution
\begin{align}
    u_{l} = \frac{1}{\gamma_{l_{\ast}}} \frac{\gamma_{1} - \gamma_{l_{\ast}}}{\gamma_{1}} \cdots \frac{\gamma_{l-1} - \gamma_{l_{\ast}}}{\gamma_{l-1}} .
\end{align}
The limiting equation for $q$ is then
\begin{align}
    q &= (1 - \gamma_{l_{\ast}})^{2} + \gamma_{l_{\ast}} \frac{1 + \alpha -  \gamma_{l_{\ast}}  }{\alpha}  (q + \eta^2) - \eta^{2} \gamma_{l_{\ast}},
\end{align}
which yields
\begin{align}
    q = \alpha \frac{1 - \gamma_{l_{\ast}}}{\alpha - \gamma_{l_{\ast}}} + \frac{\gamma_{l_{\ast}}}{\alpha - \gamma_{l_{\ast}}} \eta^{2} . 
\end{align}
By the same reasoning as in our analysis of the case $r = 1 / (\alpha-1)$, we have the limiting closed set of equations
\begin{align}
    c_{l}  = \frac{1}{\sigma^2} \gamma_{l_{\ast}}^2 \bigg[ 1 + \frac{1}{\alpha} q + \sigma^{2} c_{1} \bigg] u_{l} .
\end{align}
Using the fact that $u_{1} = 1/\gamma_{l_{\ast}}$, we have
\begin{align}
    c_{1} &= \frac{1}{\sigma^2} \gamma_{l_{\ast}} \bigg[ 1 + \frac{1}{\alpha} q + \sigma^{2} c_{1} \bigg] ,
\end{align}
which yields
\begin{align}
    c_{1} &= \frac{1}{\sigma^2} \frac{q + \alpha}{\alpha} \frac{\gamma_{l_{\ast}}}{1 - \gamma_{l_{\ast}}},
\end{align}
and thus, for $l > 1$,
\begin{align}
    c_{l} 
    &= \frac{1}{\sigma^2} \gamma_{l_{\ast}}^2 \bigg[ 1 + \frac{1}{\alpha} q + \sigma^{2} c_{1} \bigg] u_{l} 
    = \frac{1}{\sigma^2} \frac{q + \alpha}{\alpha} \frac{\gamma_{l_{\ast}}^2}{1 - \gamma_{l_{\ast}}} u_{l}. 
\end{align}
Then, for $2 \leq l \leq l_{\ast}$, we have
\begin{align}
    c_{l} &= \frac{1}{\sigma^2} \frac{q + \alpha}{\alpha} \frac{\gamma_{l_{\ast}}}{1 - \gamma_{l_{\ast}}} \frac{\gamma_{1} - \gamma_{l_{\ast}}}{\gamma_{1}} \cdots \frac{\gamma_{l-1} - \gamma_{l_{\ast}}}{\gamma_{l-1}} 
\end{align}
using the solution for $u_{l}$ obtained above. As $w_{l} = 0$ for $l \leq l_{\ast}$, we must have $c_{l} \geq 0$ for $l \leq l_{\ast}$ in order for these solutions to be physical, hence we conclude that we must have $\gamma_{l} \geq \gamma_{l_{\ast}}$ for all $l < l_{\ast}$. As $w_{l} \geq 0$ for $l > l_{\ast}$, we will not obtain further conditions on the physical validity of these solutions by solving for $c_{l}$ for $l > l_{\ast}$, hence we will not attempt to do so.

\subsection{NN model} \label{app:nn_rs_saddle_point}

For a deep network, the RS saddle point is determined by the $2(\ell+2)$-dimensional system of equations
\begin{align}
    \hat{z} &= - \frac{\alpha}{\beta^{-1} + z}
    \\
    \hat{q} &= \frac{\alpha (q+\eta^{2})}{(\beta^{-1} + Q - q)^2}
    \\
    z &= \frac{\sigma^{2} C_{1} }{1 - \sigma^{2} C_{1} \hat{z}}  
    \\
    q &= \frac{1 + \sigma^{4} C_{1}^2 \hat{q}}{(1 - \sigma^{2} C_{1} \hat{z})^2} 
    \\
    \hat{C}_{1} &= \frac{\sigma^2}{\gamma_{1}} \bigg[\frac{\hat{q} + \hat{z} ( 1 + \hat{z} - \sigma^{2} C_{1} \hat{z})}{(1 - \sigma^{2} C_{1} \hat{z})^2}\bigg]
    \\
    \hat{C}_{l} &= \frac{\gamma_{l-1}}{\gamma_{l}} \frac{\hat{C}_{l-1}}{1-\hat{C}_{l-1} C_{l} } && (l=2,\ldots,\ell) 
    \\
    C_{l} &= \frac{C_{l+1}}{1- C_{l+1} \hat{C}_{l}} && (l=1,\ldots,\ell-1) 
    \\
    C_{\ell} &= \frac{1}{1 - \hat{C}_{\ell}}
\end{align}
where, as before, we have defined $z \equiv Q-q$ and $\hat{z} \equiv \hat{Q}-\hat{q}$ for brevity. Unlike for the RF model, in this case the replica-uniform and replica-nonuniform components do not decouple nicely. However, we have fewer equations to solve. Moreover, we can exclude solutions with $C_{l} \sim \mathcal{O}(1/\beta)$, as they will be trivial. 

From the condition
\begin{align}
    z &= \frac{\sigma^{2} C_{1} }{1 - \sigma^{2} C_{1} \hat{z}}  ,
\end{align}
we have
\begin{align}
    \sigma^{2} C_{1} \hat{z} = \frac{z \hat{z}}{1 + z \hat{z}}
\end{align}
hence
\begin{align}
    q &= \frac{1 + \sigma^{4} C_{1}^2 \hat{q}}{(1 - \sigma^{2} C_{1} \hat{z})^2} = (1 + z \hat{z})^{2} + z^{2} \hat{q} 
    \\
    &= (1 + z \hat{z})^{2} + \frac{1}{\alpha} (z \hat{z})^{2} (q + \eta^2),
\end{align}
where we have noted that $\hat{q} = \hat{z}^{2} (q + \eta^2)/\alpha$. 

We now seek to eliminate the Lagrange multipliers $\hat{C}_{l}$ and all of the order parameters $C_{l}$ except for $C_{1}$. To do so, we will follow our earlier analysis of the RF model. We define
\begin{align}
    A \equiv z \frac{\hat{q} + \hat{z} ( 1 + \hat{z} - \sigma^{2} C_{1} \hat{z})}{(1 - \sigma^{2} C_{1} \hat{z})} 
\end{align}
such that
\begin{align}
    \hat{C}_{1} = \frac{A}{\gamma_{1} C_{1}} .
\end{align}
If $\ell=1$, then we can solve the equation
\begin{align}
    C_{1} &= \frac{1}{1 - \hat{C}_{1}} = \frac{1}{1 - \frac{A}{\gamma_{1} C_{1}}}
\end{align}
yielding
\begin{align}
    C_{1} = \frac{\gamma_{1} + A}{\gamma_{1}} ,
\end{align}
which will allow us to close the equations. 

We now consider deeper networks ($\ell > 1$), following the solution techniques we used for the RF model. We observe that a solution with any $C_{l} = 0$ must have all $C_{l} = 0$ and $z = 0$. Similarly, a solution with one $\hat{C}_{l} = 0$ must have all $\hat{C}_{l} = 0$ and $\hat{z} = 0$. As $C_{\ell} = 1/(1-\hat{C}_{\ell})$, these situations cannot coexist. Moreover, neither is self-consistent unless $\alpha = 0$ or $\beta$ is strictly infinite or zero. With this observation in mind, we will eliminate the Lagrange multipliers $\hat{C}_{l}$ using the same method as we did for the replica-nonuniform components of the Lagrange multipliers in the RF case. Formally defining $C_{l+1} \equiv 1$ for convenience, we have
\begin{align}
    \hat{C}_{l} = \frac{C_{l} - C_{l+1}}{C_{l}C_{l+1}}  
\end{align}
for $l = 1, \ldots, \ell$. Then, for $l = 2, \ldots, \ell$, the equation
\begin{align}
    \hat{C}_{l} &= \frac{\gamma_{l-1}}{\gamma_{l}} \frac{\hat{C}_{l-1}}{1-\hat{C}_{l-1} w_{l}}
\end{align}
yields
\begin{align}
    \frac{C_{l} - C_{l+1}}{C_{l}C_{l+1}} = \frac{\gamma_{l-1}}{\gamma_{l}} \frac{C_{l-1}}{C_{l}} \frac{C_{l-1} - C_{l}}{C_{l-1} C_{l}} .
\end{align}
Using the abovementioned fact that we can write
\begin{align}
    \hat{C}_{1} = \frac{A}{\gamma_{1} C_{1}},
\end{align}
we can see that this set of equations is analogous to what we obtained for the deviations from uniformity $w_{l}$ and $\hat{w}_{l}$ in the RF case (with, in that case, $A = z \hat{z}$). Thus, using the results of our previous calculation, we conclude that
\begin{align}
    C_{l} = \frac{(\gamma_{l} + A) (\gamma_{l+1} + A) \cdots (\gamma_{\ell} + A)}{\gamma_{l} \gamma_{l+1} \cdots \gamma_{\ell}} ,
\end{align}
which gives us closed set of equations for $z$, $\hat{z}$, $q$, $\hat{q}$, and $C_{1}$. 

With the scaling $z \sim \mathcal{O}(1)$, we have $z \hat{z} \to - \alpha$, and the condition on $q$ becomes
\begin{align}
    q = (1-\alpha)^2 + \alpha (q + \eta^{2}),
\end{align}
which yields
\begin{align}
    q = 1 - \alpha + \frac{\alpha}{1-\alpha} \eta^{2}.
\end{align}
To determine the limiting condition on $z$, we note that
\begin{align}
    A = \frac{1}{z} \alpha (1 - \alpha + \eta^{2}) - \alpha
\end{align}
in this limit, hence
\begin{align}
    (\gamma_{l} + A) z = (\gamma_{l} - \alpha) z + \alpha (1 - \alpha + \eta^{2}).
\end{align}
Therefore, we have the polynomial condition
\begin{align}
    z^{\ell+1} = \sigma^{2} (1 - \alpha) \prod_{l=1}^{\ell} \left[ \frac{( \gamma_{l} - \alpha ) z + \alpha (1 - \alpha + \eta^{2})}{\gamma_{l}} \right].
\end{align}
Given a candidate positive solution to this degree-($\ell+1$) polynomial, we must then verify that it yields a positive value for all 
\begin{align}
    C_{l} = \frac{1}{z^{l}} \prod_{l'=l}^{\ell} \frac{(\gamma_{l'} - \alpha)z + \alpha (1 - \alpha + \eta^{2}) }{\gamma_{l'} } 
\end{align}
in order for it to be a nontrivial physical solution. This implies that we must have
\begin{align}
    \frac{(\gamma_{l} - \alpha) z + \alpha (1 - \alpha + \eta^{2}) }{\gamma_{l} z} > 0 
\end{align}
for all $l$. 

For a network with a single hidden layer ($\ell=1$), the condition is just a quadratic, with solutions
\begin{widetext}
\begin{align}
    z_{\pm} = (1-\alpha) \frac{\sigma^{2} (\gamma_{1} - \alpha) \pm \sqrt{\sigma^{4} (\gamma_{1} - \alpha)^{2} + 4 \alpha \gamma_{1} \sigma^{2} (1 - \alpha + \eta^2)/(1-\alpha) }}{2 \gamma_{1}}. 
\end{align}
\end{widetext}
For any $\gamma_{1},\sigma>0$, $z_{+} \geq 0$ if $0 < \alpha < 1$, and $z_{-} \geq 0$ if $\alpha > 1$. However, noting that 
\begin{align}
    C_{1} = \frac{z}{\sigma^{2} (1-\alpha)},
\end{align}
the $z_{+}$ solution yields a non-negative value for $C_{1}$, and is therefore physical for $0<\alpha<1$, while the $z_{-}$ solution yields a non-positive value, and is therefore unphysical. 

For a network with more than a single hidden layer, the polynomial condition cannot be analytically solved in a useful way. However, we can gain some insight by solving it perturbatively in the large-width regime $\gamma_{1},\cdots,\gamma_{\ell} \gg \alpha$. Concretely, we introduce a formal expansion parameter $\lambda$, and solve the equation
\begin{align}
    z^{\ell+1} = \sigma^{2} (1-\alpha) \prod_{l=1}^{\ell} \left[ \left(1-\lambda\frac{\alpha}{\gamma_{l}}\right) z + \lambda \frac{\alpha}{\gamma_{l}} (1 - \alpha + \eta^2) \right]
\end{align}
order-by-order in $\lambda$ with the \emph{Ansatz}
\begin{align}
    z = \sum_{j=0}^{\infty} z_{j} \lambda^{j}.
\end{align}
It is easy to see that the zeroth-order condition yields
\begin{align}
    z_{0} = \sigma^{2} (1-\alpha) ,
\end{align}
and that the first-order term yields
\begin{align}
    z_{1} = [(1-\sigma^{2})(1 - \alpha) + \eta^2 ] \sum_{l=1}^{\ell} \frac{\alpha}{\gamma_{l}} .
\end{align}
To go to higher order, it is convenient to specialize to the case of equal hidden layer widths $\gamma_{1}=\gamma_{2}=\cdots=\gamma_{\ell}=\gamma$, both to simplify the calculations and to make the results easier to interpret. In this case, we can directly define the expansion parameter as $\lambda \equiv \alpha/\gamma$, hence the equation we want to solve becomes
\begin{align}
    z^{\ell+1} = \sigma^{2} (1-\alpha) \left[ (1-\lambda) z + \lambda (1 - \alpha + \eta^2) \right]^{\ell} .
\end{align}
In this simplified setting, it is relatively straightforward to work out by hand or with the aid of Mathematica that
\begin{align}
    z_{2} = (1 - \alpha + \eta^2) \left( \frac{\ell(\ell-1) \tilde{\sigma}^2}{2} - \frac{\ell(\ell+1)}{2\tilde{\sigma}^{2}} + \ell\right)
\end{align}
with $\tilde{\sigma}$ as in \eqref{eqn:sigma_tilde}, which yields the result reported in the main text.

For solutions with $z \sim \mathcal{O}(\beta^{-1})$ and $C_{1} \sim \mathcal{O}(1)$, we have the limiting equation
\begin{align}
    q = \frac{1}{\alpha} (q+\eta^2), 
\end{align}
which implies that we should have
\begin{align}
    q = \frac{\eta^2}{\alpha - 1} .
\end{align}
As $z \to 0$, we must have $q \geq 0$, hence this solution makes sense for all $\alpha > 1$. To solve for $C_{l}$ for these solutions, it is most convenient to express $A$ in terms of $C_{1} \sim \mathcal{O}(1)$. Noting that 
\begin{align}
    \hat{C}_{1} \to \frac{1}{\gamma_{1} C_{1}} \frac{1 - \sigma^{2} C_{1} + \eta^{2}/(\alpha-1)}{\sigma^{2} C_{1}} ,
\end{align}
we have
\begin{align}
    A \to \frac{1 - \sigma^{2} C_{1} + \eta^{2}/(\alpha-1)}{\sigma^{2} C_{1}} ,
\end{align}
hence $C_{1}$ is determined by degree-($\ell+1$) polynomial
\begin{align}
    \sigma^{2\ell} C_{1}^{\ell+1} = \prod_{l=1}^{\ell} \frac{1 + \sigma^{2} (\gamma_{l}-1) C_{1} + \eta^2/(\alpha-1)}{\gamma_{l}} .
\end{align}
Given a candidate positive solution for $C_{1}$, we can then determine $C_{l}$ for all $l>1$ via
\begin{align}
    C_{l} = \prod_{l'=l}^{\ell} \frac{\gamma_{l'} + A}{\gamma_{l'}} = \prod_{l'=l}^{\ell}  \frac{1 + \sigma^{2} (\gamma_{l'}-1) C_{1} + \eta^2/(\alpha-1)}{\gamma_{l'} \sigma^{2} C_{1}} .
\end{align}
This implies that we must have
\begin{align}
    C_{1} > \frac{1}{\sigma^{2} (1-\gamma_{l})} \left(1 + \frac{\eta^2}{\alpha-1}\right) 
\end{align}
for all $l$ (including $l=1$) in order for the candidate solution to be physical and non-trivial. As we are interested only in learning curves, we will not analyze this equation further.

\section{Direct computation of posterior expectations for LR and RF models}\label{app:rf_posterior_expectations}

For the LR and RF models, we can evaluate the zero-temperature posterior expectation in the definition of $\epsilon$ analytically. In particular, writing
\begin{align}
    \mathbf{F} \equiv \frac{\sigma}{\sqrt{d n_{1} \cdots n_{\ell}}} \mathbf{U}_{1} \cdots \mathbf{U}_{\ell}
\end{align}
for brevity, we have the posterior moment generating function for $\mathbf{v}$:
\begin{align}
    \mathcal{Z}(\mathbf{j}) 
    &\propto \int d\mathbf{v}\, \exp\left(-\frac{\beta}{2} \Vert \mathbf{X} \mathbf{F} \mathbf{v} - \mathbf{y} \Vert^{2} - \frac{1}{2} \Vert \mathbf{v} \Vert^{2} + \mathbf{j}^{\top} \mathbf{v} \right)
    \\
    &\propto \exp\bigg( \beta \mathbf{y}^{\top} \mathbf{X} \mathbf{F} (\mathbf{I}_{n_{\ell}} + \beta \mathbf{F}^{\top} \mathbf{X}^{\top} \mathbf{X} \mathbf{F})^{-1} \mathbf{j} \nonumber\\&\qquad\qquad + \frac{1}{2} \mathbf{j}^{\top} (\mathbf{I}_{n_{\ell}} + \beta \mathbf{F}^{\top} \mathbf{X}^{\top} \mathbf{X} \mathbf{F})^{-1} \mathbf{j} \bigg), 
\end{align}
where the implied constants of proportionality are independent of the source $\mathbf{j}$. Then, as $\mathbf{w} = \sqrt{d} \mathbf{F} \mathbf{v}$. the posterior mean and covariance of the end-to-end weight vector are given as
\begin{align}
    \langle \mathbf{w} \rangle = \sqrt{d}  \mathbf{F} (\beta^{-1} \mathbf{I}_{n_{\ell}} + \mathbf{F}^{\top} \mathbf{X}^{\top} \mathbf{X} \mathbf{F})^{-1} \mathbf{F}^{\top} \mathbf{X}^{\top} \mathbf{y}
\end{align}
and 
\begin{align}
    \langle \mathbf{w} \mathbf{w}^{\top} \rangle - \langle \mathbf{w} \rangle \langle \mathbf{w} \rangle^{\top} 
    = d \mathbf{F} (\mathbf{I}_{n_{\ell}} + \beta \mathbf{F}^{\top} \mathbf{X}^{\top} \mathbf{X} \mathbf{F})^{-1} \mathbf{F}^{\top},
\end{align}
respectively. We note that $\langle \mathbf{w} \rangle$ is simply the RF ridge regression estimator with ridge parameter $1/\beta$, as
\begin{align}
    \langle \mathbf{v} \rangle = \argmin_{\mathbf{v}} \left( \Vert \mathbf{X} \mathbf{F} \mathbf{v} - \mathbf{y} \Vert^{2} + \frac{1}{\beta} \Vert \mathbf{v} \Vert^{2} \right). 
\end{align}
The thermal bias-variance decomposition of the zero-temperature generalization error is then given as
\begin{align}
    \varepsilon_{b} 
    &\equiv \lim_{\beta \to \infty} \frac{1}{d} \Vert \langle \mathbf{w} \rangle - \mathbf{w}_{\ast} \Vert^{2} 
    \\
    &= \lim_{\beta \to \infty} \left\Vert  \mathbf{F} (\beta^{-1} \mathbf{I}_{n_{\ell}} + \mathbf{F}^{\top} \mathbf{X}^{\top} \mathbf{X} \mathbf{F})^{-1} \mathbf{F}^{\top} \mathbf{X}^{\top} \mathbf{y} - \frac{\mathbf{w}_{\ast}}{\sqrt{d}} \right\Vert^{2},
    \\
    \varepsilon_{v} &\equiv \lim_{\beta \to \infty} \frac{1}{d} \tr[\langle \mathbf{w} \mathbf{w}^{\top} \rangle - \langle \mathbf{w} \rangle \langle \mathbf{w} \rangle^{\top}] 
    \\
    &= \lim_{\beta \to \infty} \tr[\mathbf{F} (\mathbf{I}_{n_{\ell}} + \beta \mathbf{F}^{\top} \mathbf{X}^{\top} \mathbf{X} \mathbf{F})^{-1} \mathbf{F}^{\top}]. 
\end{align}
In terms of these quantities, we have
\begin{align}
    \epsilon = \lim_{d,p,n_{1},\ldots,n_{\ell} \to \infty} \mathbb{E}_{\mathcal{D}} (\varepsilon_{b}  + \varepsilon_{v}). 
\end{align}
With our data model, $\mathbf{X}$ has rank $\min\{p,d\}$ with probability one, while $\mathbf{F}$ has rank $\min\{d,n_{1},\ldots,n_{\ell}\}$ with probability one \cite{muirhead2009aspects}. Moreover, $\mathbf{X} \mathbf{F}$ has rank $\min\{p,d,n_{1},\ldots,n_{\ell}\}$ with probability one. 

If all $n_{1}, \ldots, n_{\ell} > d$ and $p > d$, both $\mathbf{F} \mathbf{F}^{\top}$ and $\mathbf{X}^{\top} \mathbf{X}$ are invertible with probability one, and, applying the push-through identity \cite{horn2012matrix}, we have
\begin{align}
    \varepsilon_{b} &= \frac{1}{d} \Vert \sqrt{d}  ( \mathbf{X}^{\top} \mathbf{X} )^{-1} \mathbf{X}^{\top} \mathbf{y} - \mathbf{w}_{\ast} \Vert^{2},
    \\
    \varepsilon_{v} &= 0. 
\end{align}

If $p < \min\{d,n_{1}, \ldots, n_{\ell}\}$, then the matrix $\mathbf{F} \mathbf{F}^{\top} \mathbf{X}^{\top} \mathbf{X}$ will not be invertible, but the matrix $\mathbf{X} \mathbf{F} \mathbf{F}^{\top} \mathbf{X}^{\top}$ will be invertible with probability one, even if $\mathbf{F} \mathbf{F}^{\top}$ is not. Then, with another application of the push-through identity and the aid of the Woodbury identity \cite{horn2012matrix}, we have
\begin{align}
    \varepsilon_{b} 
    &= \frac{1}{d} \Vert \sqrt{d}  \mathbf{F} \mathbf{F}^{\top} \mathbf{X}^{\top} ( \mathbf{X} \mathbf{F} \mathbf{F}^{\top} \mathbf{X}^{\top})^{-1}  \mathbf{y} - \mathbf{w}_{\ast} \Vert^{2},
    \\
    \varepsilon_{v} 
    &= \tr(\mathbf{F} \mathbf{F}^{\top} ) -  \tr[  \mathbf{F} \mathbf{F}^{\top} \mathbf{X}^{\top} ( \mathbf{X} \mathbf{F} \mathbf{F}^{\top} \mathbf{X}^{\top} )^{-1} \mathbf{X} \mathbf{F} \mathbf{F}^{\top} ]  .
\end{align}

Finally, if $p > \min\{n_{1},\ldots,n_{\ell}\}$ but $\min\{n_{1},\ldots,n_{\ell}\} < d$, the situation is somewhat more complicated. Let $l_{\textrm{min}} = \argmin n_{l}$ be the index of the narrowest layer. Then, let
\begin{align}
    \mathbf{A} &= \frac{\sigma}{\sqrt{d n_{1} \cdots n_{l_{\textrm{min}}}}} \mathbf{U}_{1} \cdots \mathbf{U}_{l_{\textrm{min}}} \in \mathbb{R}^{d \times n_{\textrm{min}}}
    \\
    \mathbf{B} &= \frac{1}{\sqrt{n_{l_{\textrm{min}}+1} \cdots n_{\ell}}} \mathbf{U}_{l_{\textrm{min}}+1} \cdots \mathbf{U}_{\ell} \in \mathbb{R}^{n_{\textrm{min}} \times n_{\ell}}
\end{align}
such that 
\begin{align}
    \mathbf{F} = \mathbf{A} \mathbf{B} .
\end{align}
Under the stated assumptions, the matrices $\mathbf{A}^{\top} \mathbf{X}^{\top} \mathbf{X} \mathbf{A}$ and $\mathbf{B} \mathbf{B}^{\top}$ are invertible with probability one, as is their product. Then, we have
\begin{align}
    \varepsilon_{b} &= \frac{1}{d} \Vert \sqrt{d}  \mathbf{A} ( \mathbf{A}^{\top} \mathbf{X}^{\top} \mathbf{X} \mathbf{A} )^{-1} \mathbf{A}^{\top} \mathbf{X}^{\top} \mathbf{y} - \mathbf{w}_{\ast} \Vert^{2},
    \\
    \varepsilon_{v} &= 0 .
\end{align}

We observe that, under the re-scaling of the feature map $\mathbf{F} \mapsto \sigma \mathbf{F}$ for any $\sigma > 0$, $\varepsilon_{b}$ is always constant, while $\varepsilon_{v}$ is either identically zero or degree-two homogeneous in $\sigma$. This suggests that we should be able to read off the ridgeless results from our Bayesian replica results. We also note that we have recovered the three-region phase diagram indicated by our replica calculation.

For completeness, we also remark that we can use these results to directly compute $\epsilon_{\textrm{LR}}$ without the use of the replica trick. For the LR model, we have
\begin{align}
    \mathbf{F} = \frac{\sigma}{\sqrt{d}} \mathbf{I}_{d}
\end{align}
and two phases: $p<d$ and $p>d$. We first consider the regime $p<d$. Using the fact that $\mathbb{E}_{\mathbf{w}_{\ast}} \mathbf{w}_{\ast} \mathbf{w}_{\ast}^{\top} = \mathbf{I}_{d}$ and the formula for the expectation of an inverse Wishart matrix with identity scale matrix \cite{muirhead2009aspects}:
\begin{align}
    \mathbb{E}_{\mathbf{X}} \tr[  (\mathbf{X} \mathbf{X}^{\top})^{-1}  ] = \frac{p}{d-p-1},
\end{align}
a short computation yields
\begin{align}
    \lim_{n,p\to\infty} \mathbb{E}_{\mathbf{X},\xi,\mathbf{w}_{\ast}} \varepsilon_{b} &= 1 - \alpha + \frac{\alpha}{1-\alpha} \eta^{2},
    \\
    \lim_{n,p\to\infty} \mathbb{E}_{\mathbf{X},\xi,\mathbf{w}_{\ast}} \varepsilon_{v} &= \sigma^{2} (1-\alpha). 
\end{align}
In the regime $p > d$, we have
\begin{align}
    \lim_{n,p\to\infty} \mathbb{E}_{\mathbf{X},\xi,\mathbf{w}_{\ast}} \varepsilon_{b} &= \frac{1}{\alpha - 1} \eta^{2},
    \\
    \lim_{n,p\to\infty} \mathbb{E}_{\mathbf{X},\xi,\mathbf{w}_{\ast}} \varepsilon_{v} &= 0, 
\end{align}
using the fact that
\begin{align}
    \mathbb{E}_{\mathbf{X}} \tr[  (\mathbf{X}^{\top} \mathbf{X} )^{-1}  ] = \frac{d}{p-d-1}
\end{align}
in this regime. This recovers the result of our replica computation. 

We remark that a similar, albeit more complex, procedure would likely allow one to derive the learning curve for a deep RF model rigorously using properties of products of large Gaussian random matrices \cite{akemann2013products}. However, from a physical perspective, the non-rigorous replica theory approach used here has the advantages of being more transparent and of allowing a relatively unified treatment of NN models.

\begin{widetext}

\section{Direct computation of posterior expectations for NN models}\label{app:nn_posterior_expectations}

In this appendix, we show that the zero-temperature posterior expectation in the definition of $\epsilon$ can be evaluated semi-analytically for NNs. Our approach mirrors that of our previous work in \cite{zv2021scale}: we will integrate out the weights of the first hidden layer ($\mathbf{U}_{1}$) exactly, yielding expressions for the posterior mean and variance of the end-to-end weight vector in terms of expectations over the remaining weights. These results follow by applying the results of \cite{zv2021scale} to a test dataset of $d$ examples with trivial data matrix $\hat{\mathbf{X}} = \sqrt{d} \mathbf{I}_{d}$ and then passing to the zero-temperature limit, but we will provide a detailed derivation for completeness.

Writing
\begin{align}
    \mathbf{f} \equiv \frac{\sigma}{\sqrt{d n_{1} \cdots n_{\ell}}} \mathbf{U}_{2} \cdots \mathbf{U}_{\ell} \mathbf{v}
\end{align}
for brevity, such that $\mathbf{w} = \sqrt{d} \mathbf{U}_{1} \mathbf{f}$, we can write the posterior moment generating function of $\mathbf{w}$ as
\begin{align}
    \mathcal{Z}(\mathbf{j}) 
    &\propto \mathbb{E}_{\mathcal{W} \setminus \mathbf{U}_{1}} \int d\mathbf{U}_{1}\, \exp\left(-\frac{\beta}{2} \Vert \mathbf{X} \mathbf{U}_{1} \mathbf{f} - \mathbf{y} \Vert^{2} - \frac{1}{2} \Vert \mathbf{U}_{1} \Vert^{2} + \sqrt{d} \mathbf{j}^{\top} \mathbf{U}_{1} \mathbf{f} \right),
\end{align}
where we discard irrelevant constants of proportionality. This matrix Gaussian integral can be conveniently evaluated through vectorization \cite{magnus2019matrix}. Using standard properties of the Kronecker product, we find that
\begin{align}
    \mathcal{Z}(\mathbf{j}) \propto \mathbb{E}_{\mathcal{W} \setminus \mathbf{U}_{1}}\bigg[\rho(\Vert \mathbf{f} \Vert^2)  \exp\bigg( & \beta \sqrt{d} \Vert \mathbf{f} \Vert^{2} \mathbf{y}^{\top} ( \mathbf{I}_{p} + \beta \Vert \mathbf{f} \Vert^{2} \mathbf{X} \mathbf{X}^{\top} )^{-1} \mathbf{X} \mathbf{j} \nonumber\\& + \frac{1}{2} d \Vert \mathbf{f} \Vert^{2} \mathbf{j}^{\top} [\mathbf{I}_{d} - \beta \Vert \mathbf{f} \Vert^{2} \mathbf{X}^{\top} ( \mathbf{I}_{p} + \beta \Vert \mathbf{f} \Vert^{2} \mathbf{X} \mathbf{X}^{\top} )^{-1} \mathbf{X} ] \mathbf{j} \bigg) \bigg],
\end{align}
where
\begin{align}
    \rho(\Vert \mathbf{f} \Vert^2) \equiv \det( \mathbf{I}_{p} + \beta \Vert \mathbf{f} \Vert^{2} \mathbf{X} \mathbf{X}^{\top} )^{-1/2} \exp\left(-\frac{1}{2} \beta \mathbf{y}^{\top} ( \mathbf{I}_{p} + \beta \Vert \mathbf{f} \Vert^{2} \mathbf{X} \mathbf{X}^{\top} )^{-1} \mathbf{y} \right) .
\end{align}
By varying this result with respect to the source, we thus obtain
\begin{align}
    \langle \mathbf{w} \rangle = \frac{\mathbb{E}_{\mathcal{W} \setminus \mathbf{U}_{1}} [\rho \mathbf{z} ]}{\mathbb{E}_{\mathcal{W} \setminus \mathbf{U}_{1}} \rho} 
\end{align}
and
\begin{align}
    \langle \mathbf{w} \mathbf{w}^{\top} \rangle - \langle \mathbf{w} \rangle  \langle \mathbf{w} \rangle^{\top} 
    &= \frac{\mathbb{E}_{\mathcal{W} \setminus \mathbf{U}_{1}} \left\{ \rho d \Vert \mathbf{f} \Vert^{2} [\mathbf{I}_{d} - \beta \Vert \mathbf{f} \Vert^{2} \mathbf{X}^{\top} ( \mathbf{I}_{p} + \beta \Vert \mathbf{f} \Vert^{2} \mathbf{X} \mathbf{X}^{\top} )^{-1} \mathbf{X} ] \right\}}{\mathbb{E}_{\mathcal{W} \setminus \mathbf{U}_{1}} \rho} + \frac{\mathbb{E}_{\mathcal{W} \setminus \mathbf{U}_{1}}[\rho \mathbf{z} \mathbf{z}^{\top}]}{\mathbb{E}_{\mathcal{W} \setminus \mathbf{U}_{1}} \rho}   - \langle \mathbf{w} \rangle \langle \mathbf{w} \rangle^{\top},
\end{align}
where we have defined
\begin{align}
    \mathbf{z} \equiv \beta \sqrt{d} \Vert \mathbf{f} \Vert^{2} \mathbf{X}^{\top} ( \mathbf{I}_{p} + \beta \Vert \mathbf{f} \Vert^{2} \mathbf{X} \mathbf{X}^{\top} )^{-1} \mathbf{y}
\end{align}
for brevity. This matches the result of applying \cite{zv2021scale}'s expressions to a trivial dataset with $\hat{\mathbf{X}} = \sqrt{d} \mathbf{I}_{d}$.

As for the RF model, we introduce a thermal bias-variance decomposition
\begin{align}
    \varepsilon_{b} 
    &\equiv \lim_{\beta \to \infty} \frac{1}{d} \Vert \langle \mathbf{w} \rangle - \mathbf{w}_{\ast} \Vert^{2} 
    \\
    \varepsilon_{v} &\equiv \lim_{\beta \to \infty} \frac{1}{d} \tr[\langle \mathbf{w} \mathbf{w}^{\top} \rangle - \langle \mathbf{w} \rangle \langle \mathbf{w} \rangle^{\top}] .
\end{align}
For any set of hidden layer widths, $\Vert \mathbf{f} \Vert^{2}$ is almost surely positive. Therefore, as the only matrix inverses present in these expressions are of the form $( \mathbf{I}_{p} + \beta \Vert \mathbf{f} \Vert^{2} \mathbf{X} \mathbf{X}^{\top} )^{-1}$, the NN model should have two phases: $p<d$ and $p>d$.

If $p<d$, then the matrix $\mathbf{X} \mathbf{X}^{\top}$ is invertible with probability one. Then, up to (divergent) multiplicative constants which will cancel in the ratios of expectations, we have the almost-sure pointwise limit
\begin{align}
    \lim_{\beta \to \infty} \rho \propto \Vert \mathbf{f} \Vert^{-p} \exp\left(-\frac{\mathbf{y}^{\top} (\mathbf{X} \mathbf{X}^{\top})^{-1} \mathbf{y}}{2 \Vert \mathbf{f} \Vert^{2}} \right) .
\end{align}
Similarly, we have the almost-sure limits
\begin{align}
    \lim_{\beta \to \infty} \mathbf{z} = \sqrt{d} \mathbf{X}^{\top} ( \mathbf{X} \mathbf{X}^{\top} )^{-1} \mathbf{y}
\end{align}
and
\begin{align}
    \lim_{\beta \to \infty} \Vert \mathbf{f} \Vert^{2} \tr[\mathbf{I}_{d} - \beta \Vert \mathbf{f} \Vert^{2} \mathbf{X}^{\top} ( \mathbf{I}_{p} + \beta \Vert \mathbf{f} \Vert^{2} \mathbf{X} \mathbf{X}^{\top} )^{-1} \mathbf{X} ]
    = (d-p) \Vert \mathbf{f} \Vert^{2} .
\end{align}
Therefore, noting that $\lim_{\beta \to \infty} \mathbf{z}$ is almost surely a constant function of $\mathbf{f}$, we have
\begin{align}
    \varepsilon_{b} &= \frac{1}{d} \Vert \sqrt{d} \mathbf{X}^{\top} ( \mathbf{X} \mathbf{X}^{\top} )^{-1} \mathbf{y} - \mathbf{w}_{\ast} \Vert^{2} 
    \\
    \varepsilon_{v} &= (1-\alpha) d \frac{\mathbb{E}_{\mathcal{W} \setminus \mathbf{U}_{1}}[\Vert \mathbf{f} \Vert^{2-p} \exp(-\mathbf{y}^{\top} (\mathbf{X} \mathbf{X}^{\top})^{-1} \mathbf{y} / 2 \Vert \mathbf{f} \Vert^{2} )]}{\mathbb{E}_{\mathcal{W} \setminus \mathbf{U}_{1}}[\Vert \mathbf{f} \Vert^{-p} \exp(-\mathbf{y}^{\top} (\mathbf{X} \mathbf{X}^{\top})^{-1} \mathbf{y} / 2 \Vert \mathbf{f} \Vert^{2} )]} .
\end{align}

If $p>d$, then the matrix $\mathbf{X} \mathbf{X}^{\top}$ is invertible with probability zero, but the matrix $\mathbf{X}^{\top} \mathbf{X}$ is invertible with probability one. By the Weinstein-Aronzjan identity,
\begin{align}
    \det( \mathbf{I}_{p} + \beta \Vert \mathbf{f} \Vert^{2} \mathbf{X} \mathbf{X}^{\top} ) = \det( \mathbf{I}_{p} + \beta \Vert \mathbf{f} \Vert^{2} \mathbf{X}^{\top}\mathbf{X}  ),
\end{align}
hence the determinant factors in $\rho$ will yield a factor of $\Vert \mathbf{f} \Vert^{-d}$ in the zero-temperature limit. We must be more careful in considering the exponential term in $\rho$. Letting the orthonormal eigendecomposition of $\mathbf{X}\mathbf{X}^{\top}$ be
\begin{align}
    \mathbf{X}\mathbf{X}^{\top} = \sum_{j=1}^{p} \chi_{j} \mathbf{m}_{j} \mathbf{m}_{j}^{\top}, 
\end{align}
we have the low-temperature Neumann series \cite{horn2012matrix}
\begin{align}
    \beta ( \mathbf{I}_{p} + \beta \Vert \mathbf{f} \Vert^{2} \mathbf{X} \mathbf{X}^{\top} )^{-1} = \beta \sum_{\{j\,:\,\chi_{j}=0\}} \mathbf{m}_{j} \mathbf{m}_{j}^{\top} + \frac{1}{\Vert \mathbf{f} \Vert^{2}} \sum_{\{j\,:\,\chi_{j} > 0\}} \frac{1}{\chi_{j}} \mathbf{m}_{j} \mathbf{m}_{j}^{\top} + \mathcal{O}(\beta^{-1}),
\end{align}
hence the divergent null-space projector term $\beta \sum_{\{j\,:\,\chi_{j}=0\}} \mathbf{m}_{j} \mathbf{m}_{j}^{\top}$ does not depend on $\mathbf{f}$, and will therefore cancel in the ratio of expectations. Thus, we have 
\begin{align}
    \lim_{\beta \to \infty} \rho \propto \Vert \mathbf{f} \Vert^{-d} \exp\left(-\frac{1}{2 \Vert \mathbf{f} \Vert^{2}} \sum_{\{j\,:\,\chi_{j} > 0\}} \frac{1}{\chi_{j}} (\mathbf{m}_{j}^{\top} \mathbf{y})^2 \right) .
\end{align}
By a simple application of the push-through identity, we have the almost-sure pointwise limits
\begin{align}
    \lim_{\beta \to \infty} \mathbf{z} =  \sqrt{d} (\mathbf{X}^{\top} \mathbf{X} )^{-1} \mathbf{X}^{\top} \mathbf{y}
\end{align}
and
\begin{align}
    \lim_{\beta \to \infty} \Vert \mathbf{f} \Vert^{2} \tr[\mathbf{I}_{d} - \beta \Vert \mathbf{f} \Vert^{2} \mathbf{X}^{\top} ( \mathbf{I}_{p} + \beta \Vert \mathbf{f} \Vert^{2} \mathbf{X} \mathbf{X}^{\top} )^{-1} \mathbf{X} ] = 0.
\end{align}
Therefore, noting that $\lim_{\beta \to \infty} \mathbf{z}$ is once again almost surely a constant function of $\mathbf{f}$, we have
\begin{align}
    \varepsilon_{b} &= \frac{1}{d} \Vert \sqrt{d} (\mathbf{X}^{\top} \mathbf{X} )^{-1} \mathbf{X}^{\top} \mathbf{y} - \mathbf{w}_{\ast} \Vert^{2} ,
    \\
    \varepsilon_{v} &= 0.
\end{align}

Comparing this result to the discussion of the LR model in Appendix \ref{app:rf_posterior_expectations}, we can see that the bias terms in each phase are identical to those for the LR model, hence we can apply the results given there for their dataset averages. This shows that the learning curve for the NN model is of the form \eqref{eqn:nn_learning_curve}, with
\begin{align} 
    z 
    &= \lim_{d,p,n_{1},\ldots,n_{\ell} \to \infty} \mathbb{E}_{\mathcal{D}} \varepsilon_{v} 
    \\
    &= (1-\alpha) \lim_{d,p,n_{1},\ldots,n_{\ell} \to \infty} d \mathbb{E}_{\mathcal{D}} \left\{ \frac{\mathbb{E}_{\mathcal{W} \setminus \mathbf{U}_{1}}[\Vert \mathbf{f} \Vert^{2-p} \exp(-\mathbf{y}^{\top} (\mathbf{X} \mathbf{X}^{\top})^{-1} \mathbf{y} / 2 \Vert \mathbf{f} \Vert^{2} )]}{\mathbb{E}_{\mathcal{W} \setminus \mathbf{U}_{1}}[\Vert \mathbf{f} \Vert^{-p} \exp(-\mathbf{y}^{\top} (\mathbf{X} \mathbf{X}^{\top})^{-1} \mathbf{y} / 2 \Vert \mathbf{f} \Vert^{2} )]} \right\}. \label{eqn:z_ratio}
\end{align}
We remark that evaluation of the outer dataset average without resorting to the replica trick seems likely to be challenging.

For a network with a single hidden layer, we can evaluate the average over $\mathcal{W} \setminus \mathbf{U}_{1} = \mathbf{v}$ analytically. In this case, we have
\begin{align}
    \Vert \mathbf{f} \Vert^2 = \frac{\sigma^2}{n_{1} d} \Vert \mathbf{v} \Vert^{2},
\end{align}
hence
\begin{align}
    d \frac{\mathbb{E}_{\mathbf{v}}[\Vert \mathbf{f} \Vert^{2-p} \exp(-\mathbf{y}^{\top} (\mathbf{X} \mathbf{X}^{\top})^{-1} \mathbf{y} / 2 \Vert \mathbf{f} \Vert^{2} )]}{\mathbb{E}_{\mathbf{v}}[\Vert \mathbf{f} \Vert^{-p} \exp(-\mathbf{y}^{\top} (\mathbf{X} \mathbf{X}^{\top})^{-1} \mathbf{y} / 2 \Vert \mathbf{f} \Vert^{2} )]} 
    &= \frac{\sigma^2}{n_{1} } \frac{\mathbb{E}_{\mathbf{v}}[\Vert \mathbf{v} \Vert^{2-p} \exp(-q^2 / 2 \Vert \mathbf{v} \Vert^{2} )]}{\mathbb{E}_{\mathbf{v}}[\Vert \mathbf{v} \Vert^{-p} \exp(-q^2 / 2 \Vert \mathbf{v} \Vert^{2} )]} 
\end{align}
where we have defined
\begin{align}
    q^{2} \equiv \frac{n_{1}d}{\sigma^2} \mathbf{y}^{\top} (\mathbf{X} \mathbf{X}^{\top})^{-1} \mathbf{y}
\end{align}
for brevity. Using the fact that $\Vert \mathbf{v} \Vert^{2} \sim \chi^{2}(n_{1})$ under the prior, we have
\begin{align}
    \frac{\mathbb{E}_{\mathbf{v}}[\Vert \mathbf{v} \Vert^{2-p} \exp(-q^2 / 2 \Vert \mathbf{v} \Vert^{2} )]}{\mathbb{E}_{\mathcal{W} \setminus \mathbf{U}_{1}}[\Vert \mathbf{v} \Vert^{-p} \exp(-q^2 / 2 \Vert \mathbf{v} \Vert^{2} )]} 
    &= \frac{\int_{0}^{\infty} dt\, t^{(n_{1}-p)/2-1+1} \exp(-t/2 - q^{2}/2t)}{\int_{0}^{\infty} dt\, t^{(n_{1}-p)/2-1} \exp(-t/2 - q^{2}/2t)}
    = q \frac{K_{(n_{1}-p)/2+1}(q)}{K_{(n_{1}-p)/2}(q)},
\end{align}
where $K_{\nu}(z)$ is a modified Bessel function of the second kind \cite{dlmf}. Thus, for a NN with a single hidden layer, we have
\begin{align}
    z = \sigma^{2} (1-\alpha) \lim_{d,p,n_{1},\ldots,n_{\ell} \to \infty} \mathbb{E}_{\mathcal{D}} \left\{ \frac{1}{n_{1}} q \frac{K_{(n_{1}-p)/2+1}(q)}{K_{(n_{1}-p)/2}(q)} \right\},
\end{align}
with $q$ defined as above. 

\end{widetext}

In general, it is not immediately clear how to evaluate the limit of the nested averages in \eqref{eqn:z_ratio}. However, as argued by Li and Sompolinsky \cite{li2021statistical} in the case $\gamma_{1} = \cdots = \gamma_{\ell} = \gamma$, one can make progress under the assumption that the quantity 
\begin{align}
    r \equiv \frac{1}{\sigma^{2} \alpha} \mathbf{y}^{\top} (\mathbf{X} \mathbf{X}^{\top})^{-1} \mathbf{y}
\end{align}
rapidly concentrates about its limiting mean value, which is
\begin{align}
    \lim_{d,p\to\infty} \mathbb{E}_{\mathbf{X},\mathbf{w}_{\ast},\bm{\xi}} r = \frac{1}{\tilde{\sigma}^2} 
\end{align}
by the results of Appendix \ref{app:rf_posterior_expectations}. Then, assuming that the outer dataset average in \eqref{eqn:z_ratio} can be evaluated by replacing all occurrences of $r$ with $\tilde{\sigma}^{-2}$, let us make a layer-by-layer saddle-point approximation of the integrals over $\mathcal{W} \setminus \mathbf{U}_{1}$. Li and Sompolinsky \cite{li2021statistical}'s general fixed-data analysis is not set up by first integrating out $\mathbf{U}_{1}$ as above, but, as discussed in our previous work \cite{zv2021scale} for the case $\ell=1$, can be related to this approach. Moreover, their analysis of the Gaussian covariate model (presented in Appendix D of the Supplemental Material of \cite{li2021statistical}), is framed in terms of an approximation in which correlations are neglected, but amounts to the concentration assumption stated above. 

Then, by equations (26) and (27) of the main text of \cite{li2021statistical}, or by equations (24), (26), and (30) of their Supplemental Material, the result of their approximation is that
\begin{align}
    z = \sigma^{2} (1-\alpha) u^{\ell}
\end{align}
for $u$ a solution of
\begin{align}
    1 - u = \lambda (1 - u^{-\ell} \tilde{\sigma}^{-2}) ,
\end{align}
with $\lambda \equiv \alpha/\gamma$. We would like to show that this is consistent with the result of our RS calculation, which implies that $z$ should be a non-negative root of
\begin{align}
    z^{\ell+1} = \sigma^{2} (1-\alpha) [(1-\lambda)z+\lambda (1-\alpha+\eta^2)]^{\ell} .
\end{align}
Substituting in the \emph{Ansatz} $z=\sigma^{2} (1-\alpha) u^{\ell}$, we have
\begin{align}
    u^{\ell(\ell+1)} = [(1-\lambda) u^{\ell} + \lambda \tilde{\sigma}^{-2} ]^{\ell} .
\end{align}
Taking the $\ell$-th root of both sides, we obtain
\begin{align}
    u^{\ell+1} = (1-\lambda) u^{\ell} + \lambda \tilde{\sigma}^{-2} 
\end{align}
hence, dividing by $u^{\ell}$ under the assumption that it is positive and re-arranging terms, we obtain
\begin{align}
    1 - u = \lambda (1 - u^{-\ell} \tilde{\sigma}^{-2}) ,
\end{align}
which recovers Li and Sompolinsky's result. 

\section{Comparison to large-width perturbative calculations with fixed data}\label{app:perturbative_comparison}

In this appendix, we compare the results of the replica theory calculation of the generalization error of a deep linear network in the present work to our previous perturbative results in \cite{zv2021asymptotics}. Concretely, Appendix G of \cite{zv2021asymptotics} computes the leading finite-width correction to the posterior-averaged error on a test set of $\hat{p}$ examples for fixed data in the regime $\alpha < 1$. As noted there and in \cite{zv2021scale}, this can be roughly interpreted as the leading-order correction in $\ell p/n$, as the correction that is parameterically $\mathcal{O}(1/n)$ in $n$ scales with $\ell$ and $p$. To make contact with the present work, we evaluate their result for a test dataset of $d$ examples, with trivial data matrix $\hat{\mathbf{X}} = \sqrt{d} \mathbf{I}_{d}$. Then, the perturbative approximation for the thermal bias-variance decomposition given in \cite{zv2021asymptotics} is
\begin{widetext}
\begin{align}
    \varepsilon_{b} &= \frac{1}{d} \Vert \sqrt{d} \mathbf{X}^{\top} (\mathbf{X} \mathbf{X}^{\top})^{-1} \mathbf{y} - \mathbf{w}_{\ast} \Vert^{2}
    \\
    \varepsilon_{v} &= \frac{1}{d} \tr[\mathbf{I}_{d} - \mathbf{X}^{\top} (\mathbf{X} \mathbf{X}^{\top})^{-1} \mathbf{X}  ] \left[ \sigma^{2} + \left(\sum_{l=1}^{\ell} \frac{\alpha}{\gamma_{l}}\right) \left(\frac{d}{p} \mathbf{y}^{\top} (\mathbf{X} \mathbf{X}^{\top})^{-1} \mathbf{y} - \sigma^{2} \right) \right]  + \mathcal{O}(n^{-2}).
\end{align}
\end{widetext}
We remark that the results of \cite{zv2021asymptotics,zv2021scale} show that it should be safe to interchange the limit $\beta \to \infty$ with the high-dimensional limit and the expectation over data. Using the fact that $\mathbf{X} \mathbf{X}^{\top}$ is invertible with probability one in this regime, the disorder average of the thermal bias term yields
\begin{align}
    \lim_{n,p\to\infty} \mathbb{E}_{\mathbf{X},\bm{\xi},\mathbf{w}_{\ast}} \varepsilon_{b} = 1 - \alpha + \frac{\alpha}{1-\alpha} \eta^{2},
\end{align}
where we have again used the formula for the expectation of an inverse Wishart matrix with identity scale matrix \cite{muirhead2009aspects}. Similarly, the disorder average of the thermal variance term yields
\begin{align}
    \lim_{n,p \to \infty} \mathbb{E}_{\mathbf{X},\bm{\xi},\mathbf{w}_{\ast}}  \varepsilon_{v} &= (1-\alpha) \sigma^{2} \nonumber\\&\quad + [(1-\alpha)(1 - \sigma^{2}) + \eta^{2} ] \sum_{l=1}^{\ell} \frac{\alpha}{\gamma_{l}} \nonumber\\&\quad + \mathcal{O}(n^{-2}),
\end{align}
Therefore, the perturbative result of \cite{zv2021asymptotics} implies a large-width disorder-averaged generalization error of
\begin{align}
    \epsilon_{\textrm{NN}} &= \epsilon_{\textrm{LR}}
    + [(1-\sigma^{2})(1 - \alpha) + \eta^2 ] \sum_{l=1}^{\ell} \frac{\alpha}{\gamma_{l}} + \mathcal{O}(n^{-2}), 
\end{align}
which agrees with the leading-order large-width solution of the RS result reported here. This makes sense, as we intuitively expect possible RSB effects to emerge at smaller width. Moreover, we remark that we have an exact correspondence between $Q-q$ and $q$ and the averages of the thermal variance and bias terms, respectively. Finally, we note that the coefficient of the $\mathcal{O}(\alpha/\gamma)$ correction, which is the dataset average of $d \mathbf{y}^{\top} (\mathbf{X} \mathbf{X}^{\top})^{-1} \mathbf{y}/p - \sigma^{2}$, gives the dataset average of the condition for when increasing width helps generalization noted by \cite{li2021statistical,zv2021asymptotics}.

\section{Detailed analysis of optimal network architecture}\label{app:optimal_architecture}

In this appendix, we provide a detailed analysis of how width and depth affect generalization in RF and NN models. 

\subsection{Optimal width for RF models} \label{app:rf_optimal_width}

We first consider optimizing the width of a deep random feature model. In the regime $\alpha < \min\{1,\gamma_{\textrm{min}}\}$, we have
\begin{align}
    \frac{\partial \epsilon_{\textrm{RF}}}{\partial \gamma_{l}} = \frac{\alpha (1-\alpha+\eta^{2})}{\gamma_{l}^{2}} \left[ \tilde{\sigma}^{2}  \prod_{l' \neq l} \frac{\gamma_{l'}-\alpha}{\gamma_{l'}} - \left(\frac{\gamma_{l}}{\gamma_{l}-\alpha}\right)^{2} \right] .
\end{align}
where we have defined
\begin{align}
    \tilde{\sigma}^{2} \equiv \frac{1 - \alpha}{1 - \alpha + \eta^{2}} \sigma^{2} 
\end{align}
as in \eqref{eqn:sigma_tilde}. If $\ell = 1$, this is simply
\begin{align}
    \frac{\partial \epsilon_{\textrm{RF}}}{\partial \gamma_{1}} = \frac{\alpha (1-\alpha+\eta^{2})}{\gamma_{1}^{2}} \left[\tilde{\sigma}^{2} - \left(\frac{\gamma_{1}}{\gamma_{1}-\alpha}\right)^{2} \right] .
\end{align}
For $\tilde{\sigma} \leq 1$, $\partial \epsilon_{\textrm{RF}}/\partial \gamma_{1} < 0$ for all parameter values in this regime, and $\epsilon_{\textrm{RF}}$ is minimized by taking $\gamma_{1} \to \infty$. If $\tilde{\sigma} > 1$, there is a valid (i.e., $\gamma_{1} > \alpha$) stationary point $\partial \epsilon_{\textrm{RF}}/\partial \gamma_{1} = 0$ at
\begin{align}
    \gamma_{\star} = \frac{\tilde{\sigma}}{\tilde{\sigma}-1} \alpha . 
\end{align}
For $\gamma_{1} < \gamma_{\star}$, $\partial \epsilon_{\textrm{RF}}/\partial \gamma_{1} < 0$, while for $\gamma_{1} > \gamma_{\star}$, $\partial \epsilon_{\textrm{RF}}/\partial \gamma_{1} > 0$. This point is therefore a minimum of $\epsilon_{\textrm{RF}}$. 

We now consider $\ell > 1$. In this parameter regime, we have
\begin{align}
    \prod_{l' \neq l} \frac{\gamma_{l'}-\alpha}{\gamma_{l'}} \leq 1
    \quad \textrm{and} \quad
    \left(\frac{\gamma_{l}}{\gamma_{l}-\alpha}\right)^{2} \geq 1 ,
\end{align}
where both inequalities are strict if all $\gamma_{l} < \infty$. Thus, if $\tilde{\sigma} \leq 1$, $\partial \epsilon_{\textrm{RF}} / \partial \gamma_{l}$ is always negative for all $l$, and $\epsilon_{\textrm{RF}}$ is minimized by taking all $\gamma_{l} \to \infty$. If $\tilde{\sigma} > 1$, we have a nontrivial stationary point with all
\begin{align}
    \gamma_{l} = \gamma_{\star} \qquad (l = 1, \ldots, \ell)
\end{align}
for
\begin{align}
    \frac{\gamma_{\star} - \alpha}{\gamma_{\star}} = \tilde{\sigma}^{-2/(\ell+1)} , 
\end{align}
which gives 
\begin{align}
    \gamma_{\star} = \frac{\tilde{\sigma}^{2/(\ell+1)}}{\tilde{\sigma}^{2/(\ell+1)} - 1} \alpha .
\end{align}
To check whether this is indeed a local minimum, we compute the Hessian of $\epsilon_{\textrm{RF}}$ at the stationary point, which is given by
\begin{align}
    \frac{\partial^2 \epsilon_{\textrm{RF}}}{\partial \gamma_{l} \partial \gamma_{l'}} \bigg|_{\gamma_{l} = \gamma_{\star}} =  \frac{\alpha^2 (1-\alpha+\eta^2)}{\gamma_{\star}^4} \tilde{\sigma}^{6/(\ell+1)} \left[ \delta_{ll'} + 1 \right] .
\end{align}
Diagonalizing this matrix is trivial, yielding eigenvalue
\begin{align}
    \frac{\alpha^2 (1-\alpha+\eta^2)}{\gamma_{\star}^4} \tilde{\sigma}^{6/(\ell+1)}
\end{align}
with multiplicity $\ell-1$ and eigenvalue
\begin{align}
    \frac{\alpha^2 (1-\alpha+\eta^2)}{\gamma_{\star}^4} \tilde{\sigma}^{6/(\ell+1)} (\ell+1)
\end{align}
with multiplicity one. Both of these eigenvalues are positive throughout the parameter region of interest, confirming that the Hessian is positive-definite at the stationary point. Moreover, substituting $\gamma_{\star}$ into the generalization error $\epsilon_{\textrm{RF}}$, we have
\begin{align}
    \epsilon_{\textrm{RF}} \bigg|_{\gamma_{1}=\cdots=\gamma_{\ell}=\gamma_{\star}} 
    &=  \epsilon_{\textrm{LR}} + (1-\alpha) \sigma^{2} \left(\tilde{\sigma}^{-2\ell/(\ell+1)} - 1 \right) \nonumber\\&\quad + \ell (1 - \alpha + \eta^2) \left(\tilde{\sigma}^{2/(\ell+1)}-1\right)
\end{align}
for any fixed $\alpha < \min\{1,\gamma_{\star}\}$. As $\tilde{\sigma} > 1$, both correction terms are negative. and this result is a monotonically deceasing function of the network depth $\ell$. It can therefore be seen that this result yields better generalization than that obtained by taking any subset of the hidden layers to infinity while keeping the remainder fixed at $\gamma_{\star}$. This result has the interesting feature that, for any fixed $\tilde{\sigma} > 1$ and $\alpha$, $\gamma_{\star}$ is a monotonically increasing function of $\ell$.

For RF models in the regime $\alpha > \gamma_{\textrm{min}}$ for $\gamma_{\textrm{min}} < 1$, it is easy to see that $\epsilon_{\textrm{RF}}$ is a monotonically decreasing function of $\gamma_{\textrm{min}} \in [0,1)$ if $\alpha > 1$, while if $\alpha < 1$, it is a monotonically increasing function of $\gamma_{\textrm{min}} \in [0,\alpha)$. However, in this regime, it is important to keep track of crossings in the ordering of different layer widths.

\subsection{Optimal depth for RF models} \label{app:rf_optimal_depth}

We now consider optimizing the depth of a RF model. For simplicity, we specialize to the case in which all hidden layers have the same width, i.e., $\gamma_{1} = \cdots = \gamma_{\ell} = \gamma$. Our starting point is therefore the expression \eqref{eqn:rf_equal_width_generalization}, which can sensibly be analytically continued in width. For brevity, we let
\begin{align}
    \psi \equiv \frac{\gamma - \alpha}{\gamma} ,
\end{align}
which is bounded as $0 < \psi < 1$ in the regime of interest. This gives
\begin{align}
    \frac{\epsilon_{\textrm{RF}} - \epsilon_{\textrm{LR}}}{1-\alpha + \eta^2} = \tilde{\sigma}^{2} ( \psi^{\ell} - 1 ) + \ell \left(\frac{1}{\psi} - 1 \right) ,
\end{align}
where we have defined $\tilde{\sigma}$ as in \eqref{eqn:sigma_tilde}. Treating $\ell$ as a continuous parameter, we then have
\begin{align}
    \frac{\partial \epsilon_{\textrm{RF}}}{\partial \ell} 
    &= (1-\alpha+\eta^{2}) \left[ \tilde{\sigma}^{2} \psi^{\ell} \log\left( \psi \right) + \frac{1}{\psi} - 1 \right], 
\end{align}
For all $\tilde{\sigma} > 0$ and $0 < \psi < 1$, we can see that $\partial \epsilon_{\textrm{RF}}/\partial \ell$ is a monotonically increasing function of $\ell$, as can be confirmed by inspecting
\begin{align}
    \frac{\partial^2 \epsilon_{\textrm{RF}}}{\partial \ell^2} = (1-\alpha+\eta^{2}) \tilde{\sigma}^{2} \psi^{\ell} \log(\psi)^2 > 0.
\end{align}
Using the lower bound $\log(\psi) > 1-1/\psi$, which is strict for all $0<\psi<1$, we have 
\begin{align}
    \frac{\partial \epsilon_{\textrm{RF}}}{\partial \ell} > (1-\alpha+\eta^{2}) \left( 1 - \frac{1}{\psi} \right) (1 - \tilde{\sigma}^{2} \psi^{\ell}).
\end{align}
For $\tilde{\sigma} \leq 1$, we have $\tilde{\sigma}^{2} \psi^{\ell} < 1$ for all $0 < \psi < 1$ and all $\ell \geq 1$, hence the above bound shows that 
\begin{align}
    \frac{\partial \epsilon_{\textrm{RF}}}{\partial \ell} > 0,
\end{align}
implying that $\epsilon_{\textrm{RF}}$ is a monotonically increasing function of $\ell$ if $\tilde{\sigma} \leq 1$. Therefore, shallow random feature models are optimal in this regime. This is consistent with our earlier observations regarding optimal width, as taking hidden layer widths to infinity reduces the effective depth.

If $\tilde{\sigma} > 1$, then the above lower bound shows that $\partial \epsilon_{\textrm{RF}} / \partial \ell > 0$ if $\psi \leq \tilde{\sigma}^{-2/\ell}$, i.e., if $\ell \geq - \log(\tilde{\sigma}^2)/\log(\psi)$. However, in this case, $\epsilon_{\textrm{RF}}$ is not always an increasing function of $\ell$. In particular, using the upper bound $\log(\psi) < \psi(1-1/\psi)$, which is again strict for $0<\psi<1$, we have the upper bound
\begin{align}
    \frac{\partial \epsilon_{\textrm{RF}}}{\partial \ell} 
    < (1-\alpha+\eta^{2}) \left(\frac{1}{\psi} - 1\right) ( 1 - \tilde{\sigma}^{2} \psi^{\ell+1} ) ,
\end{align}
which shows that $\partial \epsilon_{\textrm{RF}}/\partial \ell < 0$ for all $\psi \geq \tilde{\sigma}^{-2/(\ell+1)}$, i.e., if $\ell \leq - \log(\tilde{\sigma}^2)/\log(\psi) - 1$. We remark that the optimized value $\gamma_{\star}$ found above in our study of optimal width is covered by this crude bound, as it is defined by the condition $\psi |_{\gamma = \gamma_{\star}} = \tilde{\sigma}^{-2/(\ell+1)}$.

In terms of $\ell$, the intermediate region not covered by the bounds above is the open interval 
\begin{align}
    - \frac{\log(\tilde{\sigma}^2)}{\log(\psi)} - 1 < \ell < - \frac{\log(\tilde{\sigma}^2)}{\log(\psi)} . 
\end{align}
To the left of this interval, we know that $\partial \epsilon_{\textrm{RF}}/\partial \ell < 0$, while to the right of this interval, we know that $\partial \epsilon_{\textrm{RF}}/\partial \ell > 0$. Therefore, for any $\ell$ outside this open interval, $\epsilon_{\textrm{RF}}$ is strictly greater than its values anywhere within the interval. 

If $- {\log(\tilde{\sigma}^2)}/{\log(\psi)}$ is not an integer, this open interval will always include exactly one integer value of $\ell$, 
\begin{align}
    \ell_{\star} = \left\lfloor -\frac{\log(\tilde{\sigma}^2)}{\log(\psi)} \right\rfloor,
\end{align}
which will thus minimize the generalization error for fixed $\tilde{\sigma} > 1$ and $0<\psi < 1$. Restoring the definition of $\psi$ in terms of $\gamma$ and $\alpha$, this optimal value is 
\begin{align}
    \ell_{\star} = \left\lfloor \frac{\log(\tilde{\sigma}^2)}{\log[\gamma/(\gamma-\alpha)]} \right\rfloor .
\end{align}

If $- {\log(\tilde{\sigma}^2)}/{\log(\psi)}$ is an integer, then no integers are contained within the open interval that is not covered by the bounds on $\partial \epsilon_{\textrm{RF}}/\partial \ell$ derived above. In that case, the bounds on $\partial \epsilon_{\textrm{RF}}/\partial \ell$ imply that the optimal depth is then given by one of the boundary points, i.e.,
\begin{align}
    \ell_{\star} \in \left\{j,j-1 \right\},
\end{align}
where we have defined the positive integer (noting that $-\log(\tilde{\sigma}^2)/\log(\psi) > 0$ for all $\tilde{\sigma} > 1$ and $0 < \psi < 1$)
\begin{align}
    j \equiv -\frac{\log(\tilde{\sigma}^2)}{\log(\psi)} \in \mathbb{N}_{>0}.
\end{align}
For candidate depths of this form, we can use the fact that $\psi^{j} = \psi^{-{\log(\tilde{\sigma}^2)}/{\log(\psi)}} = \tilde{\sigma}^{-2}$ to obtain
\begin{align}
    \frac{\epsilon_{\textrm{RF}} |_{\ell = j + k} - \epsilon_{\textrm{RF}} |_{\ell = j}}{1-\alpha+\eta^{2}} &= k \left(\frac{1}{\psi} - 1 \right) + \psi^{k} - 1
\end{align}
for any $k \in \mathbb{Z}$ such that $j + k \geq 0$. For any $0 < \psi < 1$, this generalization gap is positive for all $k < -1$, vanishes when $k=-1$, is negative for $-1 < k < 0$, vanishes for $k=0$, and is positive for all $k>0$. To show this, we first observe that the gap is a smooth function of $\psi$ for any $k$ (including non-integer real values), with 
\begin{align}
    \left[ k \left(\frac{1}{\psi} - 1 \right) + \psi^{k} - 1\right] \bigg|_{\psi = 1} = 0 .
\end{align}
Then, we consider
\begin{align}
    \frac{\partial}{\partial \psi} \left[ k \left(\frac{1}{\psi} - 1 \right) + \psi^{k} - 1\right] = k \psi^{-2} (\psi^{k+1} - 1) ,
\end{align}
which is strictly negative for all $0 < \psi < 1$ and vanishes from below as $\psi \uparrow 1$ if $k < -1$ or if $k > 0$, and is strictly positive for all $0 < \psi < 1$ and vanishes from above as $\psi \uparrow 1$ if $-1 < k < 0$. This shows that the desired claim should hold. Therefore, in this case the two candidate depths will yield identical generalization error, and we can take either $\ell_{\star} = j$ or $\ell_{\star} = j-1$. 

We note that the condition $ -{\log(\tilde{\sigma}^2)}/{\log(\psi)} = j \in \mathbb{N}_{>0}$ implies that 
\begin{align}
    \psi = \tilde{\sigma}^{-2/j} , 
\end{align}
which gives a condition on the width as a function of $\tilde{\sigma}^2$ and $\alpha$:
\begin{align}
    \gamma = \frac{\tilde{\sigma}^{2/j}}{\tilde{\sigma}^{2/j} - 1} \alpha .
\end{align}
In our earlier study of the optimal width for fixed depth, we found solutions of precisely this form, with $j = \ell + 1$. This result for the optimal depth at fixed width is therefore internally consistent with our earlier result for the optimal width at fixed depth. 

\subsection{Optimal width for NN models} \label{app:nn_optimal_width}

For a two-layer NN in the regime $\alpha < 1$, we have
\begin{align}
    \frac{\partial \epsilon_{\textrm{NN}}}{\partial \gamma_{1}} &= \frac{\alpha (1-\alpha) \sigma^{2}}{2\gamma_{1}^{2}} \left[ 1 + \frac{  (\gamma_{1} - \alpha) - 2 \gamma_{1} \tilde{\sigma}^{-2}}{  \sqrt{(\gamma_{1}-\alpha)^2 + 4 \alpha \gamma_{1} \tilde{\sigma}^{-2} }} \right]
\end{align}
where we have defined $\tilde{\sigma}$ as in \eqref{eqn:sigma_tilde}. If $\tilde{\sigma} = 1$, we have
\begin{align}
    \frac{  (\gamma_{1} - \alpha) - 2 \gamma_{1} \tilde{\sigma}^{-2}}{  \sqrt{(\gamma_{1}-\alpha)^2 + 4 \alpha \gamma_{1} \tilde{\sigma}^{-2} }} \bigg|_{\tilde{\sigma} = 1} = - \frac{\gamma_{1} + \alpha}{  \sqrt{(\gamma_{1}+\alpha)^2 }} = -1, 
\end{align}
hence ${\partial \epsilon_{\textrm{NN}}}/{\partial \gamma_{1}}|_{\tilde{\sigma}=1} = 0$. Exactly at $\alpha = \gamma_{1}$, we have
\begin{align}
    \frac{  (\gamma_{1} - \alpha) - 2 \gamma_{1} \tilde{\sigma}^{-2}}{  \sqrt{(\gamma_{1}-\alpha)^2 + 4 \alpha \gamma_{1} \tilde{\sigma}^{-2} }}\bigg|_{\alpha = \gamma_{1}} = - \frac{1}{\tilde{\sigma}},
\end{align}
If $\alpha \neq \gamma_{1}$, we have
\begin{align}
    \frac{  (\gamma_{1} - \alpha) - 2 \gamma_{1} \tilde{\sigma}^{-2}}{  \sqrt{(\gamma_{1}-\alpha)^2 + 4 \alpha \gamma_{1} \tilde{\sigma}^{-2} }} 
    &> - \frac{ \gamma_{1} + \alpha }{  \sqrt{(\gamma_{1}-\alpha)^2 + 4 \alpha \gamma_{1} \tilde{\sigma}^{-2} }} 
    \nonumber\\
    &> - \frac{\gamma_{1} + \alpha}{  \sqrt{(\gamma_{1}+\alpha)^2 }} = -1
\end{align}
if  $\tilde{\sigma} > 1$; the reversed chain of strict inequalities holds if $\tilde{\sigma} < 1$. Therefore, we conclude that, for any $\alpha$ and $\gamma_{1}$,
\begin{align}
    \frac{\partial \epsilon_{\textrm{NN}}}{\partial \gamma_{1}} = \begin{dcases} < 0, & 0 < \tilde{\sigma} < 1
    \\
    =0, & \tilde{\sigma} = 1,
    \\
    > 0, & \tilde{\sigma} > 1,
    \end{dcases}
\end{align}
as reported in the main text.

\subsection{Difference in generalization in two-layer NN and RF models} \label{app:generalization_gap}

For two-layer networks, we have the RF-NN generalization gap
\begin{align}
    \frac{\epsilon_{\textrm{RF}} - \epsilon_{\textrm{NN}}}{1-\alpha+\eta^2}
    &= 1 - \frac{1}{\psi} + \tilde{\sigma}^{2} \frac{ \psi - \sqrt{\psi^2 + 4 \alpha /\gamma_{1} \tilde{\sigma}^{2}}}{2 }
\end{align}
in the regime $\alpha < \min\{1,\gamma_{1}\}$, where $\tilde{\sigma}$ is defined in \eqref{eqn:sigma_tilde} and $\psi \equiv (\gamma_{1}-\alpha)/\gamma_{1}$. For any finite $\gamma_{1}$, we have $0 < \psi < 1$, hence it is easy to see that both $1-1/\psi$ and $\psi - \sqrt{\psi^2 + 4\alpha/\gamma_{1}\tilde{\sigma}^2}$ are negative. If $\gamma_{1} \to \infty$, then $\psi \to 1$, and $\epsilon_{\textrm{RF}}-\epsilon_{\textrm{NN}} \downarrow 0$. This yields the conclusion reported in the main text.

\section{Numerical methods} \label{app:numerical_methods}

Below, we elaborate on the numerical methods used to validate the replica-symmetric learning curves. 
In all of the numerical simulations, we set the input dimensionality $d = 100$, and sample with a resolution of around 50 - 100 estimates per dimension. To produce the standard error bars, we sample 10 values per estimate. Numerical procedures were written with NumPy \cite{numpy} and SciPy \cite{scipy}. Plots were generated with Matplotlib \cite{matplotlib}.

\subsection{Random feature model}

Theoretical predictions for the generalization error in Bayesian random feature models can be computed directly from equation \eqref{eqn:rf_learning_curve}. Additionally, after sampling a set of initial weights, we use the results from Appendix \ref{app:rf_posterior_expectations} to directly compute the posterior expectations for the RF model. By then averaging the resulting error across multiple samples of weights, we numerically verify our theoretical curves.

\subsection{Neural network model}

Theoretical predictions for the generalization error in a Bayesian Neural Network can be computed directly from equation \eqref{eqn:nn_learning_curve}. We then use the results from Appendix \ref{app:nn_posterior_expectations} to directly compute the error for particular instantiations of weights, numerically verifying our theoretical results.

\addcontentsline{toc}{section}{References}
\bibliography{refs}

\end{document}